\documentclass[journal]{IEEEtran}
\usepackage{ifpdf}

\usepackage{amsmath}
\usepackage{CJKutf8}
\usepackage{array}
\usepackage{graphicx}
\graphicspath{{./}}
\usepackage{float}
\usepackage{algorithm}
\usepackage{algpseudocode}
\usepackage{amssymb}
\usepackage{subfloat}
\usepackage{subfig}

\usepackage[normalem]{ulem}  % 加在导言区

\usepackage{booktabs}
\usepackage{multirow}
\usepackage{xcolor}

\usepackage{cite}
\usepackage[colorlinks=true, linkcolor=blue]{hyperref}

\begin{document}
\begin{CJK}{UTF8}{gbsn}

\bibliographystyle{IEEEtranS}
%\bibliography{bib/IEEEexample}
%\bibliography{bib/IEEEexample}

%\title{Bare Demo of IEEEtran.cls\\ for IEEE Journals}
%\title{Latent Diffusion Model-based Efficient Framework for Wave Impedance Inversion}
\title{Seismic Acoustic Impedance Inversion Framework Based on Conditional Latent Generative Diffusion Model}
%with Conditional Latent Generative Diffusion Model
%把基于隐式扩散模型的高效波阻抗反演框架。改了
\author{Jie Chen,
		Hongling Chen,
        Jinghuai Gao,~\IEEEmembership{Member,~IEEE,}
        Chuangji Meng,
        Tao Yang,
        and XinXin Liang.
        
%\thanks{M. Shell was with the Department.}% <-this % stops a space
%\thanks{J. Doe and J. Doe are with Anonymous University.}% <-this % stops a space
% \thanks{
% This work was supported in part by the National Natural Science Foundation of China under Grant 42404122, in part by the China Postdoctoral Science Foundation under Grant 2024M762570, the Fundamental Research Funds for the Central Universities under Grant xzy012025050 and in part by the National Key Research and Development Programs of China under Grant 2020YFA0713403 and Grant 2020YFA0713400.(Jie Chen and Hongling Chen are co-first authors)

% The authors are with the School of Information and Communications Engineering, Xi’an Jiaotong University, Xi’an 710049, China (e-mail: 601882280@qq.com; 859311743@qq.com; jhgao@mail.xjtu.edu.cn; cjmeng@xjtu.edu.cn; yangtao610@xjtu.edu.cn; liangxinxin@stu.xjtu.edu.cn).

% }
% }

\thanks{This work was supported in part by the National Natural Science Foundation of China under Grant 42404122, the China Postdoctoral Science Foundation under Grant 2024M762570, and the Fundamental Research Funds for the Central Universities under Grant xzy012025050.}%
\thanks{Jie Chen and Hongling Chen contributed equally to this work and are co-first authors. (\textit{Corresponding author: Hongling Chen.})}%
\thanks{All authors are with the School of Information and Communications Engineering, Xi’an Jiaotong University, Xi’an 710049, China (e-mail: 601882280@qq.com; 859311743@qq.com; jhgao@mail.xjtu.edu.cn; cjmeng@xjtu.edu.cn; yangtao610@xjtu.edu.cn; liangxinxin@stu.xjtu.edu.cn).}
}

% The paper headers
%\markboth{
%IEEE TRANSACTIONS ON GEOSCIENCE AND REMOTE SENSING}%
%{Shell \MakeLowercase{\textit{et al.}}: Bare Demo of IEEEtran.cls for IEEE Journals}

% Make the title area
\maketitle

\begin{abstract}
Seismic acoustic impedance plays a crucial role in lithological identification and subsurface structure interpretation. However, due to the inherently ill-posed nature of the inversion problem, directly estimating impedance from post-stack seismic data remains highly challenging. Recently, diffusion models have shown great potential in addressing such inverse problems due to their strong prior learning and generative capabilities. Nevertheless, most existing methods operate in the pixel domain and require multiple iterations, limiting their applicability to field data. 
To alleviate these limitations, we propose a novel seismic acoustic impedance inversion framework based on a conditional latent generative diffusion model, where the inversion process is made in latent space. To avoid introducing additional training overhead when embedding conditional inputs, we design a lightweight wavelet-based module into the framework to project seismic data and reuse an encoder trained on impedance to embed low-frequency impedance into the latent space. Furthermore, we propose a model-driven sampling strategy during the inversion process of this framework to enhance accuracy and reduce the number of required diffusion steps.
Numerical experiments on a synthetic model demonstrate that the proposed method achieves high inversion accuracy and strong generalization capability within only a few diffusion steps. Moreover, application to field data reveals enhanced geological detail and higher consistency with well-log measurements, validating the effectiveness and practicality of the proposed approach.
\end{abstract}

%tradition_Gholami2015tv

%tradition_Pereira2017frontier

%tradition_Grana2022probalistic

\begin{IEEEkeywords} 
Impedance inversion, latent diffusion model, deep learning
%波阻抗反演，扩散模型，深度学习
\end{IEEEkeywords}

\section{Introduction}
\label{intro_start}
Seismic acoustic impedance inversion is a critical technique in geophysical exploration. It enables the extraction of impedance from seismic data to identify geological structures, evaluate reservoir potential, and optimize drilling locations \cite{imp_1979acoustic},\cite{imp_latimer2000interpreter}. The challenges associated with solving inversion problems stem from their inherently ill-posed nature, which arises due to factors such as the limited bandwidth of seismic frequencies, noise interference, and the approximation of physical principles during forward modeling. 

Geophysicists have developed a wide range of methodologies to address the above challenges and enhance the accuracy of impedance inversion. These methodologies are broadly classified into traditional inversion methods and deep learning-based inversion methods. 
Traditional seismic impedance inversion methods include deterministic approaches that generally solve a regularization inverse problem, where smoothness, sparsity, or total-variation constraints are imposed to stabilize the solution and enhance resolution.~\cite{tradition_yuan2015sparse,tradition_Gholami2015tv,tradition_Grana2022probalistic}
Another major class of traditional methods is stochastic inversion,
which generates multiple realizations
constrained by seismic data and geological priors, thereby enabling uncertainty
quantification~\cite{tradition_Bosch2010geos,tradition_Azevedo2019multiscale,tradition_Pereira2017frontier}. However, both deterministic and stochastic approaches rely on accurate physical prior knowledge and often struggle to produce high-resolution, noise-robust results for complex field data.
 In contrast, deep learning methods mitigate the dependency on expert-driven prior knowledge inherent in traditional inversion approaches by learning directly from data. Consequently, deep learning methods have emerged as a powerful paradigm for solving inversion problems. A seminal study by Das et al.~\cite{super_Das2018ConvolutionalNN} demonstrated the potential of one-dimensional convolutional neural networks (CNNs) for seismic trace-level impedance inversion, marking a significant milestone in the application of CNNs to this domain. Subsequent research has focused on enhancing model network performance through architectural innovations. For instance, Mustafa et al.~\cite{super_Mustafa2019EstimationOA} introduced temporal convolutional modules to improve sequence modeling capabilities, while Dodda et al.~\cite{super_Dadda2023Attention} incorporated attention mechanisms to optimize feature extraction. However, the success of these supervised learning paradigms depends on the quantity of well logs, which are expensive to acquire in practice. Although semi-supervised learning methods \cite{semi_Chen2022SeismicAI}, \cite{semi_Alfarraj2019semic}, \cite{semi_Chen2024multidimension} leverage both unlabeled and labeled data to reduce the need for extensive labeling, they still require high-quality well logs as labeled data. The misalignment between well logs and seismic records remains a challenging issue to resolve in practice. This reliance on labeled data may arise from the inherent limitations of end-to-end learning frameworks\cite{endtoend_pomerleau1988alvinn}, which require labeled data to learn mapping relationships directly.

With the development of deep learning, diffusion models offer a fundamentally different approach by employing a gradual denoising process to model data distributions. This process captures complex patterns by decomposing the data distribution into a sequence of tractable steps, thereby having the potential to address the limitations in traditional deep learning~\cite{generateTheory_LeCun2006ATO},
and offering improved stability over earlier generative approaches such as GANs and VAEs~\cite{df_Dhariwal2021guide,multimodal_llerFran2023Science}.
The effectiveness of diffusion models has been demonstrated in the visual domain, achieving impressive results in tasks such as image generation~\cite{df_Dhariwal2021guide} and super-resolution \cite{df_Saharia2021SR3}. Moreover, the explicit modeling of the data prior makes them well-suited for inferring missing information in inversion problems. This has led to a growing interest in applying diffusion models to geophysical inversion tasks~\cite{dsApply_zhang2024diffusionvel,dsApply_Meng2025stochastic,dsApply_Wang2024Reconstruction,dsApply_durall2023deep,dsApply_ravasi2025inverseproblemsmeasurementguided}.
Among these studies, Fu~Wang~et~al.~\cite{wang2023tgrs_diffusionfwi} introduced generative diffusion
models into full waveform inversion (FWI) to improve inversion stability and accuracy.
In parallel, Chen~et~al.~\cite{geo_chl2024diffusion} applied diffusion modeling to acoustic
impedance inversion and demonstrated reduced dependence on low-frequency background information.
Subsequent studies have explored different strategies for incorporating geological and geophysical conditions into diffusion-based inversion frameworks.
Zhang~et~al.~\cite{dsApply_zhang2024diffusionvel} adopted a multi-model fusion strategy, where separate conditional diffusion models were trained for seismic data, background velocity, and geological priors, and their outputs were combined during sampling.
Wang~et~al.~\cite{wang2024jgr_controllediffusion} embedded geological labels, well logs, and structural information directly into a single diffusion network during training to achieve controllable velocity generation.
Building on this, Wang~et~al.~\cite{wang2024arxiv_conditionalfwi} extended the conditional formulation to full waveform inversion, integrating geological and well-log priors through classifier-free guidance to enhance data fidelity and geological consistency.
These studies demonstrate distinct paradigms of conditional integration in diffusion-based geophysical modeling.
However, most existing frameworks still operate in the pixel domain, resulting in high computational cost and limited scalability.

In this work, to address this problem, we perform seismic acoustic impedance inversion in latent space instead of pixel space. We begin by projecting the impedance model and conditional information, including seismic data and low-frequency impedance data, into a low-dimensional latent space. They are then combined into a multi-channel input, which is used to train the diffusion model to capture the conditional distribution of impedance in the training dataset. After training, the denoising process serves as the inversion procedure in the latent space, generating impedance from noise using seismic data and low-frequency impedance as inputs. Additionally, purely data-driven learning is often sensitive to distribution shifts between the training and test datasets. We consequently integrate the seismic forward modeling process into the inversion process to enhance inversion robustness. The key contributions of this work can be summarized as follows:
\begin{enumerate}
	\item We propose a seismic acoustic impedance inversion method based on the conditional latent generative diffusion model (SAII-CLDM). This method enables large-scale inversion with improved performance.
	\item We propose a model-driven sampling strategy to improve the inversion accuracy and accelerate the inversion process.
	\item We propose a highly lightweight wavelet-based module that is jointly trained with the diffusion model to incorporate seismic data into the latent generative diffusion model without additional training overhead.
\end{enumerate}

The remainder of this article is organized as follows. Section~\ref{sec:review} provides an overview of the fundamentals of the diffusion model. Section~\ref{sec:method} describes the proposed method in detail, including the proposed inversion framework based on the diffusion model and the proposed model-driven sampling strategy. Sections~\ref{sec:synthetic_test} and~\ref{sec:field_example} validate the proposed method through synthetic data experiments and field data examples, respectively, demonstrating its effectiveness and robustness. In Section~\ref{sec:discussion}, we discuss the advantages and limitations of the proposed method, along with potential improvements. Finally, Section~\ref{sec:conclusion} presents the conclusion of this study.

%
%With the development of deep learning, diffusion models offer a fundamentally different approach by employing a gradual denoising process to model data distributions. This process captures complex patterns by decomposing the data distribution into a sequence of tractable steps, thereby having the potential to address the limitations in traditional deep learning \cite{generateTheory_LeCun2006ATO}. The effectiveness of diffusion models has been demonstrated in the visual domain, achieving impressive results in tasks such as image generation\cite{df_Dhariwal2021guide} and super-resolution\cite{df_Saharia2021SR3}. Moreover, the explicit modeling of the data prior makes them well-suited for inferring missing information in inversion problems. Geophysicists have adopted diffusion models for geophysical inversion problems to achieve higher accuracy\cite{dsApply_zhang2024diffusionvel},\cite{dsApply_Meng2025stochastic},
%\cite{dsApply_Wang2024Reconstruction},
%\cite{dsApply_durall2023deep},
%\cite{dsApply_ravasi2025inverseproblemsmeasurementguided}. Our research team has also conducted studies on unconditional diffusion models for seismic impedance inversion, as detailed in \cite{geo_chl2024diffusion}. However, most existing seismic inversion methods based on diffusion models are predominantly confined to the pixel domain, resulting in long computation time and data size limitations that impede their practical applications. 

\section{Review of Generative Diffusion Model}  
\label{sec:review}
\begin{figure*}[!htbp]
%\centering
\includegraphics[scale=0.22]{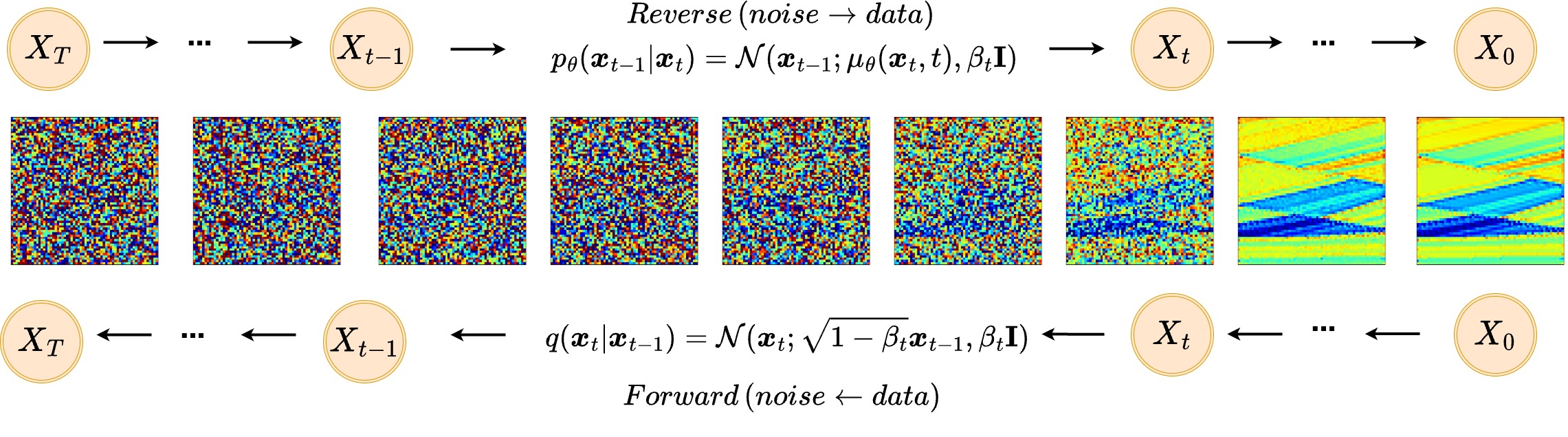}
\caption{An illustration of the forward process and reverse process in the unconditional diffusion model.}
\label{fig:ddpm_process}
\end{figure*}
This section briefly reviews the theoretical foundations of generative diffusion models, and more details are provided in~\cite{df_Ho2020ddpm}. The generative diffusion model offers a principled framework for modelling complex data distributions by progressively corrupting data with low-level noise through a forward diffusion process and then learning to reverse this corruption via a parameterized denoising neural network. A schematic illustration of the forward and reverse process is shown in Fig.~\ref{fig:ddpm_process}.

Specifically, consider the data distribution as $q(\boldsymbol{x}_0)$, where the index 0 indicates that the original data has not been corrupted. The forward process gradually corrupts $\boldsymbol{x}_0$ into pure noise $\boldsymbol{x}_T \sim \mathcal{N}(0, \mathbf{I})$ through a fixed Markov chain with Gaussian transitions:
\begin{equation}
q(\boldsymbol{x}_{1:T}|\boldsymbol{x}_0) = \prod_{t=1}^T q(\boldsymbol{x}_t|\boldsymbol{x}_{t-1})=\prod_{t=1}^T \mathcal{N}(\boldsymbol{x}_t; \sqrt{1-\beta_t}\boldsymbol{x}_{t-1}, \beta_t \mathbf{I}),
\label{eq:forward_process}
\end{equation}
where $\{\beta_t\}_{t=1}^T$ represents a predefined noise schedule that increases with time, and $\mathbf{I}$ denotes the identity matrix. By leveraging the recursive nature of the forward process, the noise sample $\boldsymbol{x}_t$ at an arbitrary step t can be directly derived from $\boldsymbol{x}_0$ as:
\begin{equation}
q(\boldsymbol{x}_t|\boldsymbol{x}_{0})=\mathcal{N}(\boldsymbol{x}_t; \sqrt{\bar{\alpha}_t}\boldsymbol{x}_{0}, (1-\bar{\alpha}_t)\mathbf{I}),
\label{eq:forward_marginal}
\end{equation}
where \(\bar{\alpha}_t = \prod_{s=1}^t (1 - \beta_s)\).

The reverse process attempts to model the original data distribution $q(\boldsymbol{x}_0)$ by 
constructing  a parameterized Markov chain that approximates the true reverse posterior \(q(\boldsymbol{x}_{t-1}|\boldsymbol{x}_t, \boldsymbol{x}_0)\):
\begin{equation}
p_\theta(\boldsymbol{x}_0) = \int p_\theta(\boldsymbol{x}_T) \prod_{t=1}^T p_\theta(\boldsymbol{x}_{t-1}|\boldsymbol{x}_t) \, d\boldsymbol{x}_{1:T}
\label{eq:reverse_process_total}
\end{equation}
\begin{equation}
p_\theta(\boldsymbol{x}_{t-1}|\boldsymbol{x}_t) = \mathcal{N}(\boldsymbol{x}_{t-1}; \mu_\theta(\boldsymbol{x}_t, t), \beta_t \mathbf{I}),
\label{eq:reverse_process}
\end{equation}
where $\mu_\theta(\boldsymbol{x}_t,t) =\frac{1}{\sqrt{\alpha_t}}(\boldsymbol{x}_t - \frac{\beta_t}{\sqrt{1 - \bar{\alpha}_t}}\boldsymbol{\varepsilon}_{\theta}(\boldsymbol{x}_t,t))$, and 
$\boldsymbol{\varepsilon}_{\theta}(\boldsymbol{x}_t,t)$ is the  output of a neural network parameterized by \(\theta\). 

The model parameters $\theta$ are optimized by maximizing the evidence lower bound (ELBO) of the data log-likelihood, defined as:
\begin{equation}
\log p_\theta(\boldsymbol{x}_0) \geq \mathbb{E}_{q(\boldsymbol{x}_{1:T}|\boldsymbol{x}_0)} \left[ \log \frac{p_\theta(\boldsymbol{x}_{0:T})}{q(\boldsymbol{x}_{1:T}|\boldsymbol{x}_0)} \right] \triangleq \mathcal{L}_{\text{ELBO}}.
\label{eq:elbo}
\end{equation}
Furthermore, Ho et al.~\cite{df_Ho2020ddpm} proved that maximizing $\mathcal{L}_{\text{ELBO}}$ can be decomposed into minimizing a series of KL divergences $D_{\mathrm{KL}}(q(\boldsymbol{x}_{t-1}|\boldsymbol{x}_t, \boldsymbol{x}_0) \| p_\theta(\boldsymbol{x}_{t-1}|\boldsymbol{x}_t))$
across all diffusion steps. The decomposition leads to a simplified training objective that reduces to learning a noise prediction network $\boldsymbol{\varepsilon}_\theta$:
\begin{equation}
\mathcal{L} = \min_{\theta}\mathbb{E}_{t, \boldsymbol{x}_0, \boldsymbol{\varepsilon}} \left[ \| \boldsymbol{\varepsilon}_t - \boldsymbol{\varepsilon}_\theta(\boldsymbol{x}_t, t) \|^2 \right],
\label{eq:ddpm_loss}	
\end{equation}
where $\boldsymbol{\varepsilon}_t$ is the noise component of $\boldsymbol{x}_t$. 

%And according to the Eq.~\eqref{eq:forward_process}, the network approximates:\(\boldsymbol{\varepsilon}_{\theta}(\boldsymbol{x}_t) \simeq \frac{\boldsymbol{x}_t - \sqrt{\bar{\alpha}_t}\boldsymbol{x}_0}{\sqrt{1 - \bar{\alpha}_t}}\).

%注意不要修改这里
%----------------------------------------

\begin{figure*}[htbp]
%\centering
\includegraphics[scale=0.24]{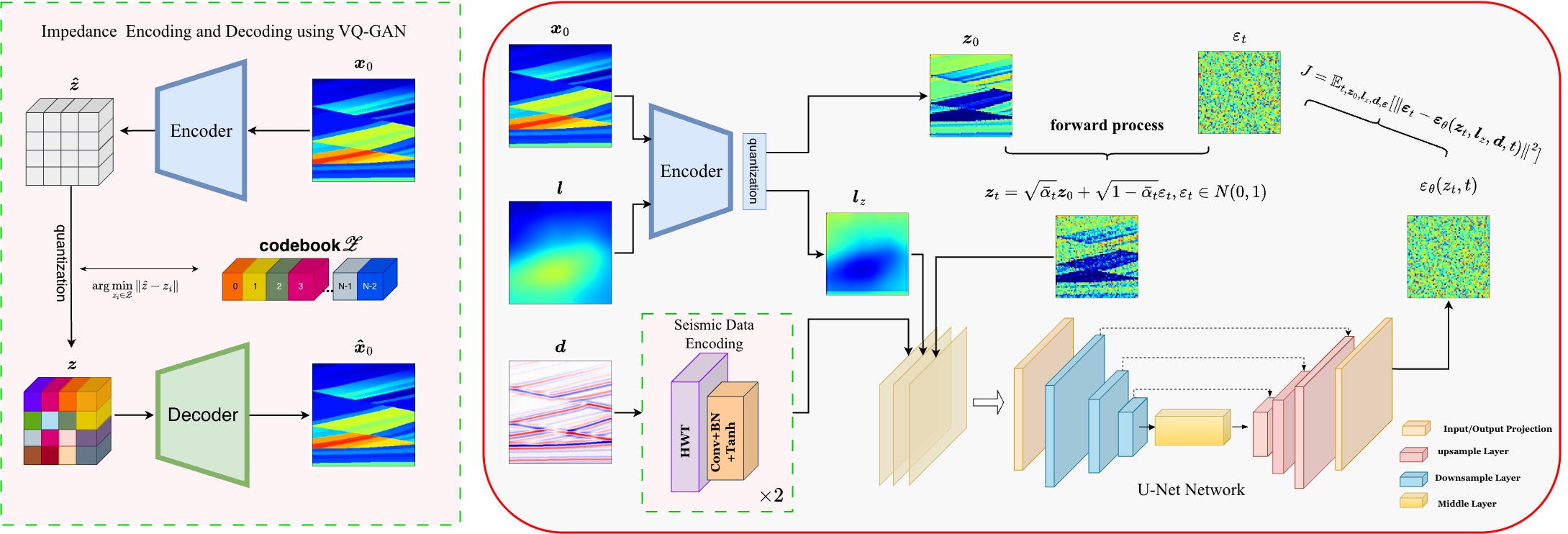}
\caption{Framework of conditional latent diffusion for seismic acoustic impedance inversion.}
\label{fig:jiagou}
\end{figure*}

\section{Method}  
\label{sec:method}

Given the powerful generative capabilities of diffusion models and their ability to learn prior distributions from data, we propose a novel impedance inversion framework based on a latent conditional diffusion model. This section is organized into two parts: First, we present the overall inversion framework. Second, we detail the proposed model-driven sampling strategy. 
\subsection{ Impedance Inversion Framework Based on Conditional Latent Generative Diffusion Model}
\label{sec:condition_ldf}
Seismic acoustic impedance aims to recover subsurface impedance from observed seismic data $\boldsymbol{d}$. Due to the inherently ill-posed nature of this problem, incorporating prior knowledge is essential to alleviate these challenges. Following the findings of Chen et al.~\cite{geo_chl2024diffusion}, integrating low-frequency impedance information $\boldsymbol{l}$ can further enhance inversion accuracy. Consequently, the proposed method utilizes seismic data $\boldsymbol{d}$ and low-frequency impedance $\boldsymbol{l}$ to perform impedance inversion, which can be expressed as the conditional probability $p(\boldsymbol{x} \vert \boldsymbol{l}, \boldsymbol{d})$.

Motivated by the fact that the denoising process can effectively recover the prior distribution, we reformulate the inversion task by modifying the denoising process to recover $p_{\theta}(\boldsymbol{x}_0\vert \boldsymbol{l}, \boldsymbol{d}) \simeq p(\boldsymbol{x}\vert \boldsymbol{l}, \boldsymbol{d})$. Thus, the gradual denoising process is expressed as:
\begin{equation}
p_{\theta}(\boldsymbol{x}_0 \vert \boldsymbol{l}, \boldsymbol{d}) = \int p_{\theta}(\boldsymbol{x}_T) \prod_{t = 1}^{T} p_{\theta}(\boldsymbol{x}_{t - 1}\vert \boldsymbol{x}_t, \boldsymbol{l}, \boldsymbol{d}) \, d\boldsymbol{x}_{1:T}.
\label{eq:inverse_ddpm}
\end{equation}
Here, the conditional distribution \( p_{\theta}(\boldsymbol{x}_{t - 1} \vert \boldsymbol{x}_t, \boldsymbol{l}, \boldsymbol{d}) \) explicitly incorporates \(\boldsymbol{l}\) and \(\boldsymbol{d}\) as additional conditions, thereby distinguishing it from the unconditional counterpart \( p_{\theta}(\boldsymbol{x}_{t - 1} \vert \boldsymbol{x}_t) \) of the standard diffusion model, which is defined in Eq.~\eqref{eq:reverse_process}. There are two primary strategies for introducing such conditions into diffusion models. The first is to train an unconditional diffusion model to obtain prior distribution $q(\boldsymbol{x}_{t - 1} \vert \boldsymbol{x}_t)$ and then modify it into the posterior \( p_{\theta}(\boldsymbol{x}_{t - 1} \vert \boldsymbol{x}_t, \boldsymbol{l}, \boldsymbol{d}) \) using Bayesian inference techniques during sample process \cite{conddf_meng2025posterior},\cite{geo_chl2024diffusion}. Although this strategy is flexible, it typically involves computationally expensive sampling procedures \cite{conddf_chung2022diffusion} and can be challenging to control in complex scenarios. In contrast, the second strategy directly learns the conditional distribution \( p_{\theta}(\boldsymbol{x}_{t-1} \vert \boldsymbol{x}_t, \boldsymbol{l}, \boldsymbol{d}) \) from prior knowledge $q(\boldsymbol{x}_{t-1} \vert \boldsymbol{x}_t, \boldsymbol{l}, \boldsymbol{d})$
\cite{dsApply_Wang2024Reconstruction}. In this work, we adopt the second strategy, as it leads to a more efficient inversion process. 

Furthermore, impedance inversion directly based on the diffusion model incurs substantial computational costs due to the inherent complexity of the diffusion process, rendering it unsuitable for large-scale inversion tasks. To address this issue, we perform impedance inversion in a low-dimensional latent space instead of the original pixel domain, which is inspired by the work of Rombach et al.~\cite{method_Rombach2021ldm}. Then, Eq.~\eqref{eq:inverse_ddpm} is written as:
\begin{equation}
p_{\theta}(\boldsymbol{z}_0 \mid \boldsymbol{l}_z, \boldsymbol{d}_z) = \int p_{\theta}(\boldsymbol{z}_T) \prod_{t = 1}^{T} p_{\theta}(\boldsymbol{z}_{t - 1} \mid \boldsymbol{z}_t, \boldsymbol{l}_z, \boldsymbol{d}_z) d\boldsymbol{z}_{1:T}.
\label{eq:reverse_cldm}
\end{equation}
where $\boldsymbol{z}_0$, $\boldsymbol{l}_z$, and $\boldsymbol{d}_z$ denote the impedance model, low-frequency impedance model, and seismic data in the low-dimensional latent space, respectively. Specifically, we employ a Vector-Quantized Generative Adversarial Network (VQ-GAN), proposed by Esser et al.~\cite{method_Esser2020vqgan}, whose encoder $\mathcal{E}$ can encode seismic acoustic impedance $\boldsymbol{x_0}$ into a low-dimensional latent representation $\boldsymbol{z_0}$, and whose decoder $\mathcal{D}$ can reconstruct $\boldsymbol{z_0}$ back to the acoustic impedance $\boldsymbol{x_0}$, as illustrated in the left panel of Fig.~\ref{fig:jiagou}. The reconstruction 
fidelity of this VQ-GAN is evaluated 
in Appendix~\ref{sec:latent_validation}. Since low-frequency impedance shares the same data domain as impedance, we also use VQ-GAN to project the $\boldsymbol{l}$ into $\boldsymbol{l}_z$. In contrast, seismic data \(\boldsymbol{d}\) reside in a different domain, and we design a wavelet-based module, named SHWT, to transform the seismic data into the low-dimensional space. The above transformations are summarized as follows:
\begin{equation}
	\boldsymbol{z}_0=\mathcal{E}(\boldsymbol{x_0}),
\end{equation}
\begin{equation}
	\boldsymbol{l}_z=\mathcal{E}(\boldsymbol{l}),
\end{equation}
\begin{equation}
	\boldsymbol{d}_z=SHWT_\phi(\boldsymbol{d}).
\end{equation}

It is worth emphasizing that the proposed method does not impose an additional training burden for incorporating condition inputs. For low-frequency impedance, we reuse the VQ-GAN that is trained on impedance data; for seismic data, the SHWT module (parameterized by $\phi$) contains only a few learnable parameters and is jointly trained with the diffusion model. As shown in Fig.~\ref{fig:jiagou}, SHWT integrates the Haar wavelet transform (HWT), convolutional layers with batch normalization (Conv+BN), and a hyperbolic tangent (Tanh) activation function. The Haar wavelet transform is chosen for its ability to provide multi-scale feature extraction and channel expansion, which is superior to naive downsampling in preserving critical information. Furthermore, we observe that noisy seismic data may affect the noise prediction performance of the diffusion model in the low-dimensional space. Therefore, we utilize learnable convolution filters to suppress non-essential components, which is inspired by Xu et al.~\cite{method_Xu2023hwd}. Additionally, by examining the forward modelling process of seismic data, we observe that preserving the polarity of seismic data is crucial, and we employ the tanh activation function. In the forward modeling case of single-trace seismic data, the seismic trace $\boldsymbol{d}$ is expressed as:
\begin{equation}
    \boldsymbol{d} = \omega * \boldsymbol{r} + \boldsymbol{n},
\end{equation}
where $\boldsymbol{\omega}$ is the seismic wavelet, $\boldsymbol{r}$ denotes the reflectivity series, and $\boldsymbol{n}$ represents random noise. The reflectivity $\boldsymbol{r}$ is nonlinearly related to the impedance $\boldsymbol{x}$ through the following equation:
\begin{equation}
    r_{j} = \frac{x_{j+1}-x_{j}}{x_{j+1} + x_{j}}.
    \label{eq:reflectivity_single}
\end{equation}
This equation illustrates that the polarity trends of seismic data contain crucial information about relative impedance variations. 

In summary, the proposed method first projects the impedance $\boldsymbol{x}_0$, low-frequency impedance $\boldsymbol{l}$, and seismic data $\boldsymbol{d}$ into a low-dimensional space. In this space, a conditional diffusion model is trained to learn the distribution $q_(\boldsymbol{z_0} |\boldsymbol{l}_z, \boldsymbol{d}_z)$.
The right panel of Fig.~\ref{fig:jiagou} shows the overall training process. We can see that the encoded impedance $\boldsymbol{z}_0$ is perturbed by  noise $\boldsymbol{\varepsilon_t}$ as:
\begin{equation}
\boldsymbol{z}_t = \sqrt{\bar{\alpha}_t} \boldsymbol{z}_0 + \sqrt{1-\bar{\alpha}_t} \boldsymbol{\varepsilon}_t, \quad \boldsymbol{\varepsilon}_t \sim \mathcal{N}(0, \mathbf{I}).
\label{eq:add_noise}
\end{equation}  
This formulation is based on Tweedie's formula, which is used to derive the posterior mean under the Gaussian assumption as defined in Eq.~\eqref{eq:forward_marginal}. Subsequently, $\boldsymbol{z}_t$ is concatenated with the condition inputs $\boldsymbol{l}_z$ and $\boldsymbol{d}_z$ to form multi-channel inputs to the neural network. The loss function is defined as:
\begin{equation}
	J = \mathbb{E}_{t, \boldsymbol{z}_0, \boldsymbol{l}_z, \boldsymbol{d}, \boldsymbol{\varepsilon_t}}[\|\boldsymbol{\varepsilon}_t - \boldsymbol{\varepsilon}_{\theta+\phi}(\boldsymbol{z}_t, t, \boldsymbol{l}_z, \boldsymbol{d})\|^2]
\label{eq:cldm_loss}
\end{equation}
Although the diffusion model is conditioned on auxiliary condition inputs $\boldsymbol{l}$ and $\boldsymbol{d}$, the loss function Eq.~\eqref{eq:cldm_loss} remains consistent with Eq.~\eqref{eq:ddpm_loss}, and we will provide more proof in the Appendix~\ref{appen:condition_ddpm_format}. 

After training, the proposed inversion framework accepts seismic data $\boldsymbol{d}$ and low-frequency impedance $\boldsymbol{l}$ as conditional inputs. It then generates the impedance from Gaussian noise via a standard denoising process, similar to the reverse process depicted in Fig.~\ref{fig:ddpm_process}. However, the standard denoising process----i.e., the Denoising Diffusion Probabilistic Model (DDPM)--- requires a long runtime due to the large number of timesteps. To address this drawback, we adopt the diffusion denoising implicit model (DDIM) introduced by Song et al.\cite{method_Song2020ddim} to reduce the sampling timesteps. The generation of a sample from $\boldsymbol{z}_{t+1}$ to $\boldsymbol{z}_{t}$ is given by:
\begin{multline}
\boldsymbol{z}_{t}=\sqrt{\bar{\alpha}_{t}}\left(\frac{\boldsymbol{z}_{t+1}-\sqrt{1-\bar{\alpha}_{t+1}} \boldsymbol{\varepsilon}_{\theta + \phi}(\boldsymbol{z}_{t+1},t+1,\boldsymbol{l}_z, \boldsymbol{d})}{\sqrt{\bar{\alpha}_{t+1}}}\right) \\
+ \sqrt{1-\bar{\alpha}_{t}-\sigma_{t+1}^{2}}\boldsymbol{\varepsilon}_{\theta + \phi}(\boldsymbol{z}_{t+1},t+1,\boldsymbol{l}_z, \boldsymbol{d})  + \sigma_{t+1} \boldsymbol{\varsigma}, \boldsymbol{\varsigma} \sim \mathcal{N}(0, \boldsymbol{I}),
\label{eq:ddim_sample}
\end{multline}
where $\sigma_{t+1}=\eta \sqrt{\left(1-\bar{\alpha}_{t}\right) /\left(1-\bar{\alpha}_{t+1}\right)} \sqrt{\left(1-\bar{\alpha}_{t+1} / \bar{\alpha}_{t}\right)}$ and $\eta$ is a hyperparameter that controls the stochasticity of the sampling process. 
Furthermore, to improve inversion accuracy, we make improvements in the sampling process, which will be detailed in Section~\ref{sec:model-driven_sample_strategy}.

%即使上个章节的我们提出的反演框架已经可以产生不错的结果了，但是当训练数据集和测试集存在偏差时不可避免精度都会下降。因此，我们提出使用模型驱动来矫正为扩散去噪的反演过程。具体而言，我们使用a提出的stochsticresample的方法，建立一个新的后验分布，该分布融合融合了$z_t$和$\boldsymbol{z}_0'(\boldsymbol{d})$的信息，且保证了从该分布采样出的$z_t'$与 $z_t$不会有过大的方差，从而使得这种修正不会破坏马尔可夫链。
\subsection{Model-Driven Sampling of SAII-CLDM}
\label{sec:model-driven_sample_strategy}

Although the framework proposed in Section~\ref{sec:condition_ldf} can produce favorable inversion results, the accuracy inevitably degrades due to distribution shifts between the training and test datasets. To alleviate this limitation, we introduce a model-driven sampling strategy that complements the data-driven learning pattern, significantly enhancing inversion robustness. Notably, experimental results indicate that this strategy also contributes to reduced sampling time. While the DDIM sampling method enables reducing sampling steps, it often suffers from performance degradation at a few timesteps. In contrast, integrating DDIM with the proposed model-driven strategy enables accurate inversion with fewer sampling timesteps, outperforming the standard DDPM-based approach\cite{df_Ho2020ddpm}. Experimental validation of this claim is provided in Section~\ref{sec:synthetic_test}.

The model-driven sampling strategy is implemented by utilizing model-driven optimization to indirectly correct $\boldsymbol{z}_t$, which is sampled as described in Eq.~\eqref{eq:ddim_sample}. 
Specifically, the model-driven optimization, which can be flexibly controlled through the learning rate and the number of iterations, is formulated as follows:
\begin{equation}
\boldsymbol{z}_0'(\mathcal{D}(\hat{\boldsymbol{z}}_0), \boldsymbol{d}) = \underset{\mathcal{D}({\hat{\boldsymbol{z}}_0})}{\arg\min} \|\boldsymbol{d}-f(\mathcal{D}(\hat{\boldsymbol{z}}_0))\|_{2}^{2},
\label{eq:model_driven_optimization}
\end{equation}
where $\boldsymbol{d}$ denotes the observed seismic data and $f(\cdot)$ represents the forward-modeling function given by:
\begin{equation}
    f(\boldsymbol{x}) = \boldsymbol{W} \mathcal{C}(\boldsymbol{x}),
    \label{eq:synthetic_record_noise}
\end{equation}
Here, $\boldsymbol{x} \in \mathbb{R}^{m\times n}$ represents the impedance model, consisting of $n$ traces each of length $m$. The operator $\mathcal{C}(\cdot)$ computes the reflectivity trace-by-trace using Eq.~\eqref{eq:reflectivity_single}, while $\boldsymbol{W}$ is a Toeplitz matrix constructed from the seismic wavelet $\omega$. Note that in Eq.~\eqref{eq:model_driven_optimization}, $f(\cdot)$ is applied to $\mathcal{D}(\hat{\boldsymbol{z}}_0)$, which decodes the latent $\hat{\boldsymbol{z}}_0$ into an impedance model in the pixel domain. To incorporate the model-driven optimization into the inversion process, we use $\hat{\boldsymbol{z}}_0(\boldsymbol{z}_t)$ as the initial value, which is derived from Eq.~\eqref{eq:add_noise} as:
\begin{equation}
\hat{\boldsymbol{z}}_0=\frac{1}{\sqrt{\bar{\alpha}_t}}\left(\boldsymbol{z}_t - \sqrt{1 - \bar{\alpha}_t}\boldsymbol{\varepsilon}_{\theta+\phi}(\boldsymbol{z}_t, t, \boldsymbol{l}_z, \boldsymbol{d})\right).
\label{eq:predicet_z0}
\end{equation}
This formulation projects $\boldsymbol{z}_t$ into the domain where the latent impedance model $\boldsymbol{z}_0$ resides, enabling the model-driven optimization Eq.~\eqref{eq:model_driven_optimization} to produce $\boldsymbol{z}_0'(\boldsymbol{z}_t)$. Subsequently, we employ the StochasticResample approach, introduced by Song et al.~\cite{method_song2024resample}, to project $\boldsymbol{z}_0'$ back into the data domain defined by $\boldsymbol{z}_t$ without disrupting the reverse diffusion transition. This approach constructs a new posterior distribution $\widetilde{p}(\boldsymbol{z}_t'|\boldsymbol{z}_t, \boldsymbol{z}_0'(\boldsymbol{d}))$, which ensures that the variance between the resampled result \(\boldsymbol{z}_t'\) and \(\boldsymbol{z}_t\) remains controlled, thereby preserving the reverse diffusion transition. The posterior $\widetilde{p}(\boldsymbol{z}_t'|\boldsymbol{z}_t, \boldsymbol{z}_0'(\boldsymbol{d}))$ is defined as:
\begin{equation}
\mathcal{N} \left( \frac{\kappa_t^2\sqrt{\bar{\alpha}_t}\hat{\boldsymbol{z}}_0'(\boldsymbol{d}) + (1-\bar{\alpha}_t)\boldsymbol{z}_t}{\kappa_t^2 + (1-\bar{\alpha}_t)}, \frac{\kappa_t^2(1-\bar{\alpha}_t)}{\kappa_t^2 + (1-\bar{\alpha}_t)}\boldsymbol{I} \right),
\label{eq:stochatic}
\end{equation}
where $\kappa_{t}^{2}=\gamma\left(\frac{1-\bar{\alpha}_{t-1}}{\bar{\alpha}_{t}}\right)\left(1-\frac{\bar{\alpha}_{t}}{\bar{\alpha}_{t-1}}\right)$, $\gamma$ is a hyperparameter that regulates the trade-off between prior consistency and optimization constraints. By applying Tweedie's formula\cite{df_efron2011tweedie}, samples from this distribution are expressed as:
\begin{multline}
\hat{\boldsymbol{z}}_t= \frac{\kappa_t^2\sqrt{\bar{\alpha}_t}\hat{\boldsymbol{z}}_0'(\boldsymbol{d}) + (1-\bar{\alpha}_t)\boldsymbol{z}_t}{\kappa_t^2 + (1-\bar{\alpha}_t)}+ \sqrt{\frac{\kappa_t^2(1-\bar{\alpha}_t)}{\kappa_t^2 + (1-\bar{\alpha}_t)}}\boldsymbol{\varsigma}, \\
\boldsymbol{\varsigma}\sim \mathcal{N}(0, \mathbf{I})
\label{eq:tweedie_stochastic}
\end{multline}

Besides, as illustrated in the complete sampling process shown in Algorithm~\ref{algc:sample}, the proposed strategy is executed at regular intervals across timesteps following DDIM sampling, rather than at every step. This is attributed to the fact that adjacent samples inherently retain similar semantic structural information \cite{df_yu2023freedom}. Consequently, the proposed strategy operates effectively without introducing significant computational overhead.

\begin{algorithm*}
\caption{Model-Driven Sampling of SAII-CLDM}
\label{alg:conditional_sampling}
\textbf{Input:} Low-frequency impedance $\boldsymbol{l}$, seismic data $\boldsymbol{d}$ \\
\textbf{Hyperparameters:} Sample timesteps $T$
\begin{algorithmic}[1]
\Require $f(\cdot)$, Encoder $\mathcal{E}(\cdot)$, Decoder $\mathcal{D}(\cdot)$, and score model $\varepsilon_{\theta}(\cdot)$.
\State $\boldsymbol{l}_z \gets \mathcal{E}(\boldsymbol{l})$ \Comment{Encode low-frequency impedance}
\State $\boldsymbol{z}_T \sim \mathcal{N}(0, \boldsymbol{I})$,  \Comment{Initialize from noise}
\For{$t = T-1, \ldots, 0$}
    \State $\hat{\boldsymbol{\varepsilon}}_{t+1} \gets \boldsymbol{\varepsilon}_{\theta+\phi}(\boldsymbol{z}_{t+1}, t+1, \boldsymbol{l}_z, \boldsymbol{d})$ \Comment{The neural network trained by Eq.~\eqref{eq:cldm_loss}})
    \State $\boldsymbol{z}_{t} \gets \sqrt{\bar{\alpha}_{t}}\left(\frac{\boldsymbol{z}_{t+1}-\sqrt{1-\bar{\alpha}_{t+1}} \boldsymbol{\hat{\varepsilon}}_{t+1}}{\sqrt{\bar{\alpha}_{t+1}}}\right)+\sqrt{1-\bar{\alpha}_{t}-\sigma_{t+1}^{2}} \hat{\boldsymbol{\varepsilon}}_{t+1} + \sigma_{t+1} \boldsymbol{\varsigma}, \boldsymbol{\varsigma} \sim \mathcal{N}(0, \boldsymbol{I})$
    \Comment{DDIM step described in Eq.~\eqref{eq:ddim_sample}}
    \If{$t \equiv 0 \pmod{\text{interval}}$} \Comment{Apply model-driven strategy every $\text{interval}$ steps}
        \State $\hat{\boldsymbol{z}}_0(\boldsymbol{z}_{t+1}) \gets \frac{1}{\sqrt{\bar{\alpha}_{t+1}}}(\boldsymbol{z}_{t+1}-\sqrt{1-\bar{\alpha}_{t+1}}\hat{\boldsymbol{\varepsilon}}_{t+1})$ \Comment{Compute initial estimate as in Eq.~\eqref{eq:predicet_z0}}
        \State $\hat{\boldsymbol{x}}_0' \gets \underset{\mathcal{D}({\hat{\boldsymbol{z}}_0})}{\arg\min} \|\boldsymbol{d}-f(\mathcal{D}(\hat{\boldsymbol{z}}_0))\|_{2}^{2},$ \Comment{described in Eq.~\eqref{eq:model_driven_optimization}}
        \State $\hat{\boldsymbol{z}}_0' \gets \mathcal{E}({\hat{\boldsymbol{x}}_0'})$ 
%        \Comment{Solve $\boldsymbol{z}_0'$ via data fidelity optimization}
        \State $\boldsymbol{z}_t' \gets \text{StochasticResample}(\hat{\boldsymbol{z}}_0', \boldsymbol{z}_t)$\Comment{described in Eq.~\eqref{eq:tweedie_stochastic}}
        \State $\boldsymbol{z}_t \gets \boldsymbol{z}_t'$ 
%        \Comment{Transform $\widetilde{p}(\boldsymbol{z}_t)$ to $\widetilde{p}(\boldsymbol{z}_t'|\boldsymbol{z}_t, \boldsymbol{z}_0'(\boldsymbol{y}))$}
    \EndIf
\EndFor
\State $\boldsymbol{x} \gets \mathcal{D}(\boldsymbol{z}_0)$ \Comment{Output reconstructed impedance data}
\end{algorithmic}
\label{algc:sample}
\end{algorithm*}

 \section{Synthetic Data Numerical Example}
 \label{sec:synthetic_test}
This section presents the experimental evaluation conducted on synthetic data. We first describe the training and testing configurations, followed by a detailed analysis of the experimental results. The goal is to validate the effectiveness and robustness of the proposed method under controlled conditions.

\subsection{Training Details and Test Details}
\subsubsection{Training Datasets}
\label{sec:training_dataset}

 \begin{figure}[htbp]
  \centering
  \includegraphics[width=0.5\textwidth]{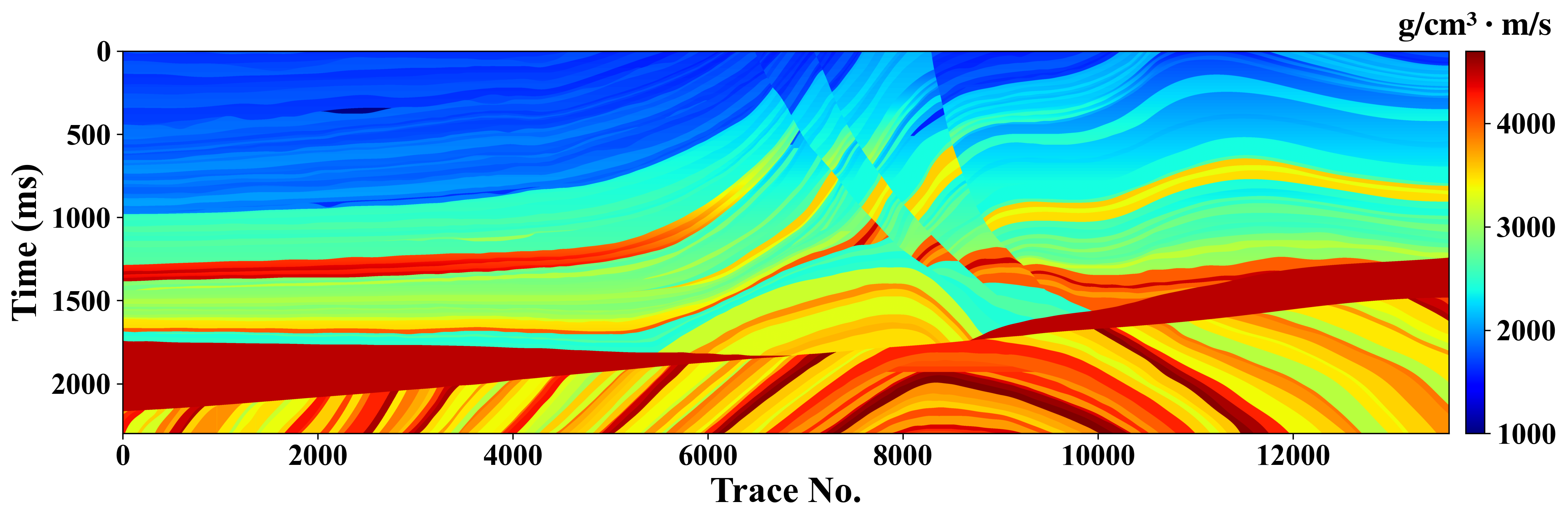}
  \caption{ Marmousi model.}
  \label{fig:Marmousi_model}
\end{figure}

 We adopt the Marmousi II impedance model (Fig.~\ref{fig:Marmousi_model}) \cite{marmousi_Martin2006Marmousi2AE} to generate the training data due to its complex structures. 
First, we divide the model into $256\times256$ patches, yielding 313 distinct acoustic impedance models. 
To enhance the diversity of the dataset, we then apply augmentation techniques, including vertical and horizontal flipping and elastic deformation, expanding the training dataset to 1,565 impedance models. 
Seismic data are synthesized through Eq.~\eqref{eq:synthetic_record_noise}, where $\boldsymbol{W}$ is constructed using Ricker wavelets with dominant frequencies randomly selected from (25~Hz, 30~Hz, 35~Hz). 
Subsequently, 20~dB band-pass coherent noise is added. 
Specifically, the synthetic noise is produced by applying a two-pass three-point moving-average filter to standard Gaussian noise, which introduces temporal correlation and a band-limited structure.
The low-frequency impedance conditions are generated by applying a low-pass filter with random cutoff frequencies of 3~Hz, 6~Hz, 12~Hz, and 18~Hz.

\subsubsection{Training Procedure}
\label{sec:training_paramerter_synthetic}
The model is trained on an NVIDIA RTX 3090 GPU. Initially, we train the VQ-GAN using impedance data from the Marmousi model, following the methodology described in \cite{method_Esser2020vqgan}. The training process takes approximately 500 epochs, with an initial learning rate of $4.5 \times 10^{-6}$ and a batch size of 5. 
Subsequently, the trained VQ-GAN encodes both the impedance and low-frequency impedance into latent representations, compressing the spatial resolution by a factor of 4. 
 In parallel, the seismic data are processed by a lightweight SHWT module that transforms them into latent features; this module is jointly optimized with the DDPM. 
We then train the DDPM model for approximately 500 epochs, using these encoded representations with a learning rate of $4.5 \times 10^{-6}$ and a batch size of 24. 
A linear noise schedule with 1000 timesteps is employed for DDPM. 
The hyperparameter configuration was selected with reference to commonly used settings reported in \cite{method_Rombach2021ldm} to ensure stable convergence, 
and the number of training epochs was determined based on the convergence behavior of the loss curves.

\subsubsection{Test Details}
\label{sec:compare_method}
To underscore the advantage of the proposed SAII-CLDM, we compare it with three classical benchmark methods: a supervised deep learning method (SDL)%\cite{super_Das2018ConvolutionalNN}
, an unsupervised deep learning method (USDL)
%~\cite{unsup_Feng2020AnUD}
, and the Total Variation inversion method (2D-TV)~\cite{tv_Ravasi2021pd}. These methods differ in their learning mechanisms compared to SAII-CLDM. The SDL method employs a U-Net architecture to directly learn the mapping between seismic data and impedance models, whereas SAII-CLDM learns data distributions. Both methods use the same training dataset to train the deep neural model.

In contrast, USDL does not use the training dataset~\cite{geo_chl2024diffusion}. It uses only the test dataset (without a true impedance model) for training with a prior constraint. The corresponding loss function is as follows:
\begin{equation}
J = \|\boldsymbol{d}-f(T_{\phi}(\boldsymbol{d}))\|_2^2 + \mu_1\|T_{\phi}(\boldsymbol{d})-\boldsymbol{l}\|_2^2 + \mu_2\|T_{\phi}(\boldsymbol{d})\|_{tv},	
\label{eq:unsup_loss}
\end{equation}
where $T_{\phi}$ denotes the U-Net with parameters $\phi$, $\boldsymbol{l}$ represents the low-frequency impedance, $\|\cdot\|_{tv}$ is the 2D total variation constraint, $f(\cdot)$ is forward model described in Eq.~\eqref{eq:synthetic_record_noise}, $\mu_1$ and $\mu_2$ are trade-off parameters. The SDL and USDL baselines are implemented following the same baseline configurations
adopted in Chen et al.~\cite{geo_chl2024diffusion}.

Unlike the other three methods, the 2D-TV method is a non-learning method. It adopts a similar loss function to that in Eq.~\eqref{eq:unsup_loss}. Here, we employ the Primal-Dual algorithm proposed by Ravasi et al. ~\cite{tv_Ravasi2021pd} as a specific implementation for comparison.

Additionally, to validate the effectiveness of our proposed model-driven sampling strategy, we compared it with a baseline that employs a 1000-timestep standard denoising process using the same trained model. Since the training process also uses 1000 timesteps, this ensures a fair comparison between the two sampling methods. As the standard diffusion method is based on the DDPM approach proposed by Ho et al.~\cite{df_Ho2020ddpm}, we refer to this baseline as SAII-LDDPM in the following text.

\begin{figure*}[htbp]
  \centering
  \subfloat[\label{fig:model_overthrust_speed}]{\includegraphics[width = 0.23\textwidth]{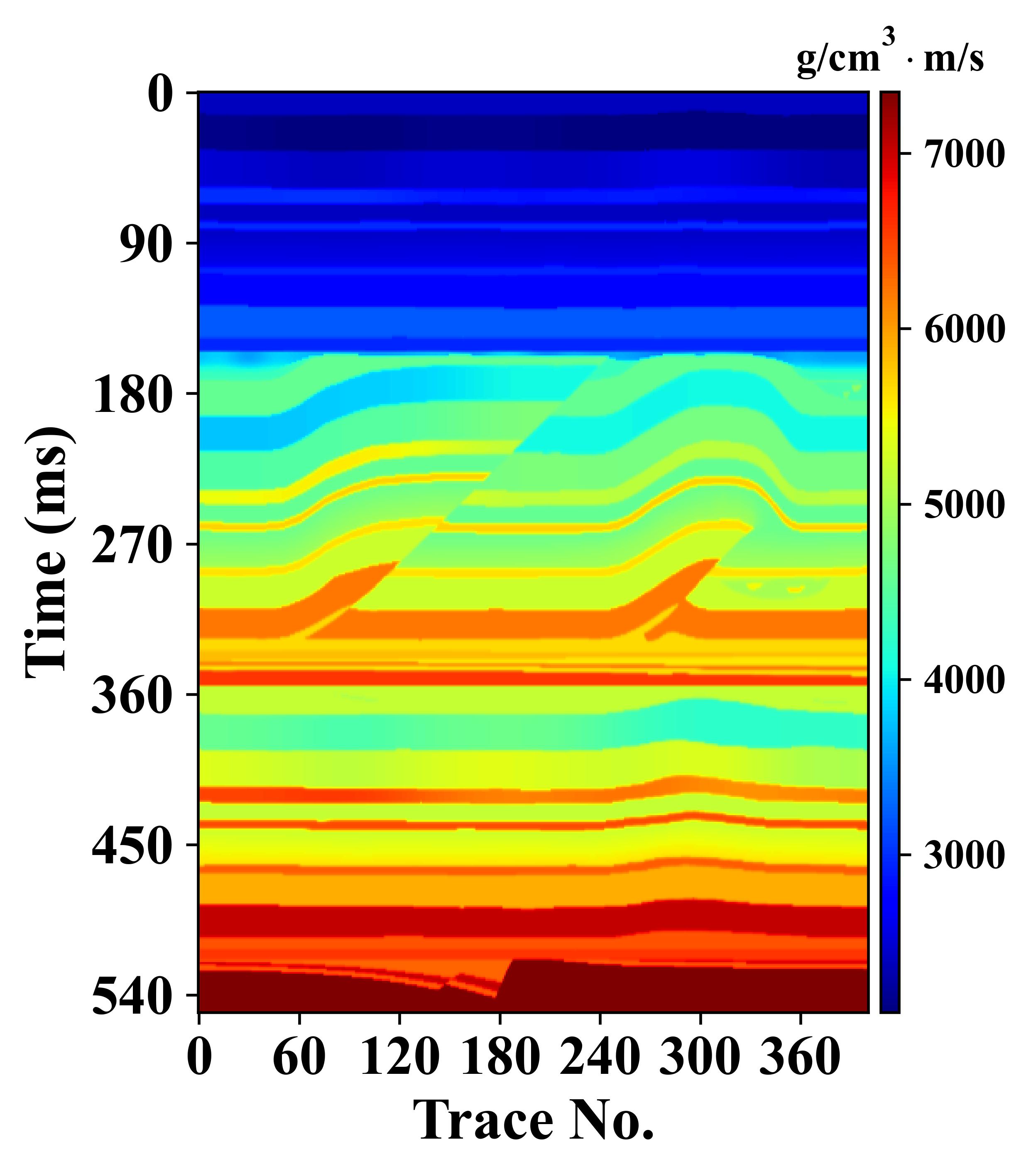}}
  \subfloat[\label{fig:model_overthrust_dipin}]{\includegraphics[width = 0.23\textwidth]{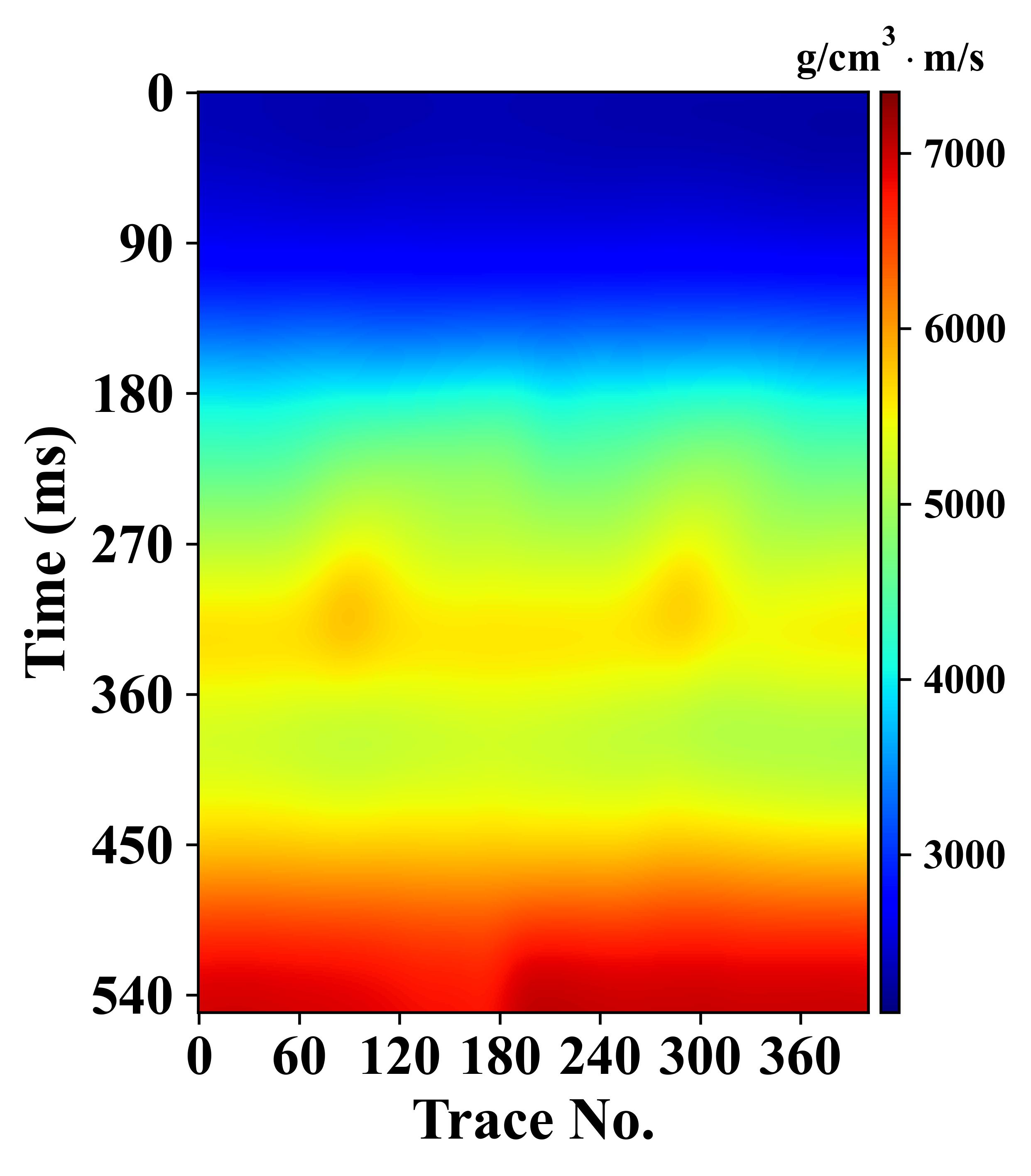}}\quad
  \subfloat[\label{fig:model_overthrust_clearRecord}]{\includegraphics[width = 0.23\textwidth]{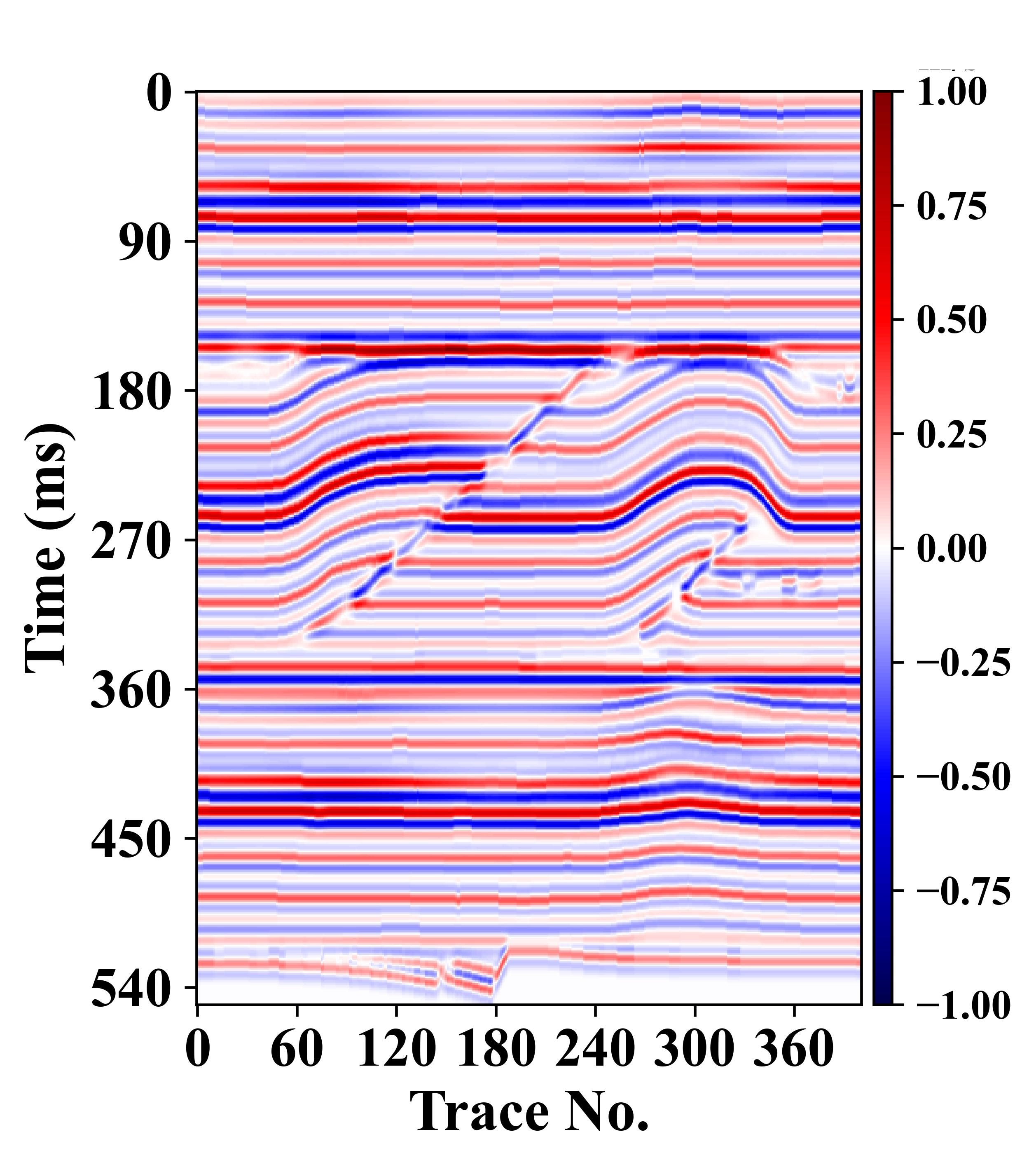}}\quad
  \subfloat[\label{fig:model_overthrust_noisyRecord}]{\includegraphics[width = 0.23\textwidth]{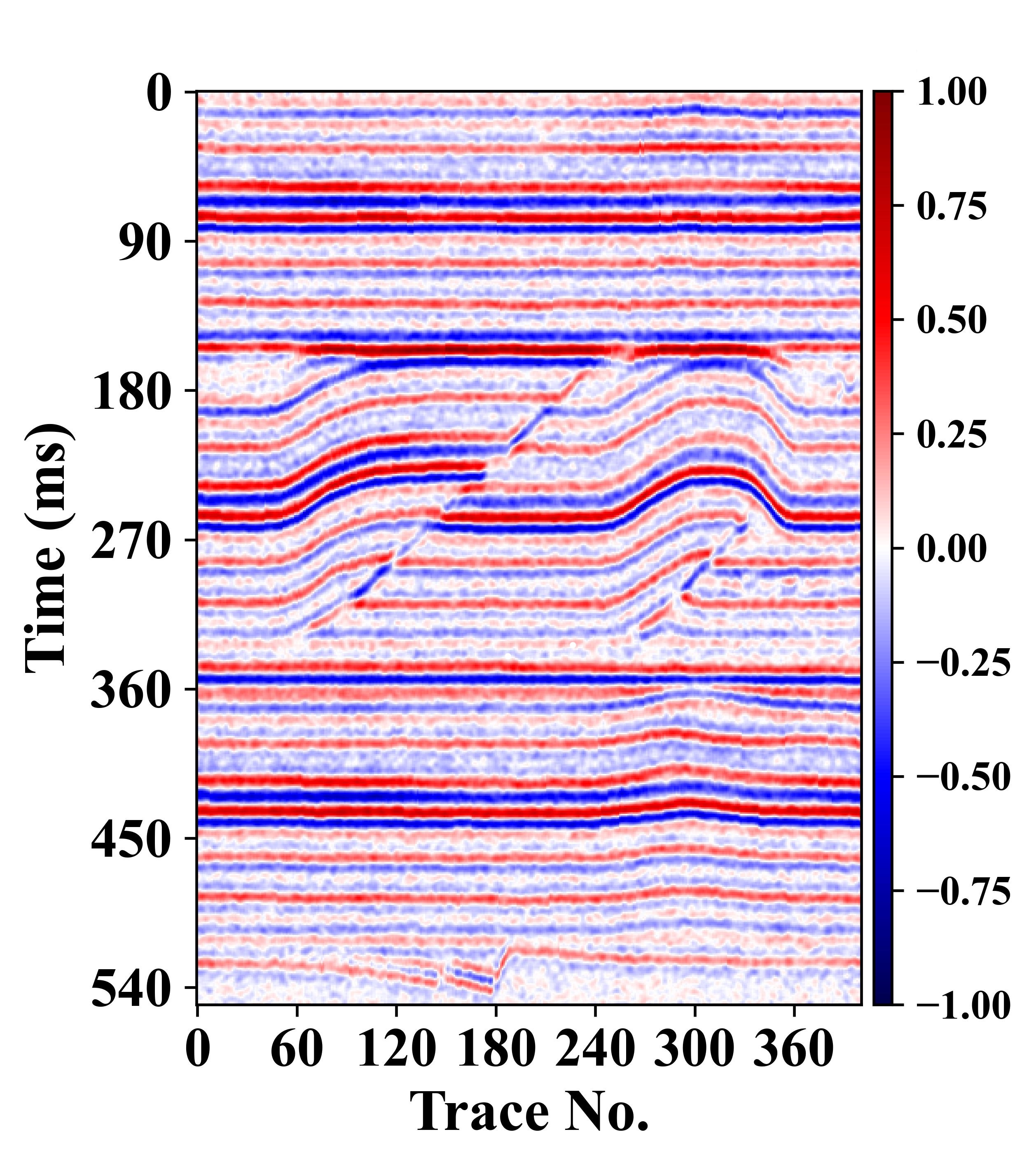}}
  \caption{Overthrust model. (a) accurate impedance, (b) low-frequency impedance, (c) free-noise synthetic seismic data, and (d) synthetic seismic data with 15~dB band-pass coherent noise.}
  \label{fig:model_overthrust}
\end{figure*}

\begin{figure*}[htbp]
  \centering
    \subfloat[\label{fig:imp_cldm_0}]{\includegraphics[width = 0.18\textwidth]{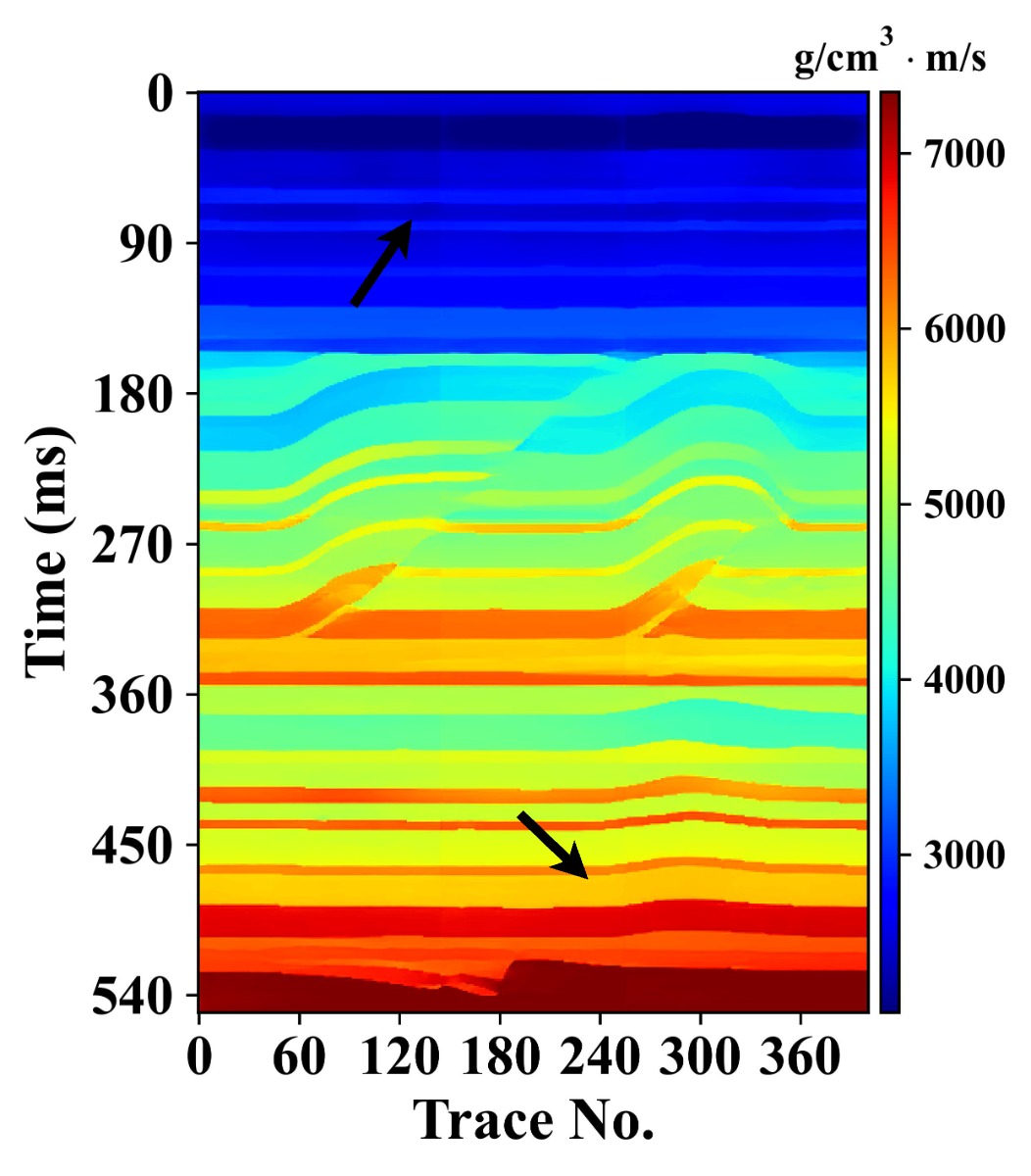}}\quad
      \subfloat[\label{fig:imp_ddpm_0}]{\includegraphics[width = 0.18\textwidth]{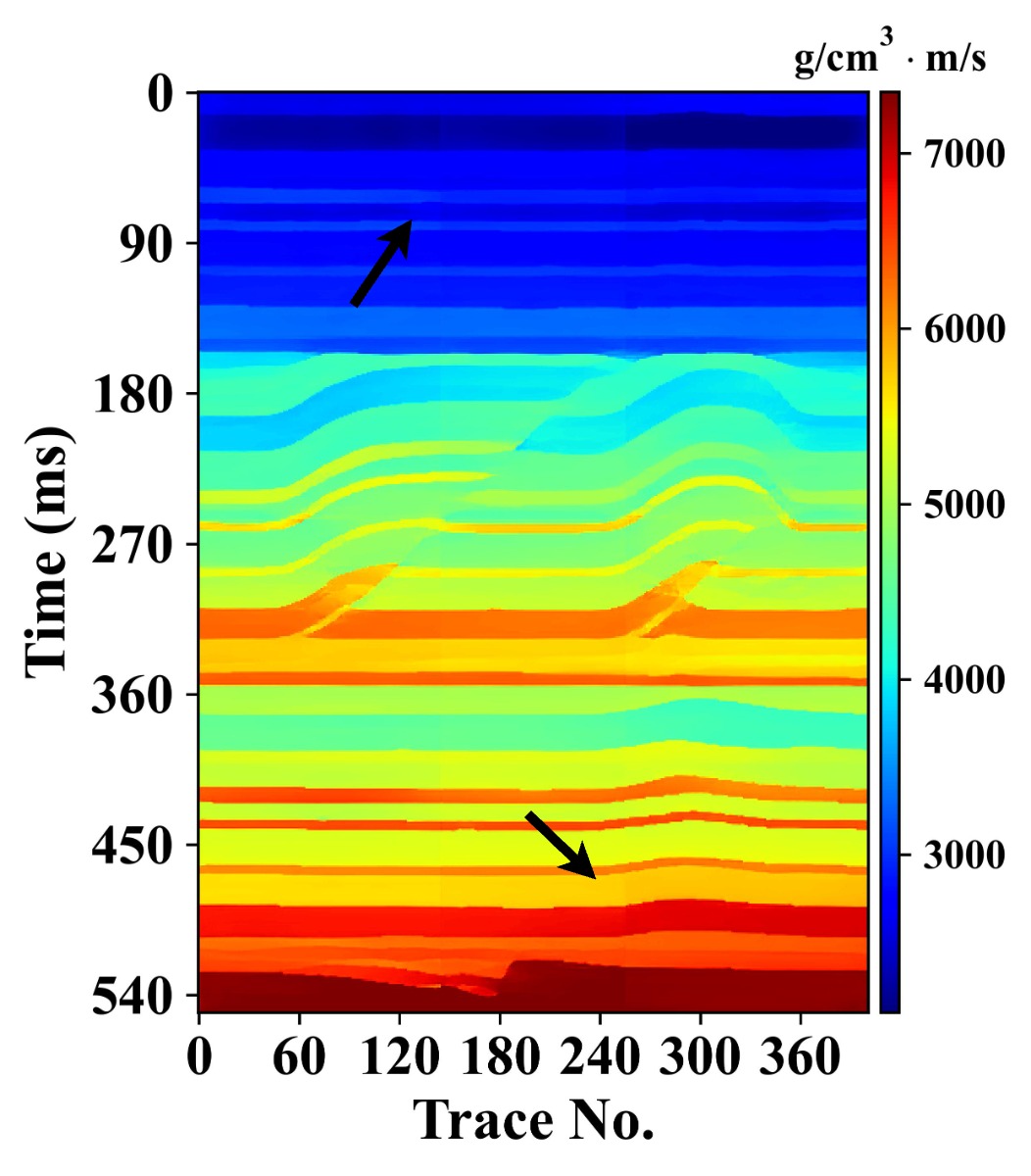}} \quad
        \subfloat[\label{fig:imp_sup_0}]{\includegraphics[width = 0.18\textwidth]{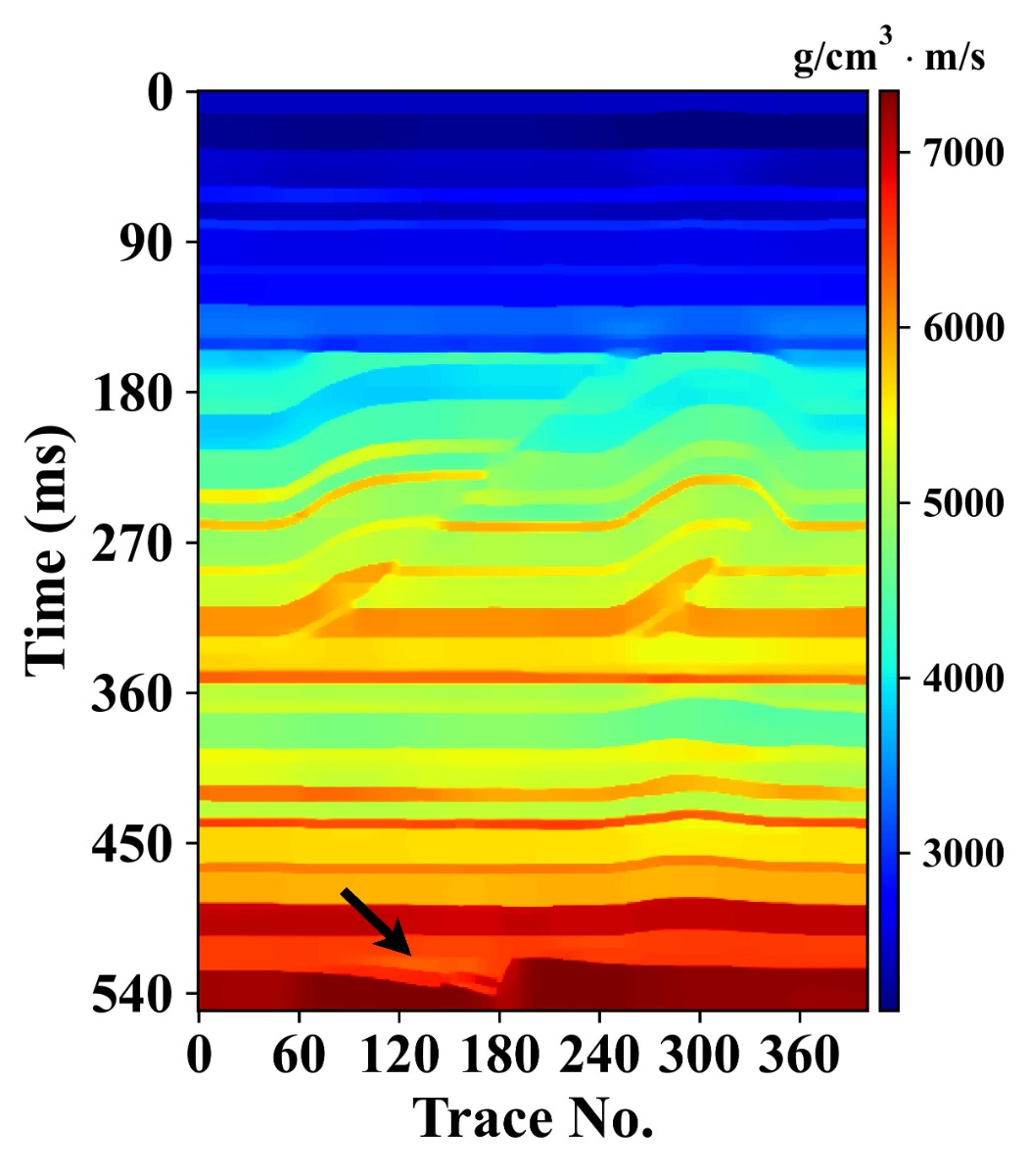}}\quad
  \subfloat[\label{fig:imp_unsup_0}]{\includegraphics[width = 0.18\textwidth]{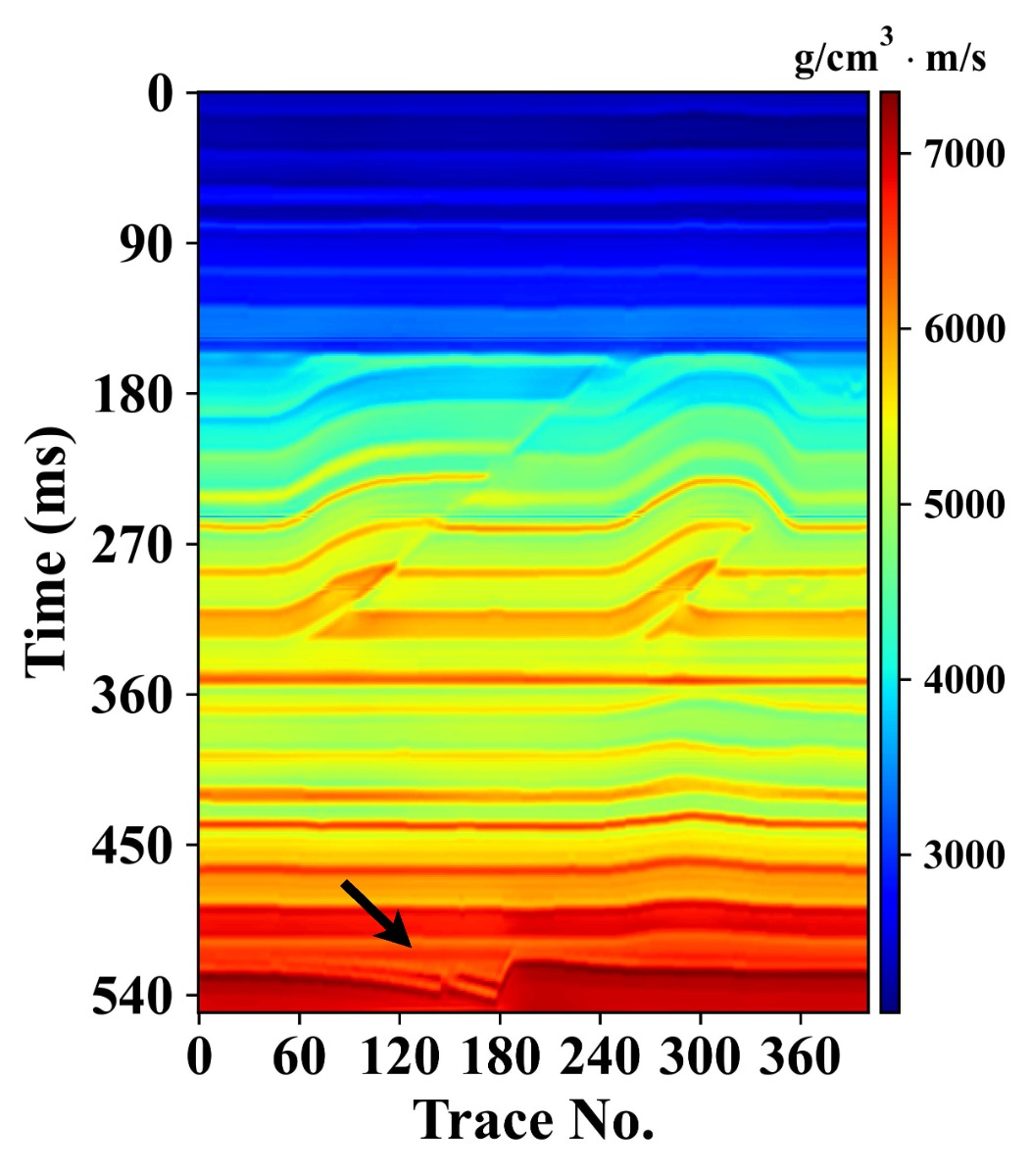}}\quad
  \subfloat[\label{fig:imp_tv_0}]{\includegraphics[width = 0.18\textwidth]{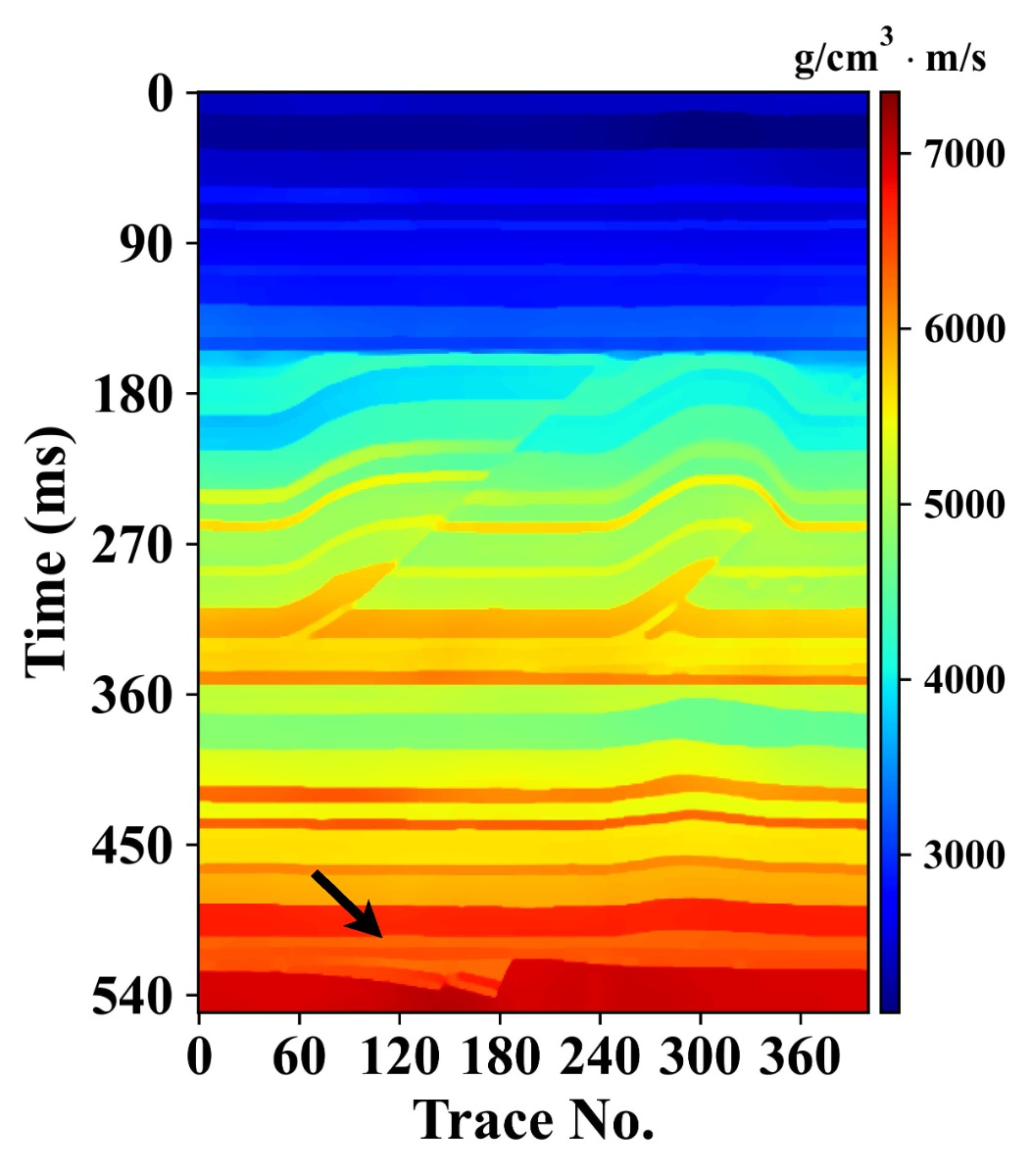}} \\
    \subfloat[\label{fig:imp_cha_cldm_0}]{\includegraphics[width = 0.18\textwidth]{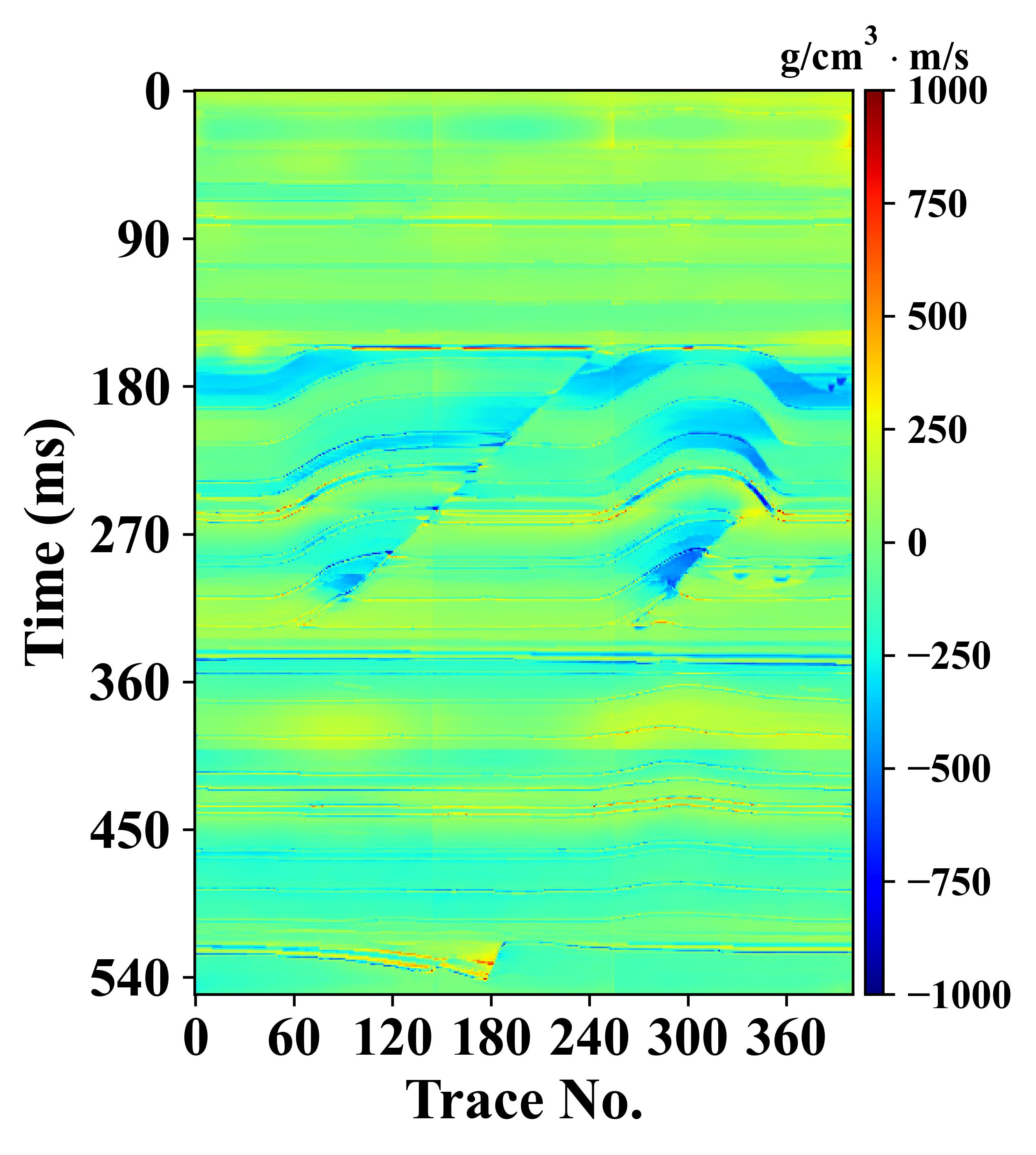}}\quad
  \subfloat[\label{fig:imp_cha_ddpm_0}]{\includegraphics[width = 0.18\textwidth]{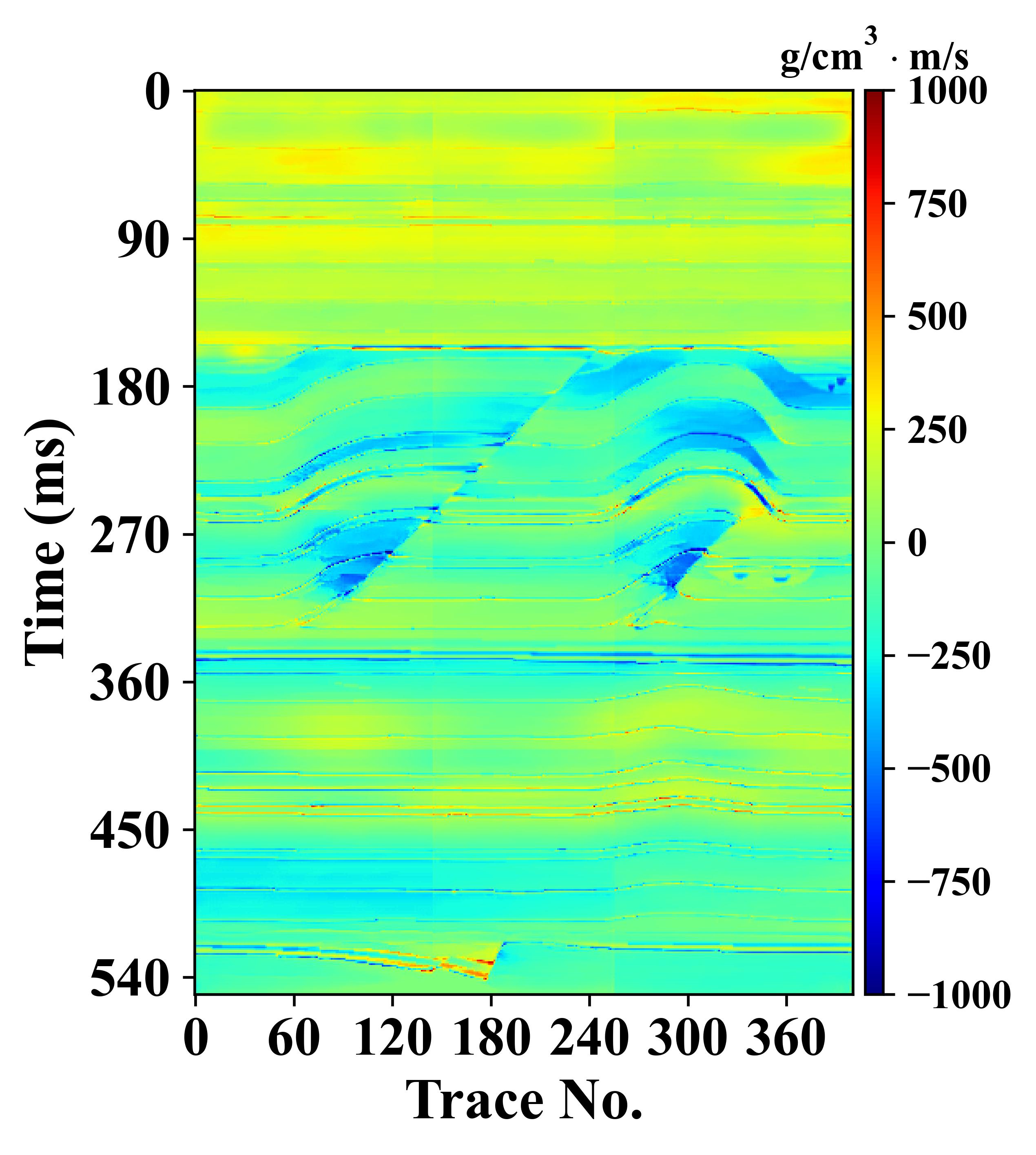}}\quad
   \subfloat[\label{fig:imp_cha_sup_0}]{\includegraphics[width = 0.18\textwidth]{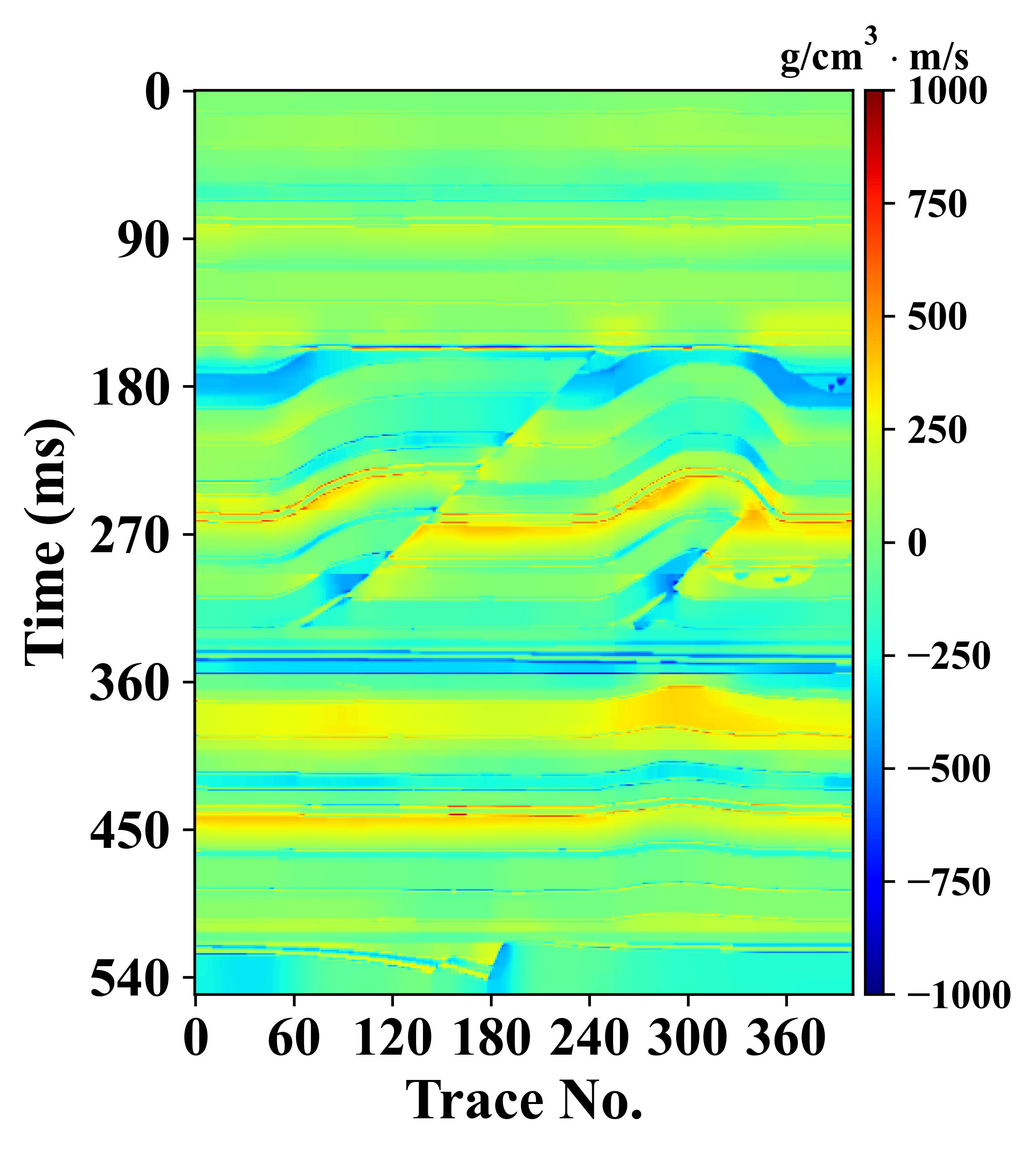}}\quad
     \subfloat[\label{fig:imp_cha_unsup_0}]{\includegraphics[width = 0.18\textwidth]{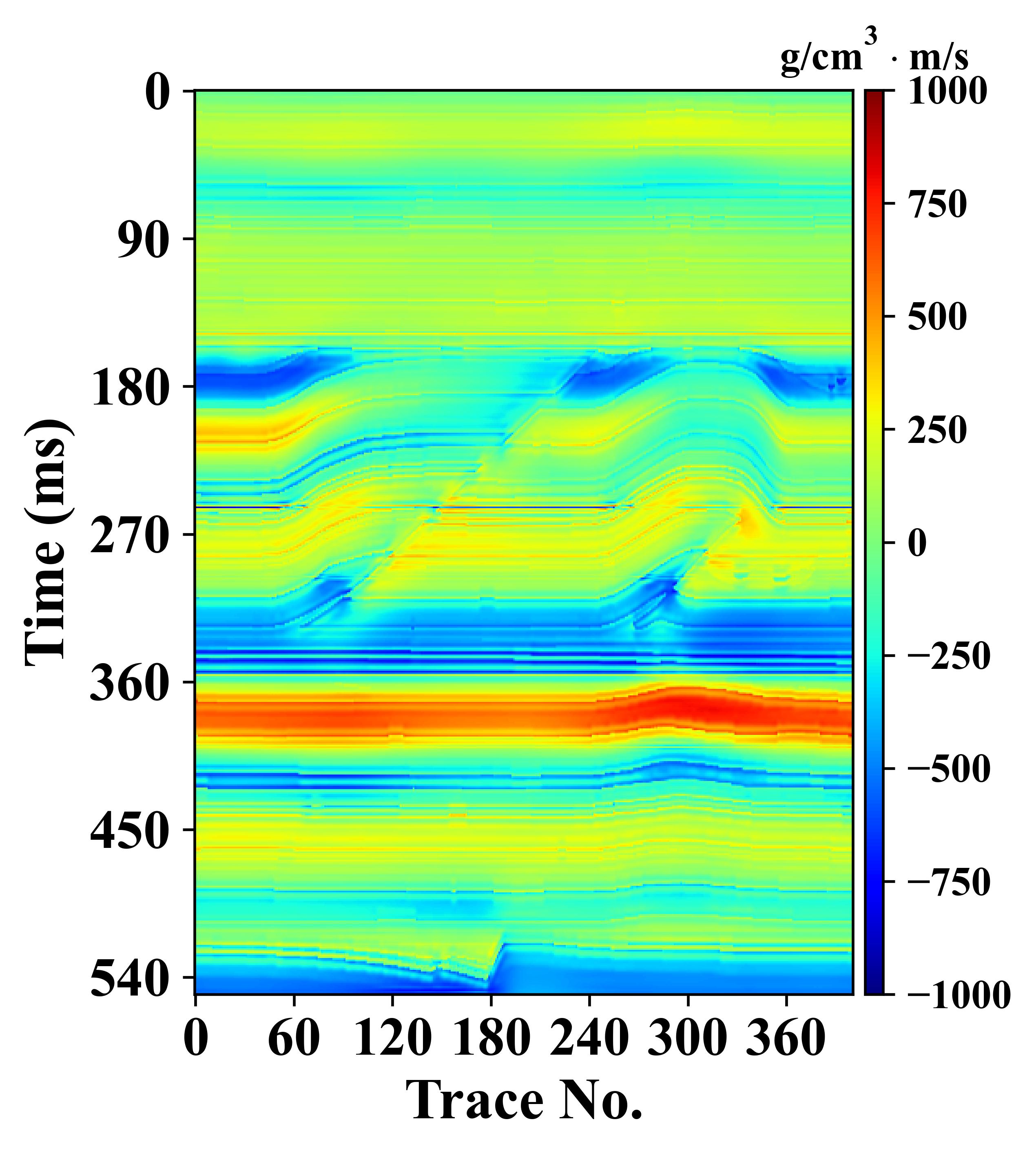}}\quad
   \subfloat[\label{fig:imp_cha_tv_0}]{\includegraphics[width = 0.18\textwidth]{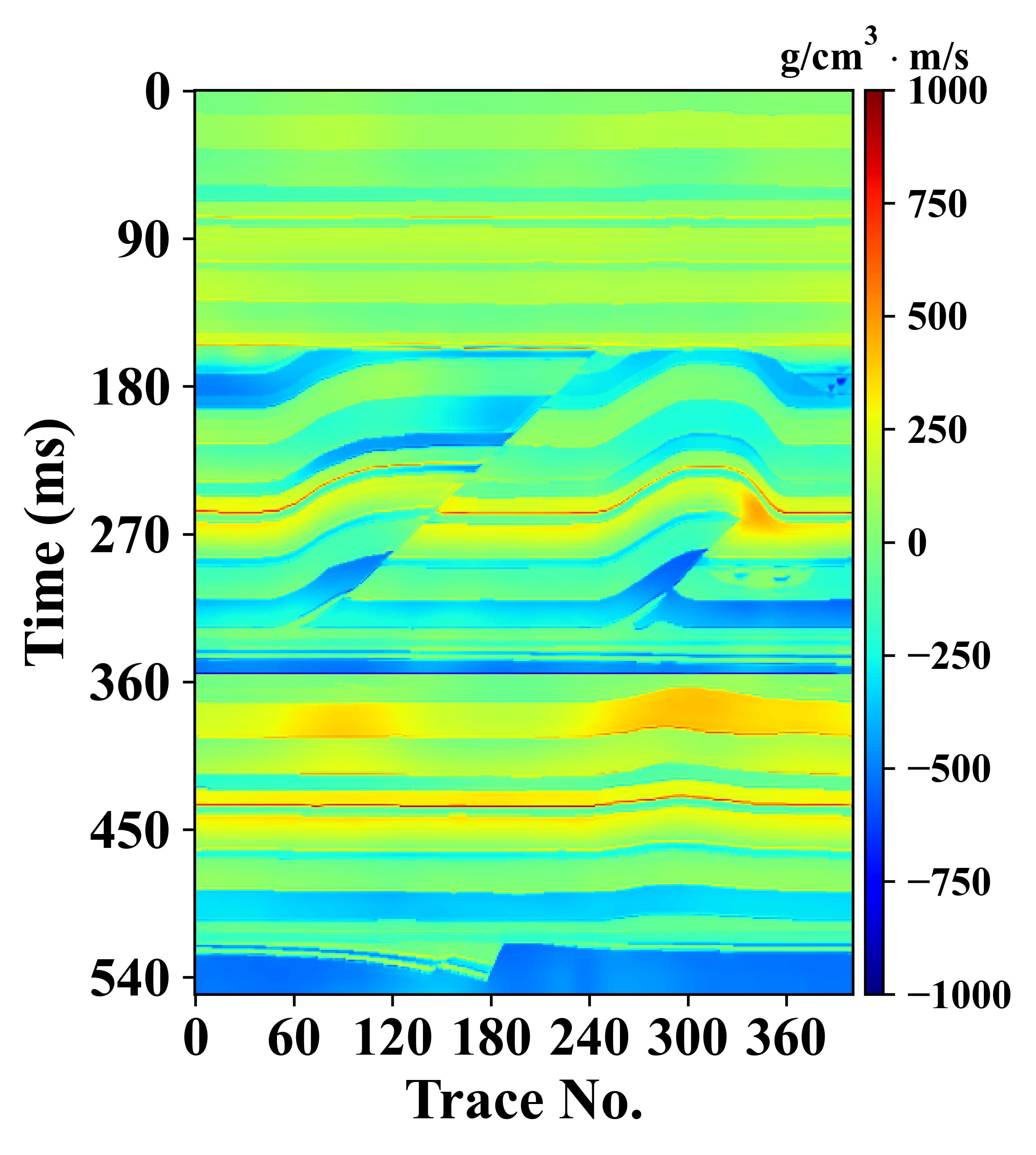}}\quad
  \caption{Impedance results of noise-free synthetic data and their residuals on Overthrust model. (a) SAII-CLDM, (b) SAII-LDDPM, (c) SDL, (d) USDL, (e) 2D-TV, (f-j) corresponding residuals in the first row.}
  \label{fig:imp_0}
\end{figure*}

\begin{figure*}[htbp]
  \centering
   \subfloat[\label{fig:imp_cldm_15}]{\includegraphics[width = 0.18\textwidth]{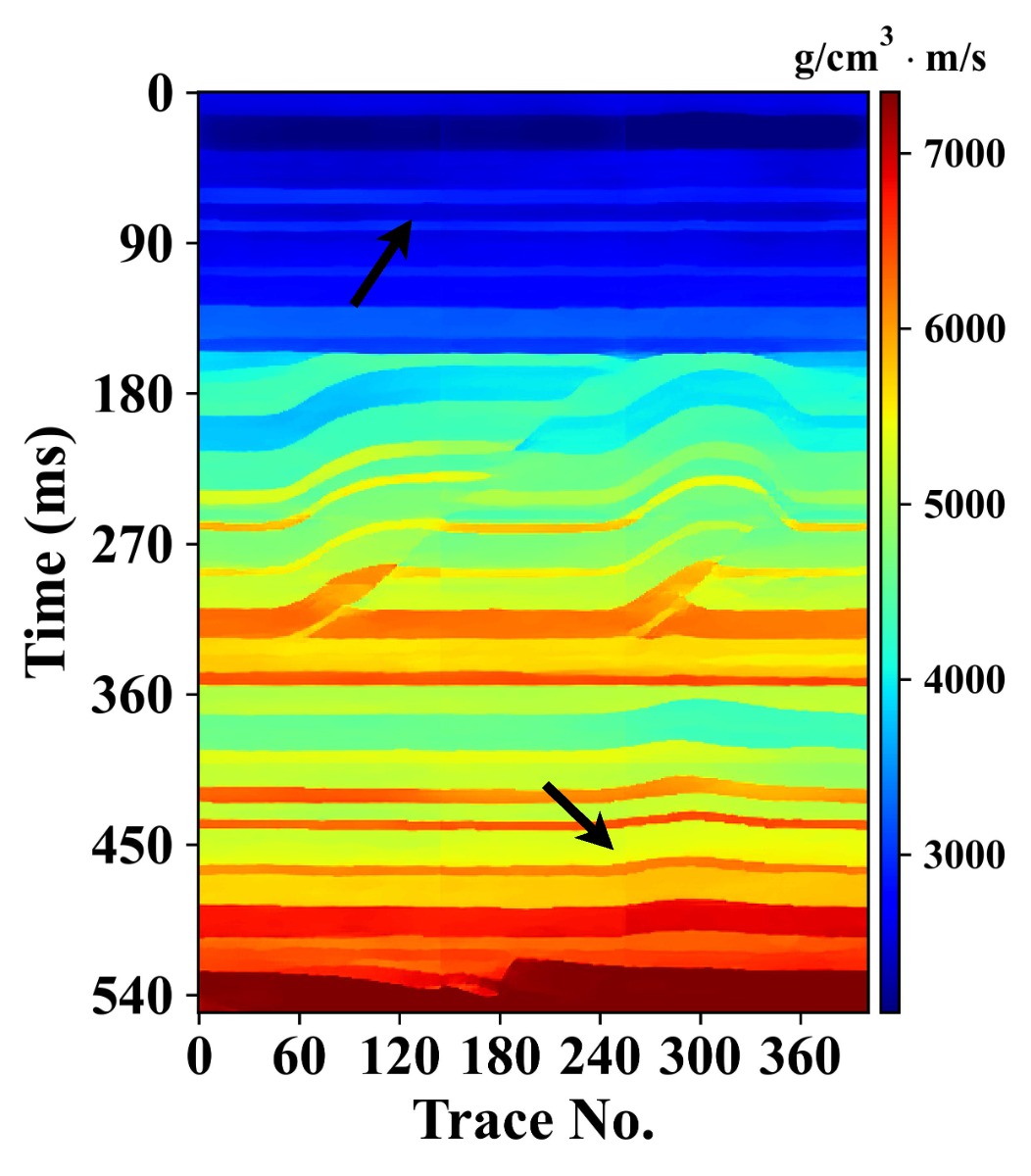}}\quad
  \subfloat[\label{fig:imp_ddpm_15}]{\includegraphics[width = 0.18\textwidth]{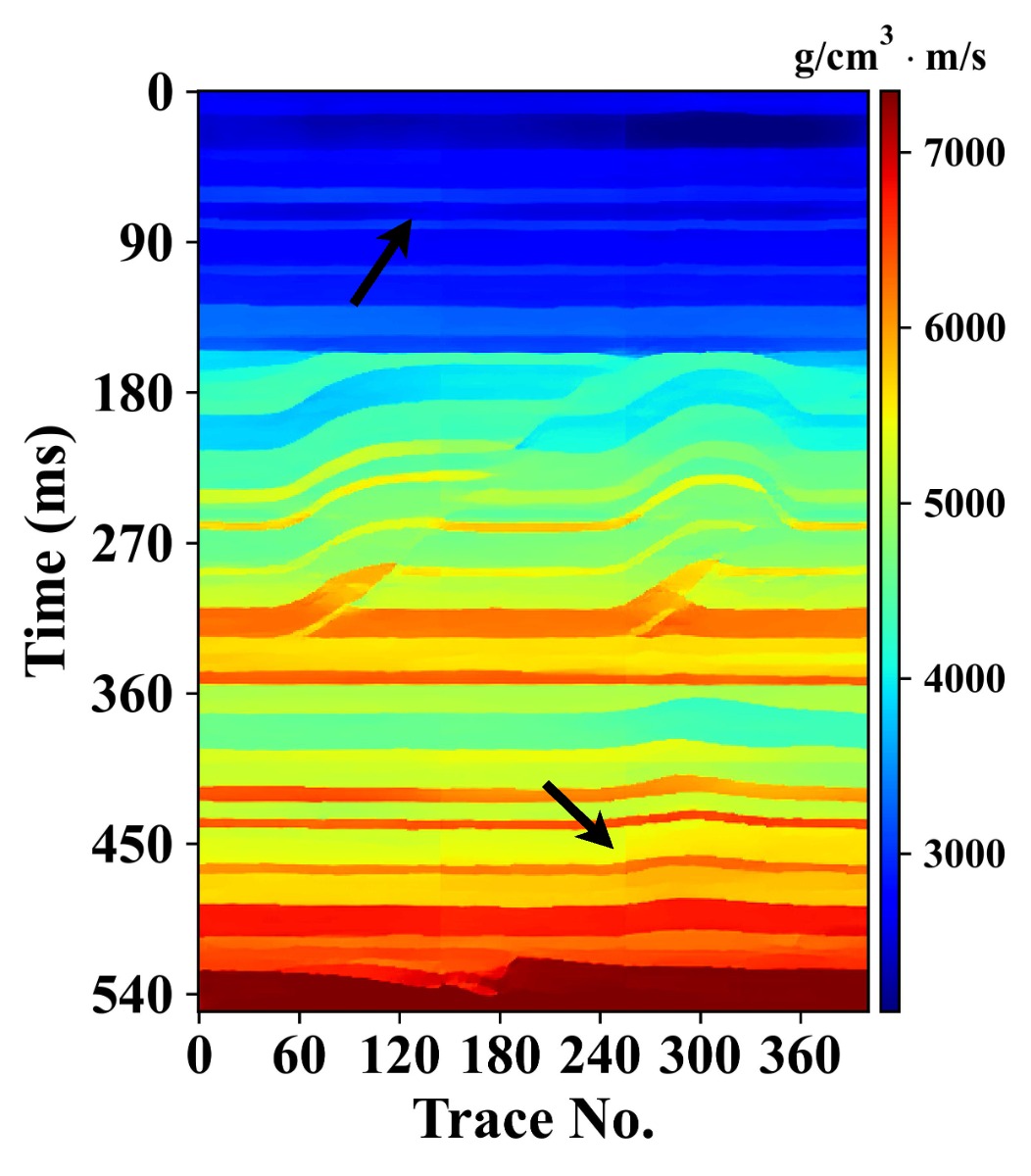}}\quad
    \subfloat[\label{fig:imp_sup_15}]{\includegraphics[width = 0.18\textwidth]{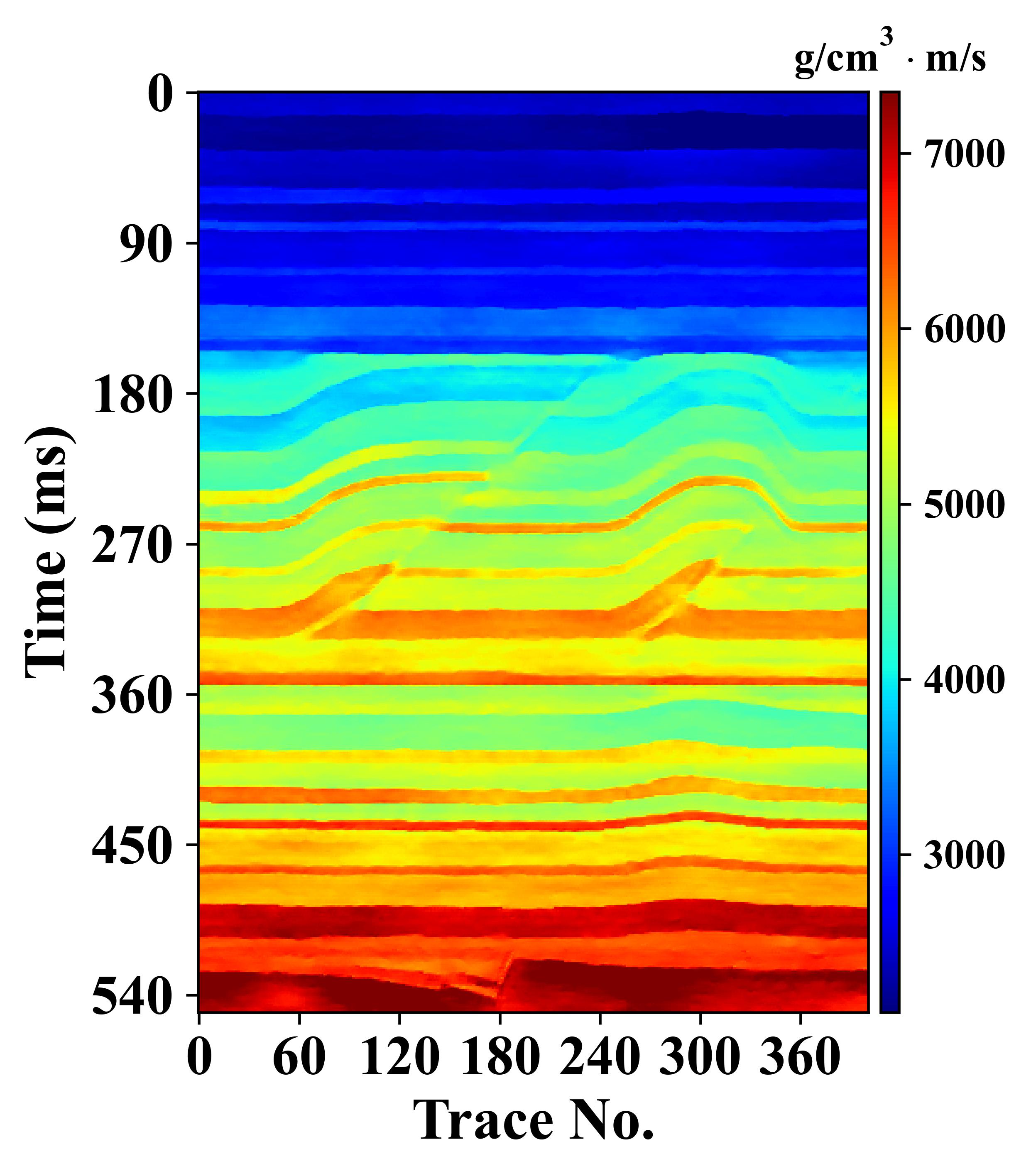}}\quad
  \subfloat[\label{fig:imp_unsup_15}]{\includegraphics[width = 0.18\textwidth]{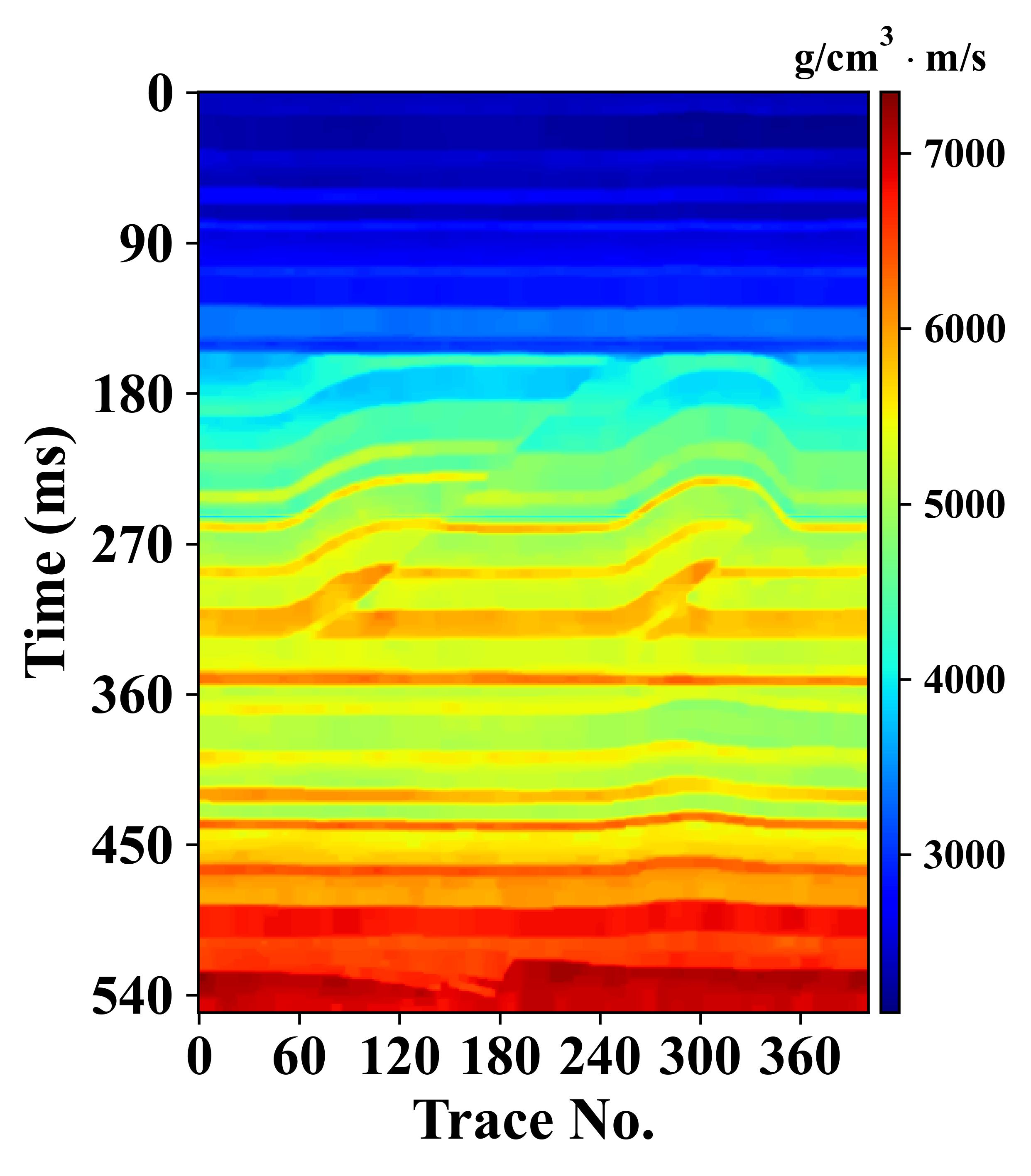}}\quad
  \subfloat[\label{fig:imp_tv_15}]{\includegraphics[width = 0.18\textwidth]{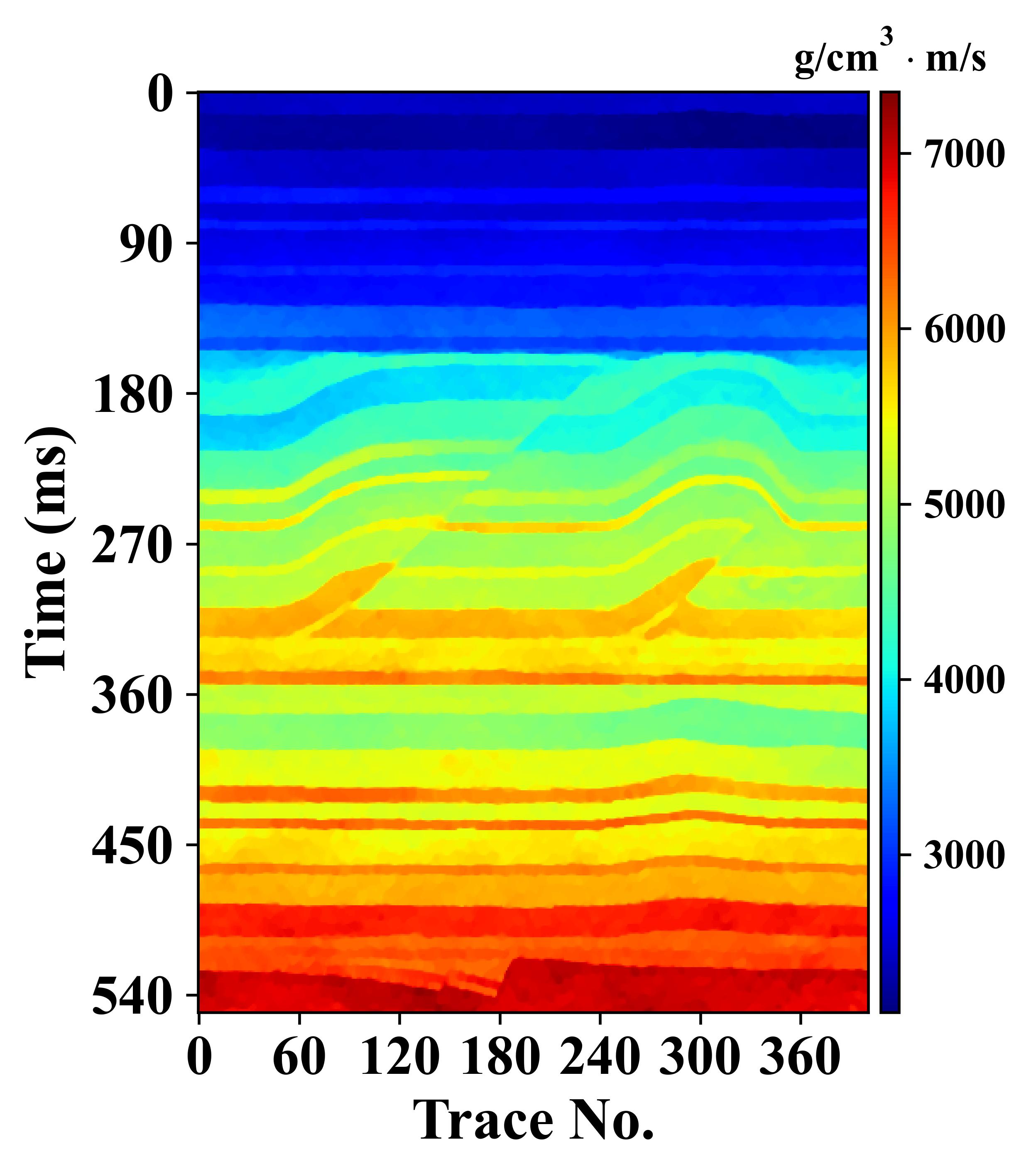}} 
    \\ 
  \subfloat[\label{fig:imp_cha_cldm_15}]{\includegraphics[width = 0.18\textwidth]{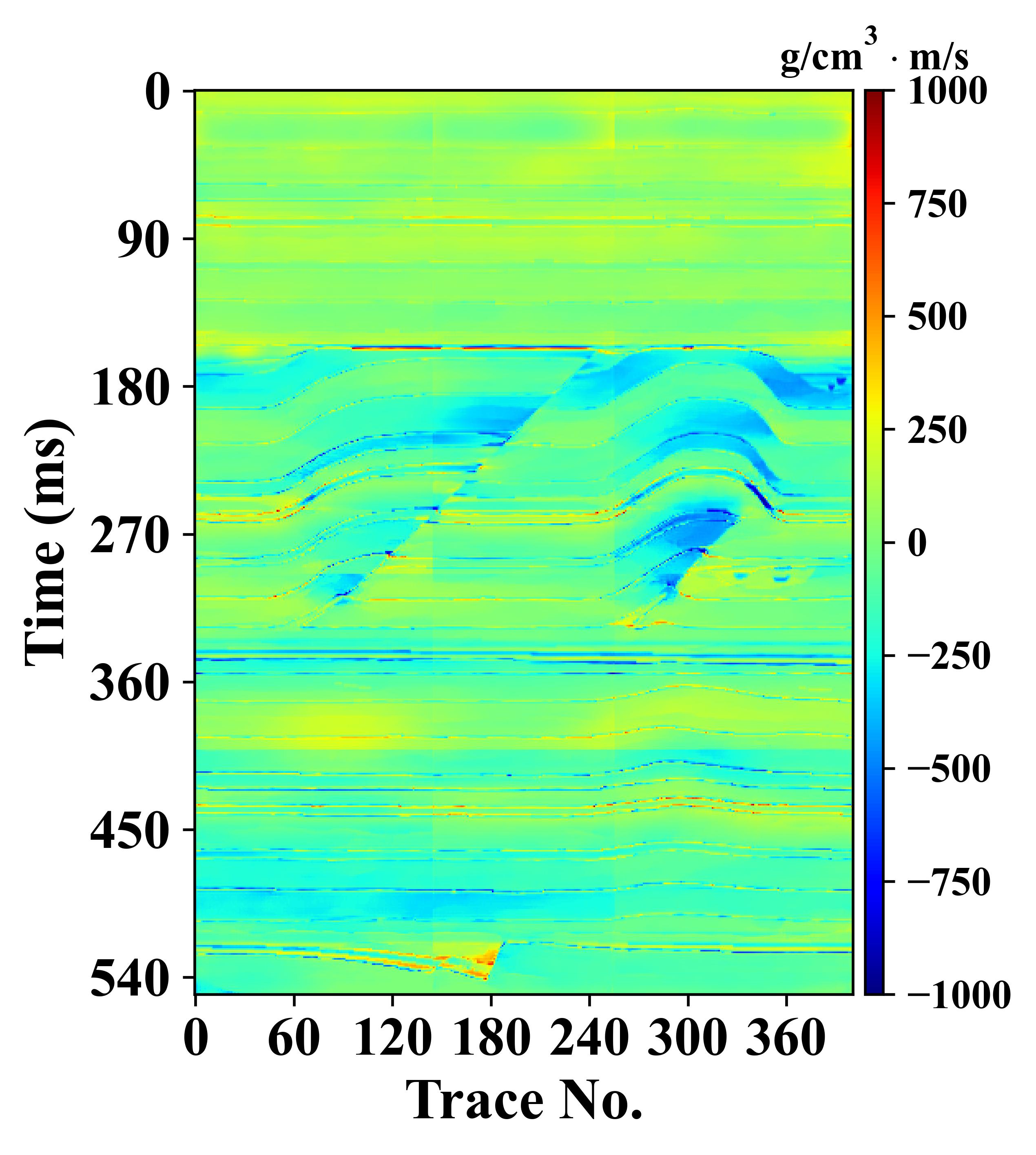}}\quad
  \subfloat[\label{fig:imp_cha_ddpm_15}]{\includegraphics[width = 0.18\textwidth]{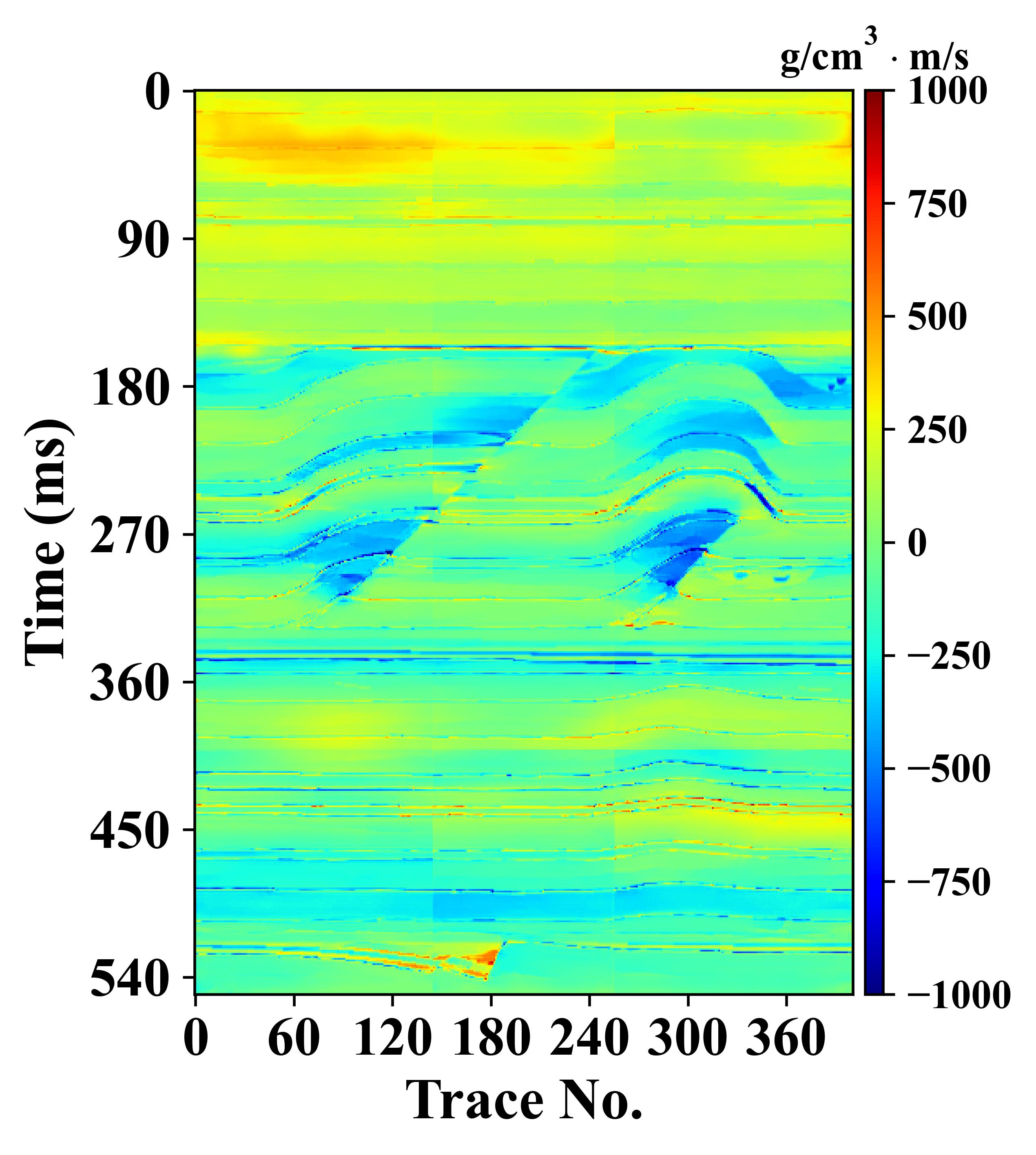}}\quad
  \subfloat[\label{fig:imp_cha_sup_15}]{\includegraphics[width = 0.18\textwidth]{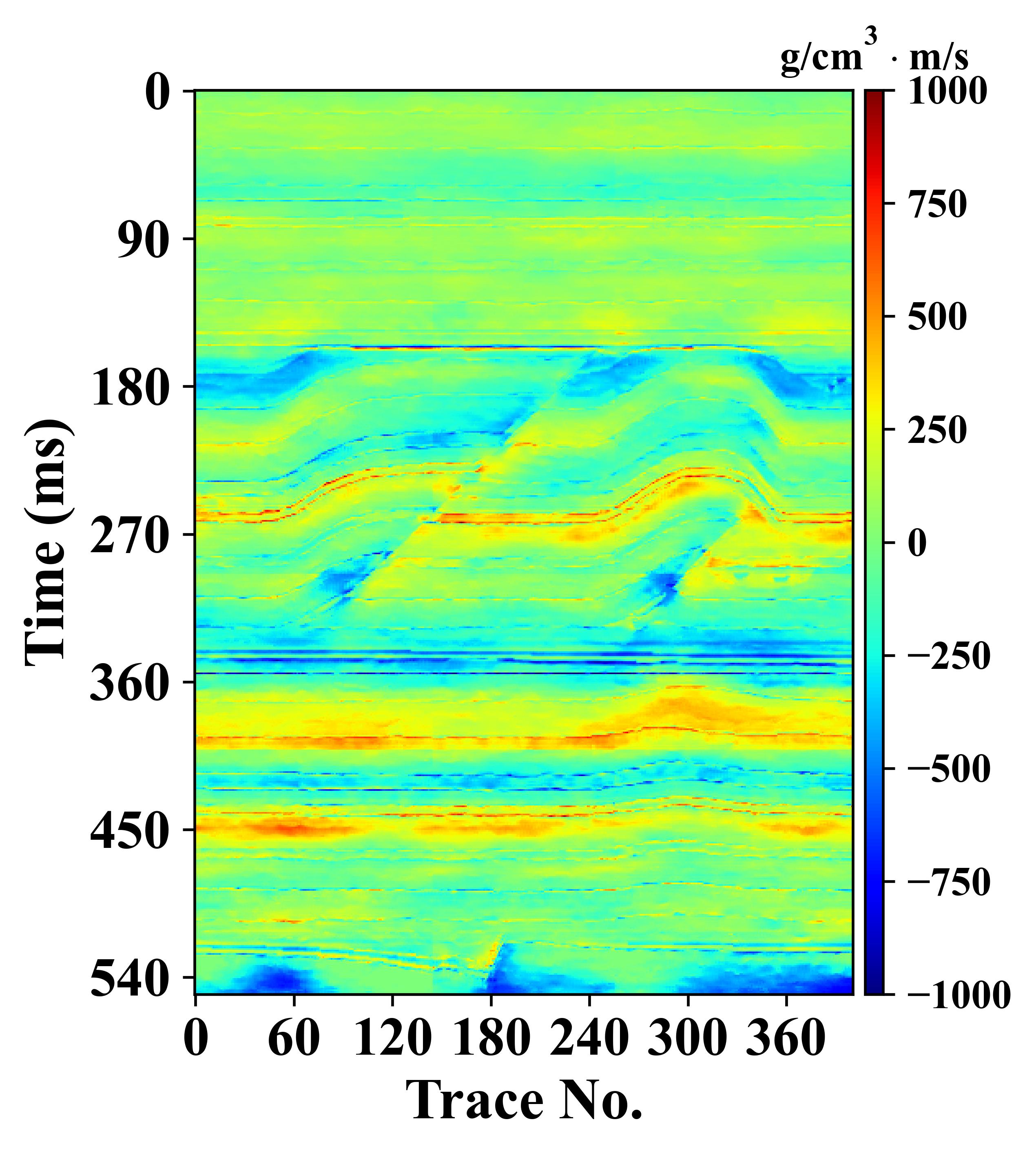}}\quad
    \subfloat[\label{fig:imp_cha_unsup_15}]{\includegraphics[width = 0.18\textwidth]{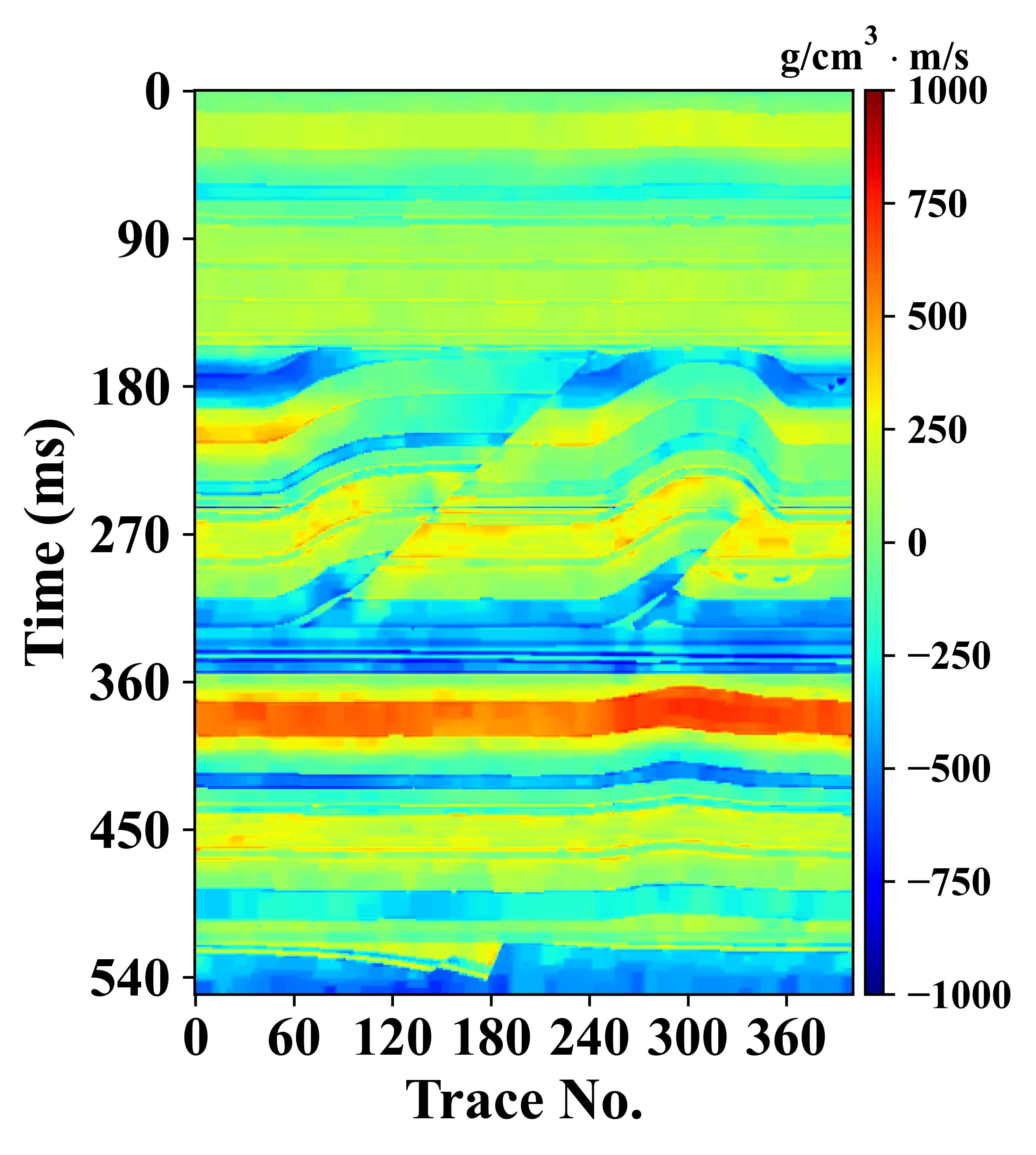}}\quad
     \subfloat[\label{fig:imp_cha_tv_15}]{\includegraphics[width = 0.18\textwidth]{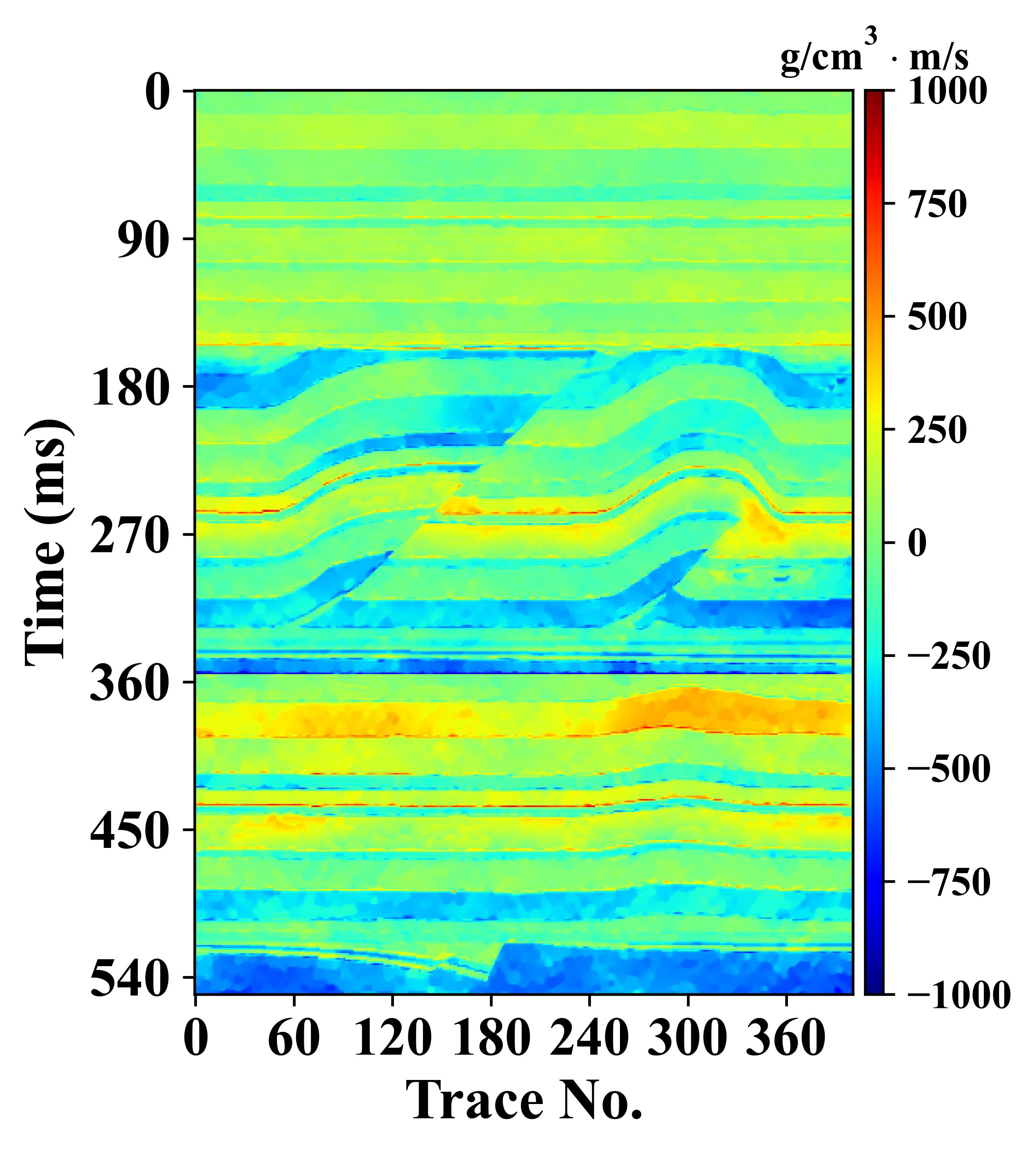}}
  \caption{Impedance results of 15~dB noisy synthetic data and their residuals on Overthrust model. (a) SAII-CLDM, (b) SAII-LDDPM, (c) SDL, (d) USDL, (e) 2D-TV, (f-j) corresponding residuals in the first row.}
    \label{fig:imp_15}
\end{figure*}

\begin{table*}[ht]
    \centering
%    \left
    \caption{Numerical Comparison Of Inversion Results on Overthrust Model}
    \begin{tabular}{lcccccccc}
        \toprule
        \multirow{2}{2cm}{Method} & \multicolumn{4}{c}{noise-free} & \multicolumn{4}{c}{15~dB noise} \\
        \cmidrule(lr){2-5} \cmidrule(lr){6-9}
        & PSNR $\uparrow$ & SSIM $\uparrow$ & PCC $\uparrow$ & RRE $\downarrow$ & PSNR $\uparrow$ & SSIM $\uparrow$ & PCC $\uparrow$ & RRE $\downarrow$ \\
        \midrule
        SAII-CLDM & \textbf{34.7593} & \textbf{0.9639} & \textbf{0.9967} & \textbf{0.0277} & \textbf{33.7181} & \textbf{0.9595} & \textbf{0.9961} & \textbf{0.0313} \\
        SAII-LDDPM & 32.6575 & 0.9605 & 0.9955 & 0.0353 & 32.0887 & 0.9553 & 0.9949 & 0.0377 \\
        SDL & 33.1399 & 0.9617 & 0.9944 & 0.0333 & 32.1987 & 0.9436 & 0.9930 & 0.0373 \\
        USDL & 29.0685 & 0.9432 & 0.9862 & 0.0528 & 28.8653 & 0.9459 & 0.9860 & 0.0535 \\
        2D-TV & 31.0512 & 0.9558 & 0.9932 & 0.0405 & 30.5057 & 0.9452 & 0.9917 & 0.0440 \\
        \bottomrule
    \end{tabular}
    \label{tab:performance}
\end{table*}

\subsection{Test Experiments}
To validate the effectiveness and robustness of the proposed method, we conduct experiments using the Overthrust model. The evaluation is divided into three parts. First, we assess the inversion accuracy and noise robustness of SAII-CLDM by comparing it with other methods under both noise-free and 15~dB noisy seismic data scenarios. Next, we investigate the impact of low-frequency impedance on inversion accuracy. Finally, we analyze the robustness of the proposed method against wavelet variations.

\subsubsection{Evaluation of Inversion Accuracy and Noise Robustness}\label{sec:eval_synthetic_inversion_noise}
In this section, we evaluate five methods' inversion accuracy and noise susceptibility using the Overthrust impedance model, the size of which is $551 \times 401$. We divide the model into six patches of size $256 \times 256$, matching the size of the training dataset. Fig.~\ref{fig:model_overthrust} shows the accurate impedance model and the low-frequency impedance model, respectively, where the latter is obtained by applying a 6~Hz cutoff frequency filter to the former. We then generate synthetic seismic data using a 30~Hz Ricker wavelet within the forward model described in Eq.~\eqref{eq:synthetic_record_noise}. Fig.~\ref{fig:model_overthrust_clearRecord} presents the noise-free seismic data, and Fig.~\ref{fig:model_overthrust_noisyRecord} contains 15~dB band-pass noise. Figs.~\ref{fig:imp_0} and \ref{fig:imp_15} compare the inversion results and residuals obtained using SAII-CLDM, SAII-LDDPM, SDL, USDL, and 2D-TV. Table~\ref{tab:performance} provides quantitative comparisons between the inverted and true impedance models using four metrics: peak signal-to-noise ratio (PSNR), structural similarity index measure (SSIM), pearson correlation coefficient (PCC), and relative root error (RRE). In the following analysis, we first compare the proposed SAII-CLDM and SAII-LDDPM methods to validate the effectiveness of the proposed model-driven sampling strategy described in Section~\ref{sec:model-driven_sample_strategy}. Finally, we compare the proposed method with three other comparison methods to demonstrate its advantages.

By comparing the inversion results of SAII-CLDM and SAII-LDDPM in Figs.~\ref{fig:imp_0} and \ref{fig:imp_15}, it can be observed that SAII-LDDPM fails to recover the shallow structures marked by arrows, as more clearly shown by the residuals. Furthermore, Table~\ref{tab:performance} shows that SAII-CLDM outperforms SAII-LDDPM across all metrics, with a particularly significant improvement in PSNR. Additionally, SAII-CLDM completes 30 timesteps in approximately 4 seconds, whereas SAII-LDDPM requires around 30 seconds for 1000 timesteps, demonstrating a substantial acceleration. Based on the above analysis, the proposed model-driven sampling strategy not only improves inversion performance but also significantly reduces the inversion time. 
Moreover, when compared with the method introduced in~\cite{geo_chl2024diffusion}, the proposed SAII-CLDM framework shows a slight reduction in accuracy. This can be attributed to the inherent accuracy loss in latent diffusion, which represents a necessary trade-off for computational efficiency.\label{Review1:4.3}

Next, we compare SAII-CLDM with three other methods: SDL, USDL, and 2D-TV. Fig.~\ref{fig:imp_0} presents the inversion results of noise-free synthetic seismic data. Through comprehensive observation, the inversion results obtained using both SAII-CLDM and SDL achieve high overall accuracy. However, SDL has difficulty resolving fine-scale geological features, as the black arrow indicates. The USDL method produces inaccurate results with substantial residual errors. The 2D-TV result appears spatially smooth but exhibits deviations in high-impedance regions marked by the arrow, likely due to inaccuracies in the low-frequency impedance model. To further assess noise susceptibility, Fig.~\ref{fig:imp_15} shows the inversion result of 15 dB synthetic seismic data in Fig.~\ref{fig:model_overthrust_noisyRecord}. It can be observed that the proposed SAII-CLDM maintains high accuracy and smoothness, closely matching its noise-free performance in Fig.~\ref{fig:imp_cldm_0}. In contrast, SDL exhibits significant discontinuities, while both USDL and 2D-TV suffer from boundary blurring compared to their noise-free inversion result. These observations highlight the superior noise robustness of the proposed method. Furthermore, the quantified results in Table~\ref{tab:performance} confirm that SAII-CLDM outperforms all other methods across all metrics, especially on SSIM.

%我们引入PSNR和SSIM指标，psnr时是基于网格间的均方误差计算，更能关注反演结果与真值的差异，然而难以准确体现反演结果的局部细节准确性。SSIM是基于局部窗口，对反演结果局部结构的准确性有更好的反映。

\begin{figure*}[htbp]
  \centering
  \subfloat[]{\includegraphics[width = 0.42\textwidth]{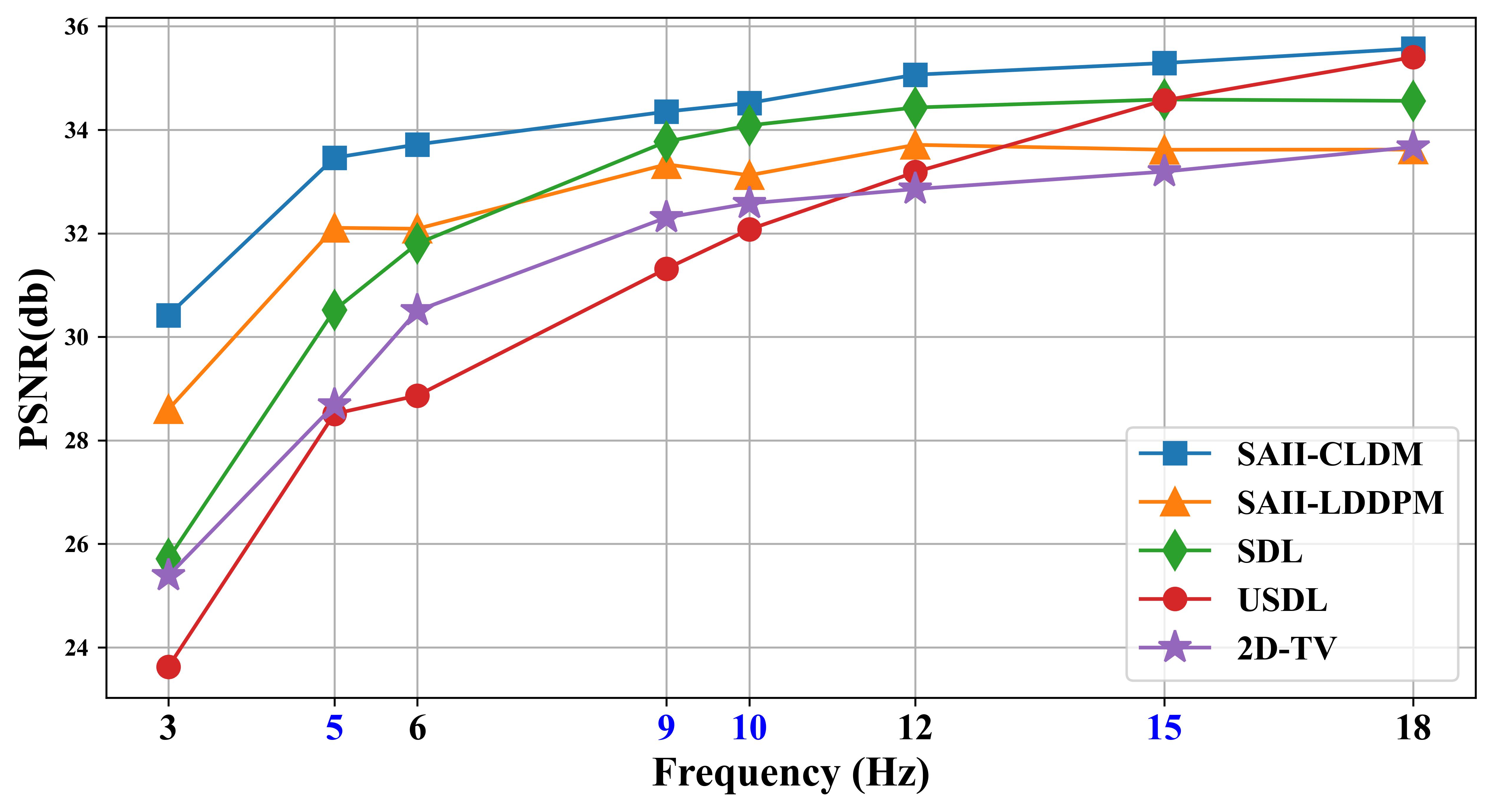}\label{fig:dipin_psnr}}\quad
  \subfloat[]{\includegraphics[width = 0.42\textwidth]{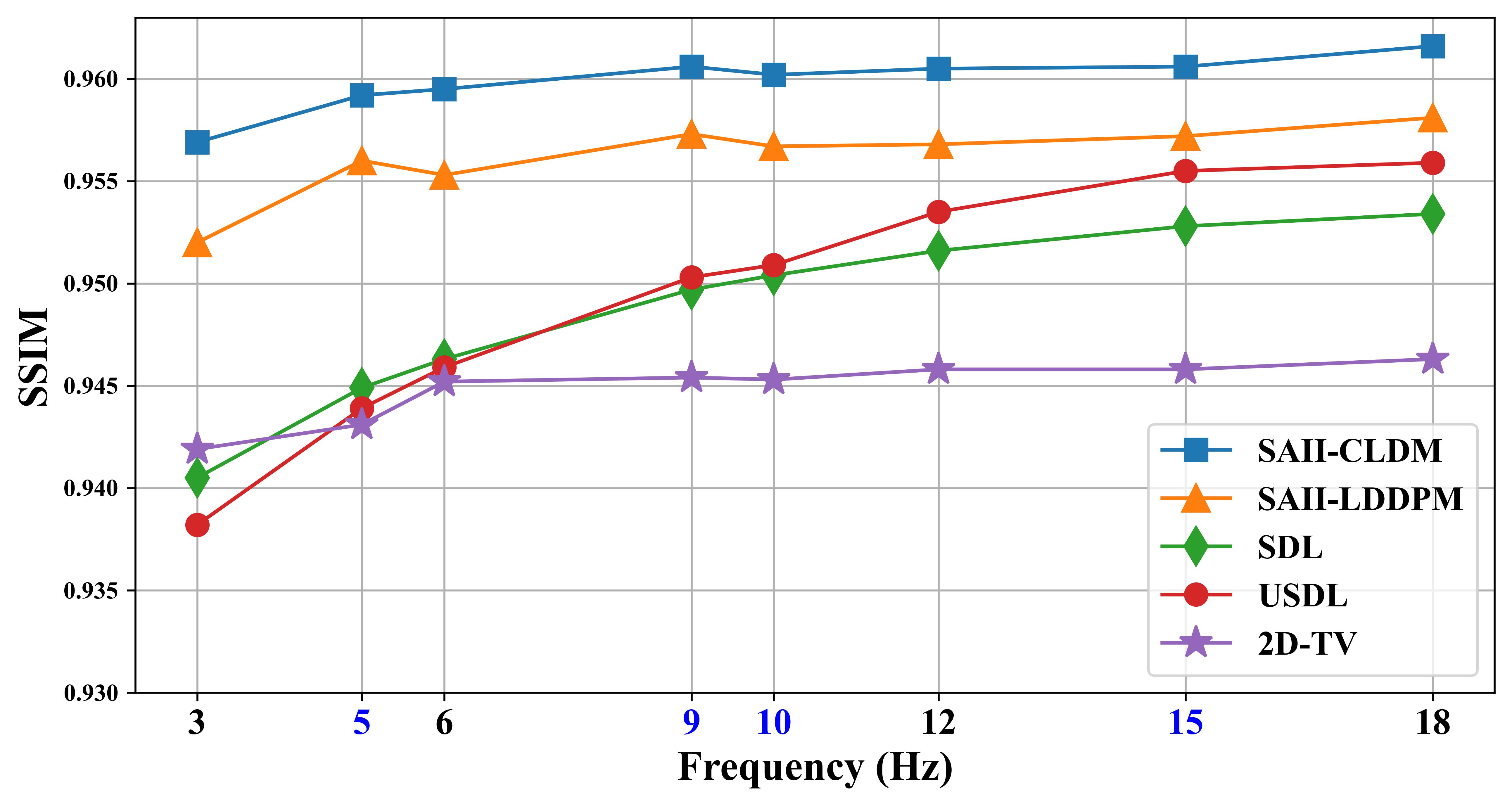}\label{fig:dipin_ssim}}\quad
  \caption{Influence of low‑frequency impedance on inversion accuracy, evaluated using (a) PSNR and (b) SSIM, for models generated with cutoff frequencies of 5, 9, 10, and 15 Hz that were not included in the training set.}
  \label{fig:dipin_compare}
\end{figure*}

\begin{figure*}[htbp]
  \centering
  \subfloat[]{\includegraphics[width = 0.42\textwidth]{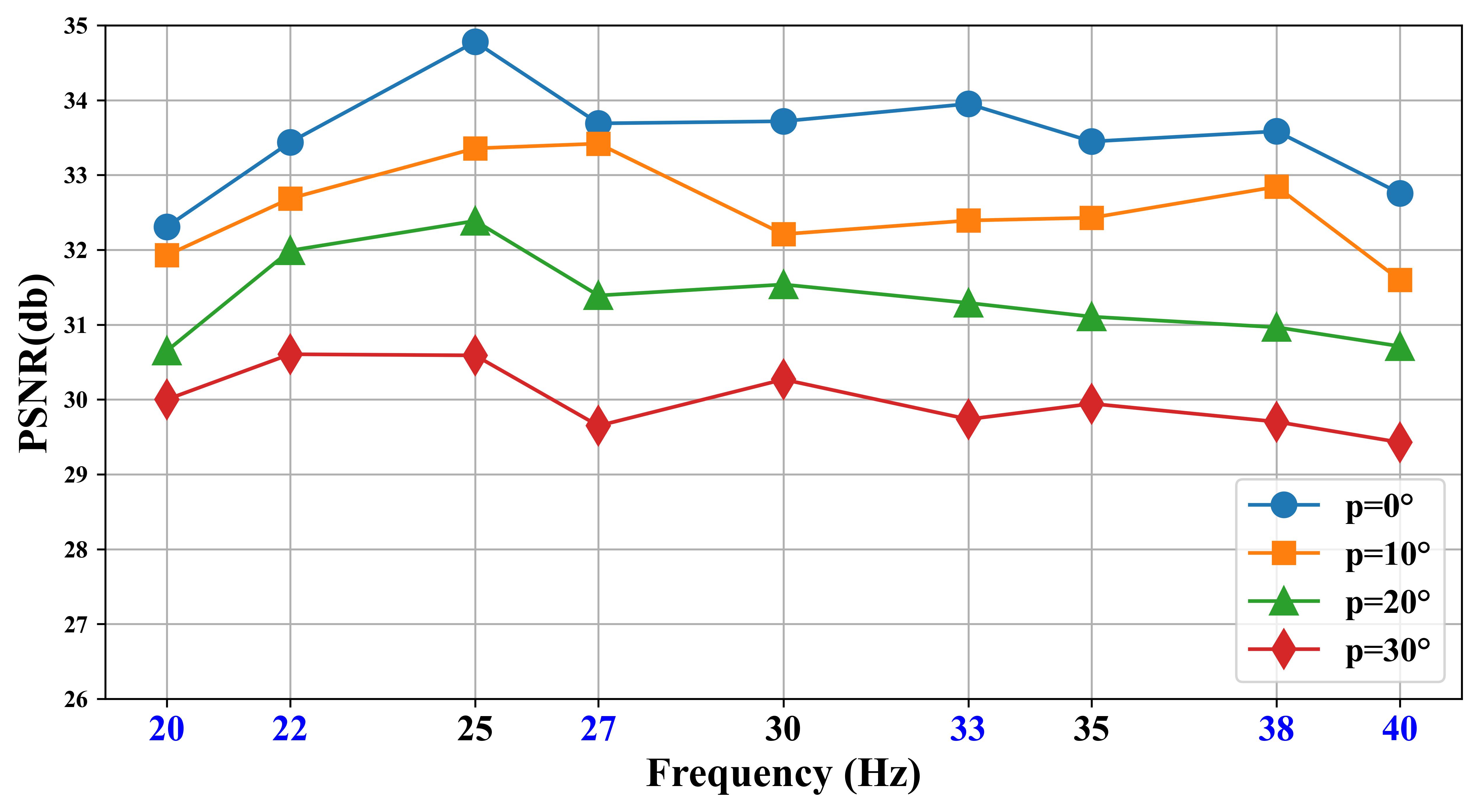}\label{fig:wavelet_psnr}}\quad
  \subfloat[]{\includegraphics[width = 0.42\textwidth]{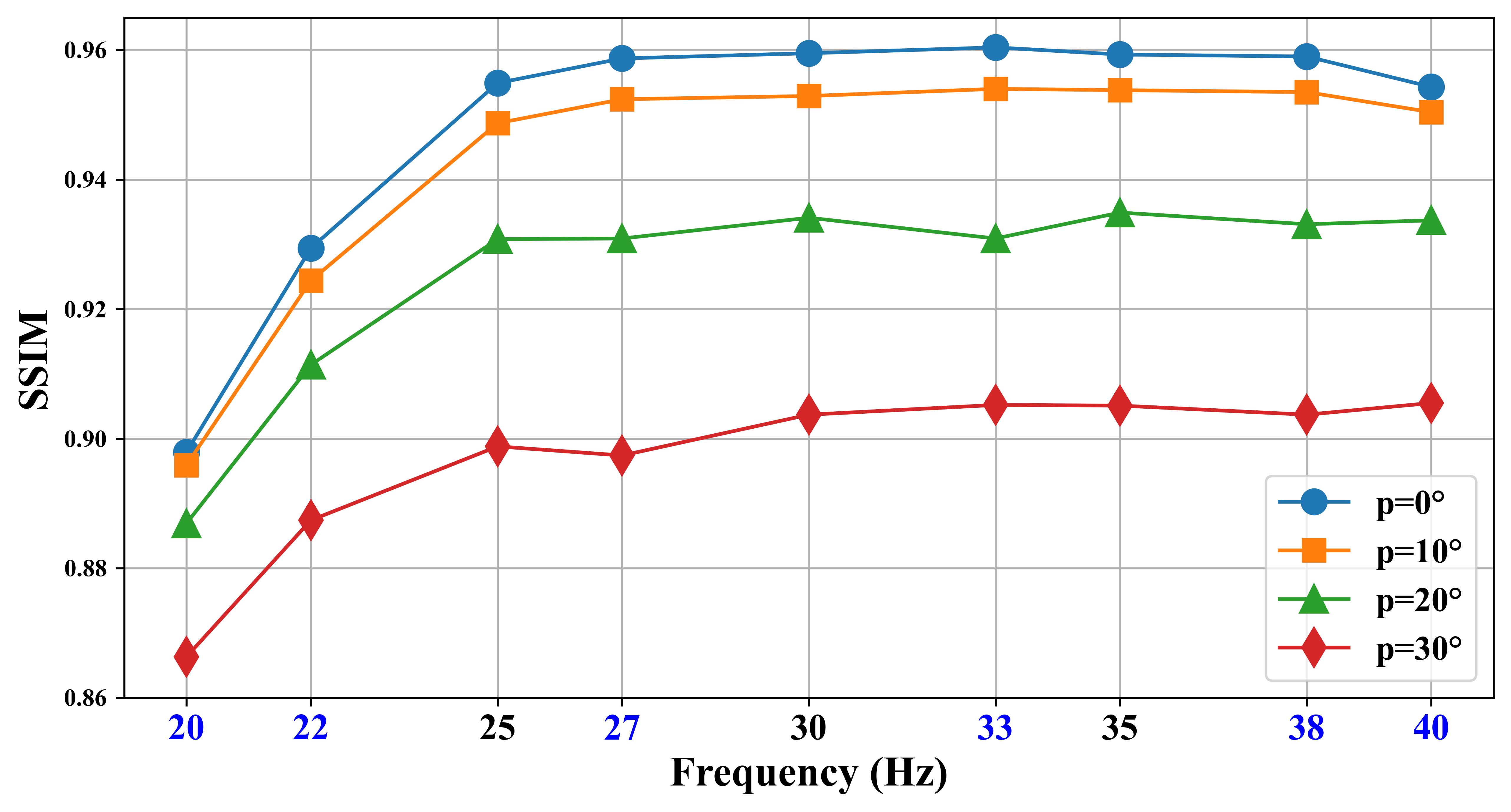}\label{fig:wavelet_ssim}}\quad
  \caption{Effects of wavelet dominant frequency and phase shift (p=$0^\circ$, $10^\circ$, $20^\circ$, and $30^\circ$) on impedance inversion accuracy, evaluated using (a) PSNR and (b) SSIM.
 }
  \label{fig:wavelets_test}
\end{figure*}
\subsubsection{Impact of Low-Frequency Impedance on Inversion Accuracy}
Low-frequency impedance plays a crucial role in inversion accuracy. To quantify its impact, we use PSNR and SSIM as evaluation metrics. PSNR based on mean squared error, captures overall differences between the inversion result and ground truth but is less sensitive to local structural details. In contrast, SSIM, computed within local windows, effectively assesses structural fidelity at finer scales. By combining these metrics, we provide a comprehensive evaluation. We evaluate inversion performance under ​15 dB synthetic seismic data while varying the accuracy of the low-frequency impedance. The training dataset excludes low-frequency impedance models generated using a low-pass filter with 5~Hz, 9~Hz, 10~Hz, or 15~Hz cutoff frequencies. Thus, neither SAII-CLDM nor SDL encounters these conditions during training, whereas USDL is retrained for each case. 

Fig.~\ref{fig:dipin_compare} shows that all five methods improved the inversion accuracy as the frequency increased, indicating that more accurate low-frequency impedance enhances inversion accuracy. The proposed SAII-CLDM outperforms other methods, particularly when the low-frequency impedance is inaccurate (below 9 Hz). Notably, SAII-CLDM achieves significantly higher PSNR and SSIM scores than SAII-LDDPM, demonstrating the effectiveness of the proposed sampling strategy. Compared to SAII-CLDM, SDL achieves a similar PSNR but a significantly lower SSIM. This discrepancy arises because SSIM is a window-based metric that is highly sensitive to local structural discontinuities, whereas PSNR measures global error and is less affected by localized variations. For instance, a comparison between the inversion results of SDL (Fig.~\ref{fig:imp_sup_15}) and SAII-LDDPM (Fig.~\ref{fig:imp_ddpm_15}) clearly illustrates this effect. The SDL inversion result exhibits noticeable local discontinuities. Although both methods achieve a similar PSNR of 32, their SSIM values differ significantly, with SAII-LDDPM achieving 0.9553 and SDL achieving 0.9436. Additionally, we observe that the performance of the USDL method is highly sensitive to the quality of the low-frequency impedance model. When provided with high-quality low-frequency impedance, USDL performs exceptionally well. Regarding the 2D-TV method, enhanced low-frequency impedance information significantly improves PSNR, while SSIM remains relatively unchanged. This suggests that the method struggles to recover local structural details. As shown in Fig.~\ref{fig:imp_tv_15}, the inversion results exhibit unclear boundaries. One possible explanation is that the 2D-TV method relies on linear assumptions, which may fail to effectively handle the nonlinear interference inherent in seismic data.

Besides, it is worth noting that the proposed method performs well even under unseen low-frequency impedance conditions during training, as indicated in blue frequency labels in Fig.~\ref{fig:dipin_compare}. This demonstrates that the proposed method exhibits a certain degree of generalization to unseen low-frequency impedance conditions in the training dataset.

%1. 从频率变化的趋势看，主频越高，提供的高频成分越多，SSIM越高，细节信息越好
%2. 从相位的变化看，相位与训练数据集差距越大，psnr变化不大，说明整体变化不大，而ssim变化大，说明细节变化大。
\subsubsection{Robustness Analysis Against Wavelet Variations}  
When the seismic data in the training dataset differ significantly from those in the test dataset, the inversion performance may degrade. Since seismic data are primarily influenced by wavelet characteristics, we evaluate the robustness of the proposed method under varying wavelet conditions. During tests, we generate test noisy seismic datasets using wavelets with dominant frequencies ranging from 20~Hz to 40~Hz (20, 22, 25, 27, 30, 33, 35, 38, and 40~Hz) and phase shifts of p = $ 0^\circ, 10^\circ, 20^\circ, 30^\circ$. It is worth noting that the training dataset consists of synthetic seismic records generated using zero-phase Ricker wavelets with dominant frequencies of 25~Hz, 30~Hz, and 35~Hz, as described in Section~\ref{sec:training_dataset}. Fig.~\ref{fig:wavelets_test} presents two line charts depicting the relationship between wavelet characteristics and inversion accuracy: one illustrating PSNR trends and the other showing SSIM variations. Below, we analyze the influence of wavelet characteristics based on Fig.~\ref{fig:wavelets_test}:

As the dominant frequency increases, neither PSNR nor SSIM shows a significant correlation with the variations in dominant frequency. This can be attributed to the inclusion of synthetic records with dominant frequencies of 25 Hz, 30 Hz, and 35 Hz in the training dataset, which ensures comprehensive coverage of the tested dominant frequency range. However, when the dominant frequency is below 9 Hz, the SSIM is noticeably lower. This is likely due to low-frequency wavelets failing to supply the richer high-frequency components necessary for resolving inversion details, resulting in lower SSIM values. Regarding wavelet phase shifts, both PSNR and SSIM demonstrate sensitivity, primarily because the training dataset contains only zero-phase wavelets. When the phase shift exceeds $20^\circ$, a noticeable decline in both PSNR and SSIM is observed. Based on the above analysis, accurately estimating the wavelet beforehand and incorporating corresponding synthetic records in training can help enhance inversion accuracy.

To further evaluate the influence of wavelets within the training range, we analyze cases where the dominant frequency falls between 25~Hz and 35~Hz, and the phase shift remains below 20∘. Under these conditions, both metrics demonstrate strong performance, with PSNR consistently exceeding 32~dB and SSIM remaining above 0.95. Even in boundary cases, such as at 38~Hz or when the phase shift reaches 20∘, performance remains acceptable, with PSNR above 31 dB and SSIM exceeding 0.93. These results confirm that the proposed method is effective within the training range and can be reliably applied in practical scenarios where frequency and phase variations are limited.

\section{Field Data Numerical Example}
\label{sec:field_example}
In this section, we evaluate the effectiveness of the proposed method on field seismic data and compare its performance with SDL, USDL, and 2D-TV approaches as described in Section~\ref{sec:compare_method}.

\subsection{Training Details and Test Details}
\subsubsection{Training dataset}

Given the significant differences between synthetic and field data, constructing a new dataset for network retraining is essential. The available 3D field dataset contains 10 well logs. To generate a representative training impedance dataset, we first analyze the distribution of these well logs and set various lateral ranges in the variogram function to simulate the training impedance dataset via Sequential Gaussian Simulation without treating the well logs as hard constraints. Additionally, we apply Geometric Transformation to augment the dataset, resulting in 2000 datasets, each with a size of $128\times 128$. The low-frequency impedance in the training dataset is generated by applying a 6~Hz cutoff frequency filter to the training impedance models. Furthermore, wavelets are estimated from the available well logs and used with a nonlinear forward model to generate the seismic data for training.

\subsubsection{Training Procedure}
\label{sec:field_training_procedure}

The training process follows the same two-step procedure and hyperparameter selection principle as those used in the synthetic-data training described in Section~\ref{sec:training_paramerter_synthetic}. First, the VQ-GAN is trained for approximately 200 epochs with an initial learning rate of $4.5 \times 10^{-5}$ and a batch size of 12. Then, the DDPM model is trained for approximately 100 epochs with a learning rate of $2 \times 10^{-6}$ and a batch size of 24.

 \subsubsection{Test Details}
\begin{figure}[htbp]
  \centering
  \includegraphics[width=0.49\textwidth]{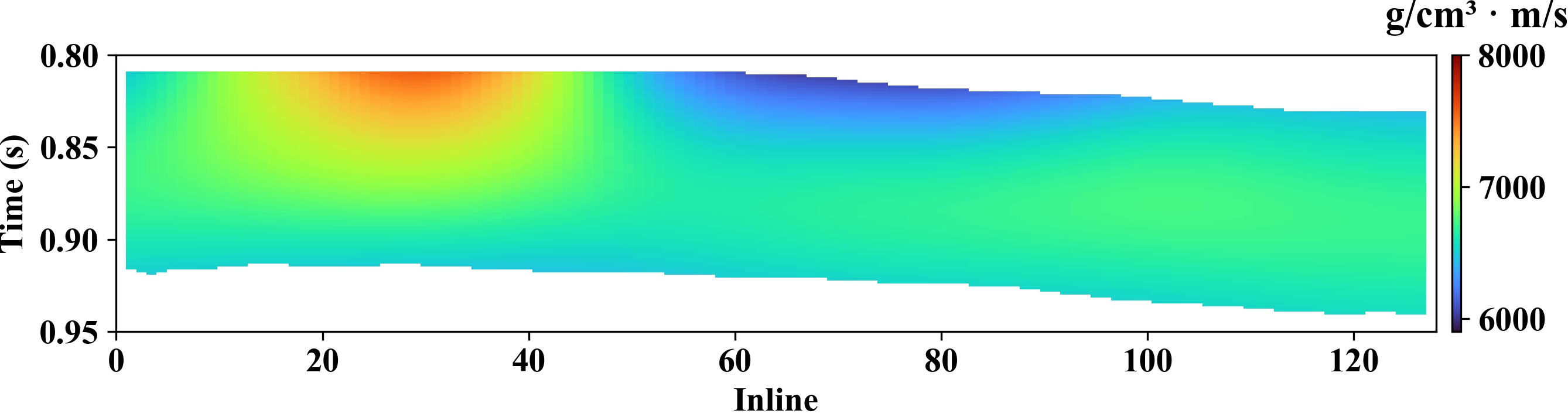}
  \caption{Low-frequency impedance of the field data.}
  \label{fig:field_dipin}
\end{figure}

 For evaluation, we select a crossline profile from the 3D field dataset as the 2D test seismic data, which includes two impedance logs located at inline positions 99 and 121. The low-frequency impedance is estimated by interpolating the 10 well logs and applying a 6 Hz cutoff frequency filter, as illustrated in Fig.~\ref{fig:field_dipin}. The seismic wavelet used in testing is the same as in the training dataset.
 
We compare the proposed method with three approaches: the SDL method, the USDL method, and the 2D-TV method, as described in Section~\ref{sec:compare_method}. Similar to the synthetic data experiments, both SAII-CLDM and SDL are trained on the training datasets, whereas USDL is trained only on the test datasets.

%
%\begin{figure*}[htbp]
%  \centering
%    \subfloat[]{\includegraphics[width = 0.5\textwidth]{pic/exp3_field/ddim_inv.jpg}\label{fig:field_ddim}} 
%    \subfloat[]{\includegraphics[width = 0.5\textwidth]{pic/exp3_field/resample_4_inv.jpg}\label{fig:field_resample1}}
%     
%      \subfloat[]{\includegraphics[width = 0.5\textwidth]{pic/exp3_field/resample_5_inv.jpg}\label{fig:field_resample2}}
%     \quad
%
%    \quad
%    \subfloat[]{\includegraphics[width = 0.3\textwidth]{pic/exp3_field/ddim_seismic.jpg}\label{fig:field_ddim}}     
%     \subfloat[]{\includegraphics[width = 0.3\textwidth]{pic/exp3_field/resample_4_seismic.jpg}\label{fig:field_resample2_seismic}}
%     \subfloat[]{\includegraphics[width = 0.3\textwidth]{pic/exp3_field/resample_5_seismic}\label{fig:field_resample2_seismic}}
%
%  
%  \caption{Inverted impedance of the field data. (a) inverted impedance of SAII-CLDM, (b) inverted impedance of SDL, (c) inverted impedance of USDL, (c) inverted impedance of 2D-TV.}
%  \label{fig:field_imps}
%\end{figure*}

\subsection{Evaluation of Inversion Accuracy on Field Data}
\label{sec:Evaluation_on_Field}
\begin{figure*}[htbp]
  \centering
    \subfloat[]{\includegraphics[width = 0.48\textwidth]{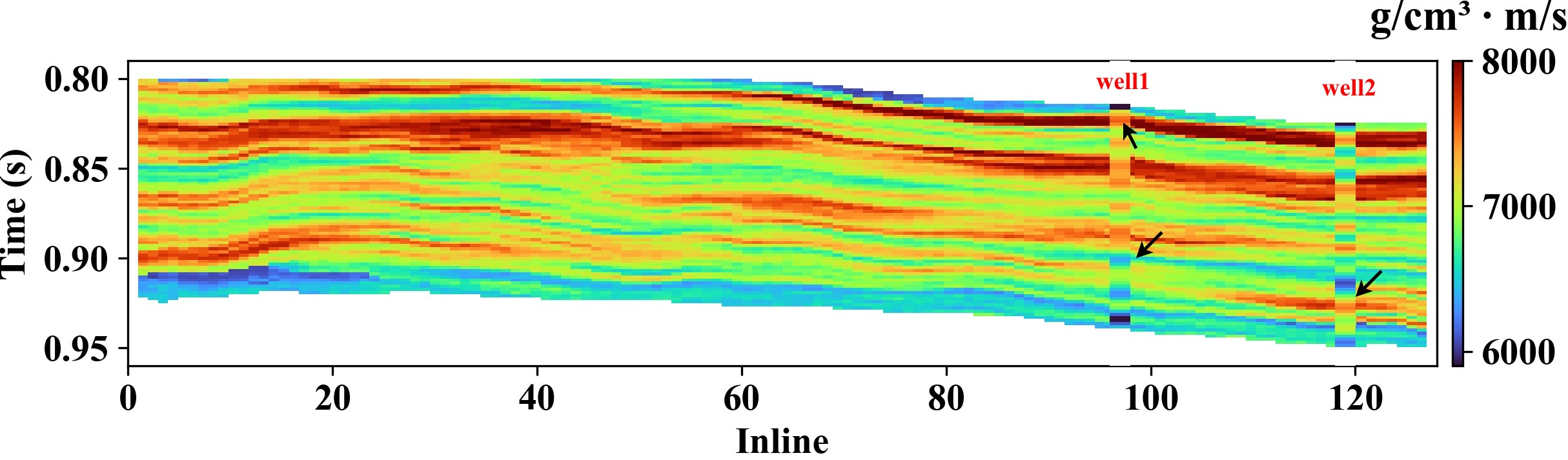}\label{fig:field_ldm}} \quad \quad 
     \subfloat[]{\includegraphics[width = 0.48\textwidth]{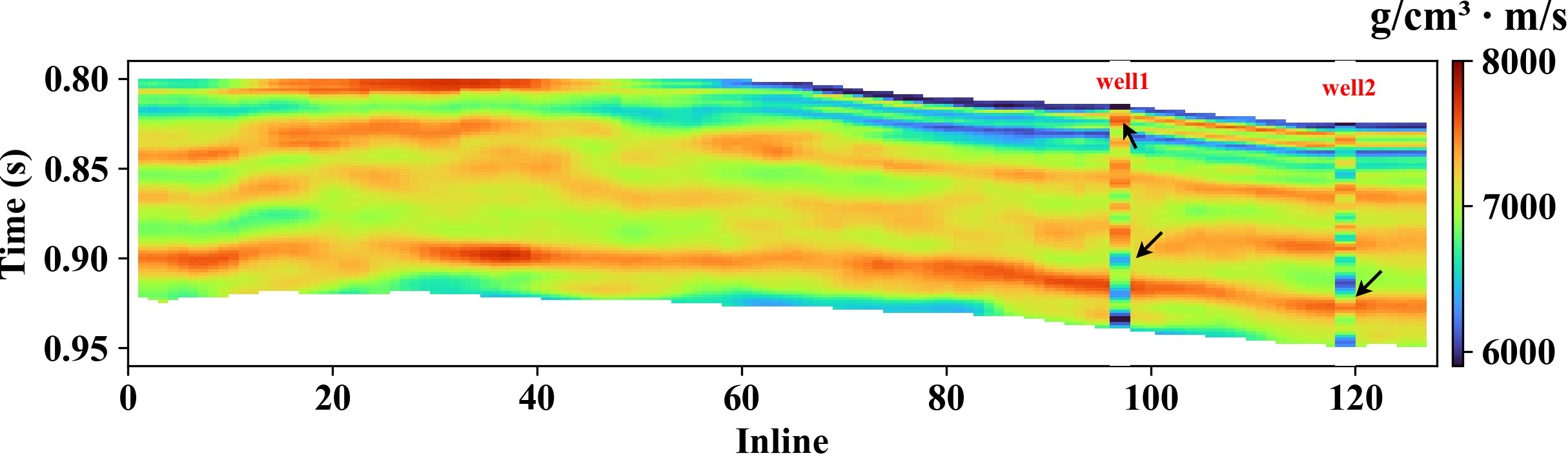}\label{fig:field_sup}}\\
     \subfloat[]{\includegraphics[width = 0.48\textwidth]{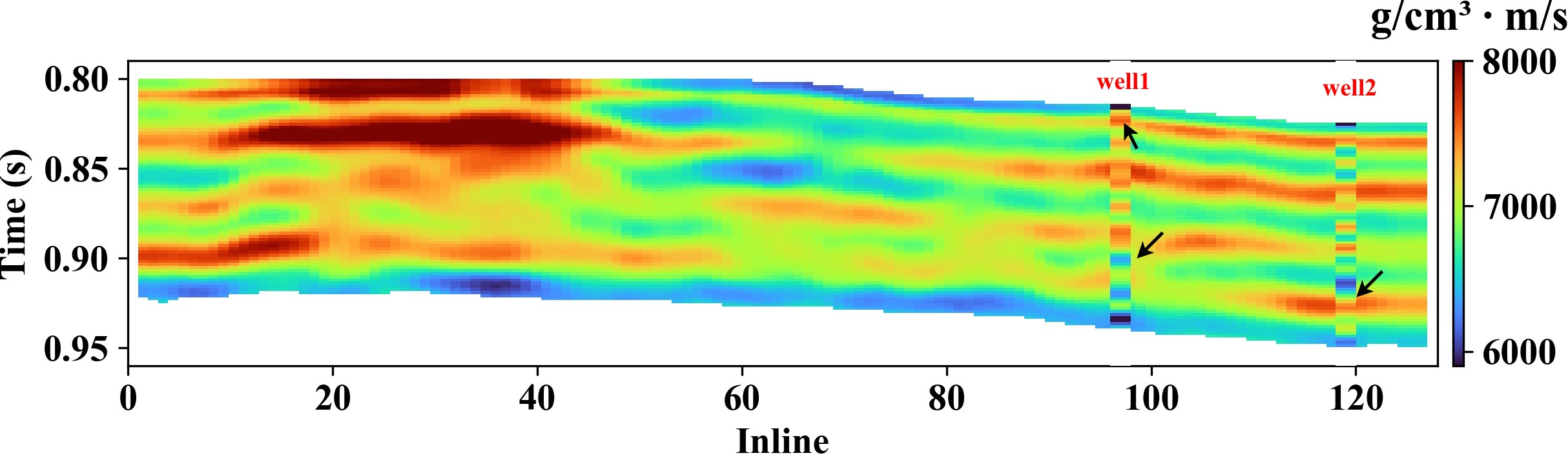}\label{fig:field_unsup}} \quad \quad 
      \subfloat[]{\includegraphics[width = 0.48\textwidth]{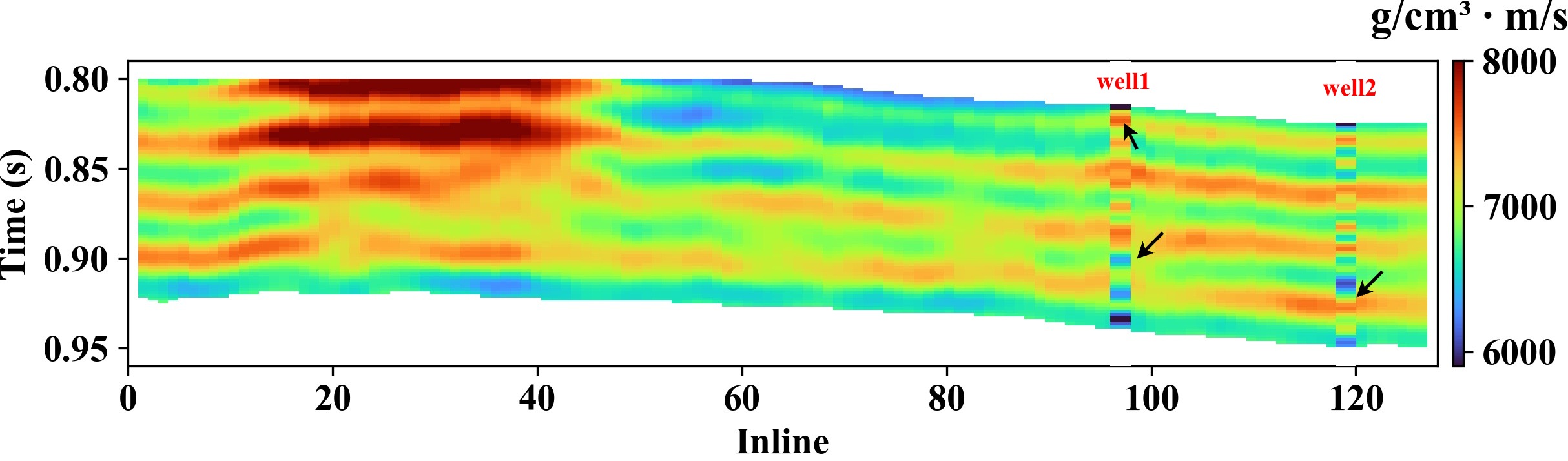}\label{fig:field_pd}}  
  \caption{Inverted impedance of the field data. (a) inverted impedance of SAII-CLDM, (b) inverted impedance of SDL, (c) inverted impedance of USDL, (d) inverted impedance of 2D-TV.}
  \label{fig:field_imps}
\end{figure*}

Figs.~\ref{fig:field_imps} shows the inverted impedance results obtained using the proposed SAII-CLDM, along with SDL, USDL, and 2D-TV. Two impedance logs are embedded within the inversion results to facilitate an objective assessment. Figs.~\ref{fig:field_imps} demonstrate that the proposed SAII-CLDM exhibits superior continuity, clearer boundaries, and higher resolution compared to the other methods. Notably, it shows a more precise alignment with the impedance logs, particularly in the regions indicated by the arrows. In contrast, although SDL produces relatively high-resolution results, it fails to capture subtle details and exhibits unclear boundaries. When compared with the impedance logs, the overall trend is consistent, but the details are less accurate. This limitation arises because SDL learns from the training dataset using the end-to-end framework, which directly learns the mapping relationship of input and output. However, the absence of true labels in the training dataset hinders SDL's ability to generalize effectively, leading to suboptimal performance. Compared to the proposed method, USDL and 2D-TV generate lower-resolution and coarser results, showing only a rough alignment with the well logs. This discrepancy stems from their reliance on the convolution model, which does not fully align with the actual forward model, resulting in reduced resolution and less detailed outputs. Furthermore, we record the Pearson Correlation Coefficient (PCC) between each inverted impedance and the corresponding well log in Table~\ref{tab:well_pcc}. The proposed method achieves a higher PCC value than the other methods, indicating better agreement with the well-log data. This confirms the effectiveness of the proposed method. It should be noted that the PCC values of these four methods are not exceptionally high, primarily due to the difficulty of accurately aligning well logs with seismic data near well locations, which remains a challenging task in practice.

\begin{figure*}[htbp]
  \centering
%  \left
  \subfloat[]{\includegraphics[width = 0.32\textwidth]{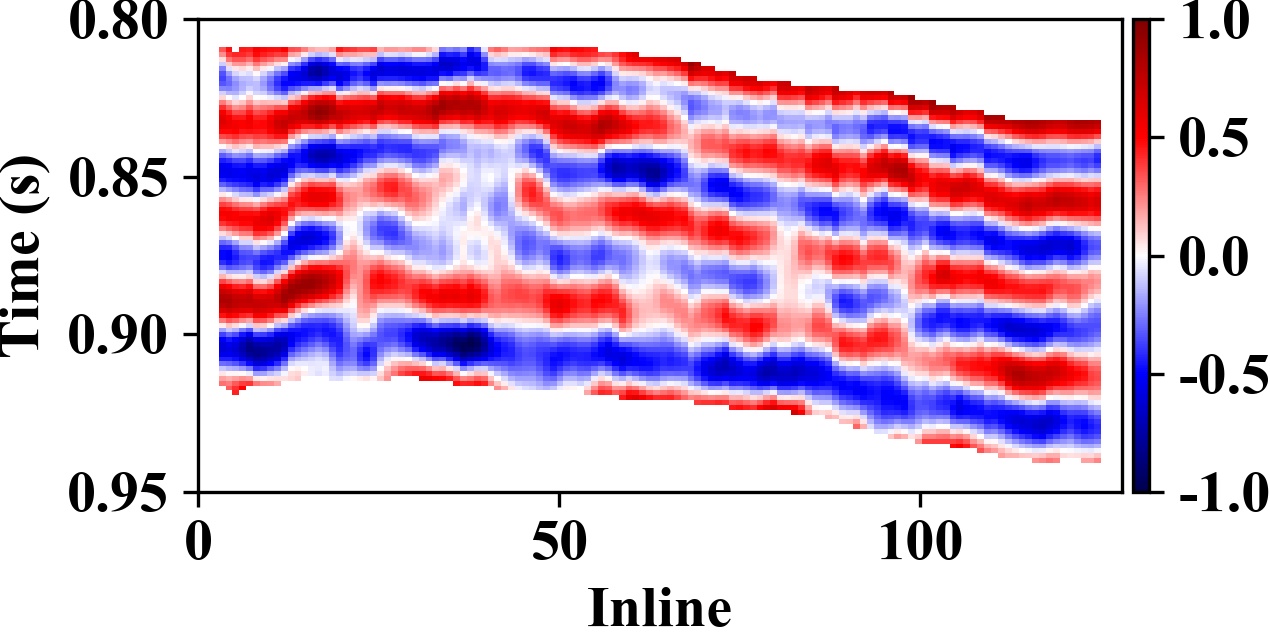}\label{fig:field_record_true}} \quad 
  \subfloat[]{\includegraphics[width = 0.32\textwidth]{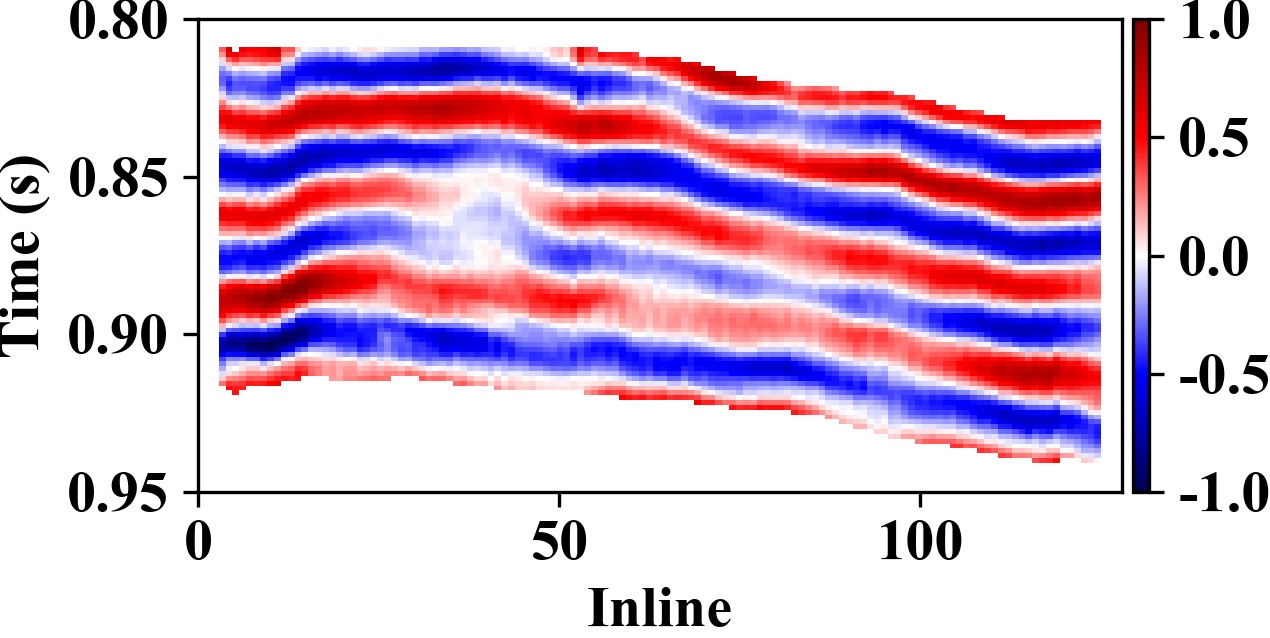}\label{fig:field_record_ldm}} \quad 
  \subfloat[]{\includegraphics[width = 0.32\textwidth]{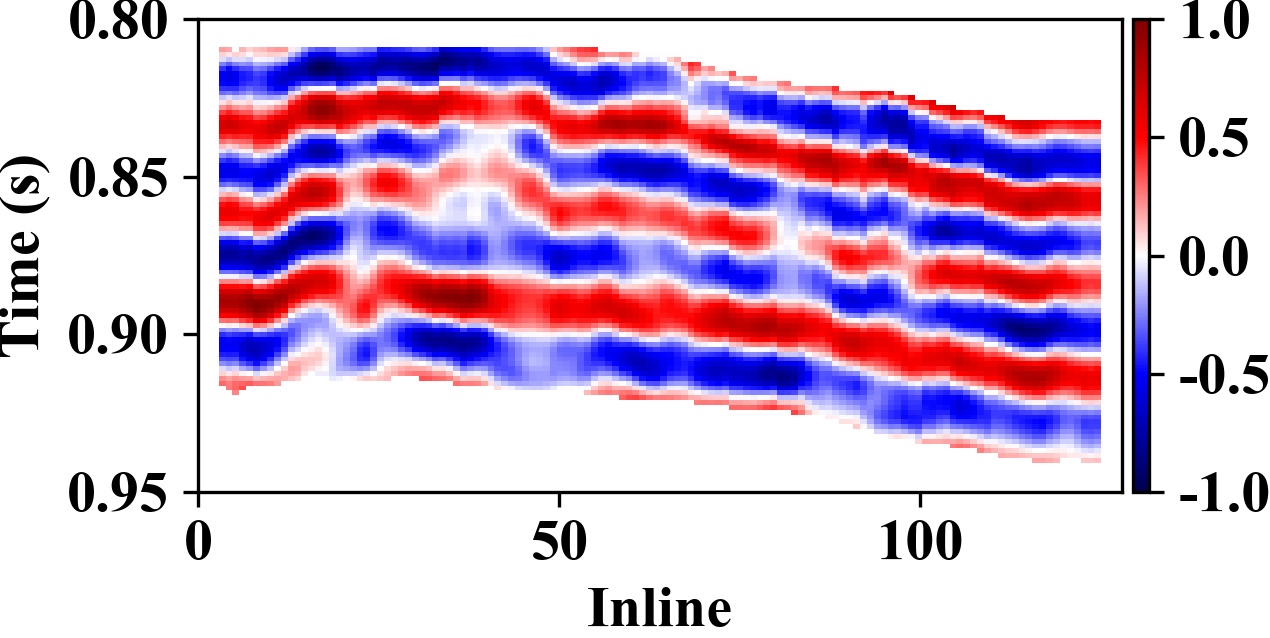}\label{fig:field_record_sup}} \\
  \subfloat[]{\includegraphics[width = 0.32\textwidth]{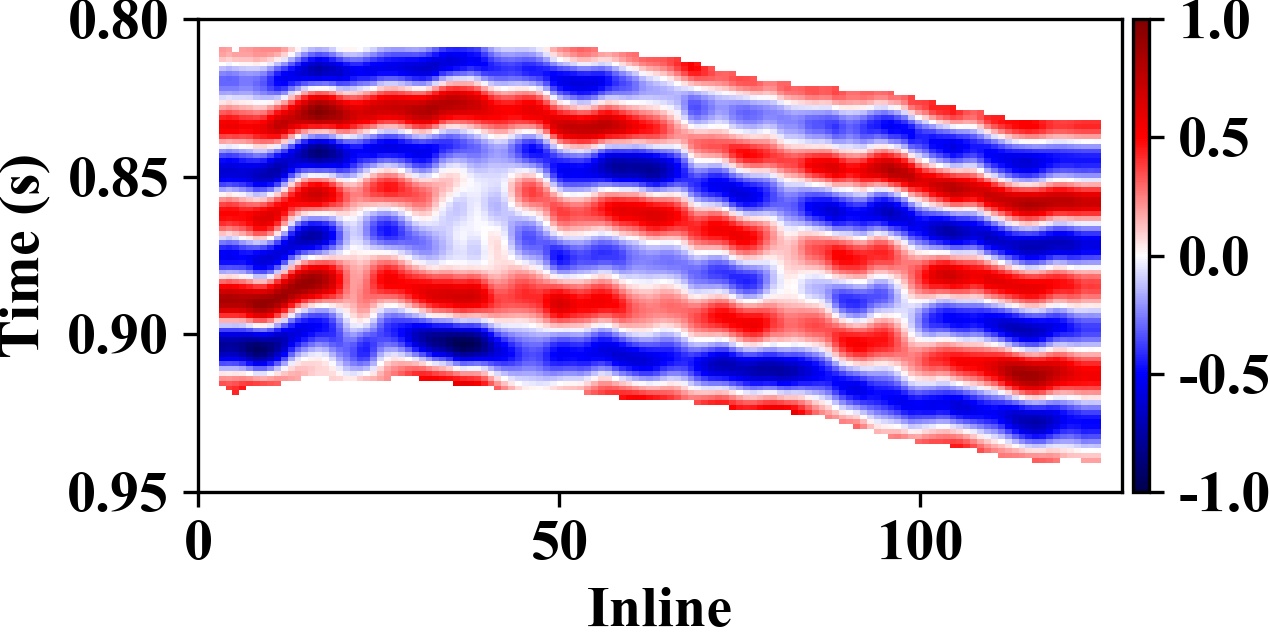}\label{fig:field_record_unsup}}\quad \quad
  \subfloat[]{\includegraphics[width = 0.32\textwidth]{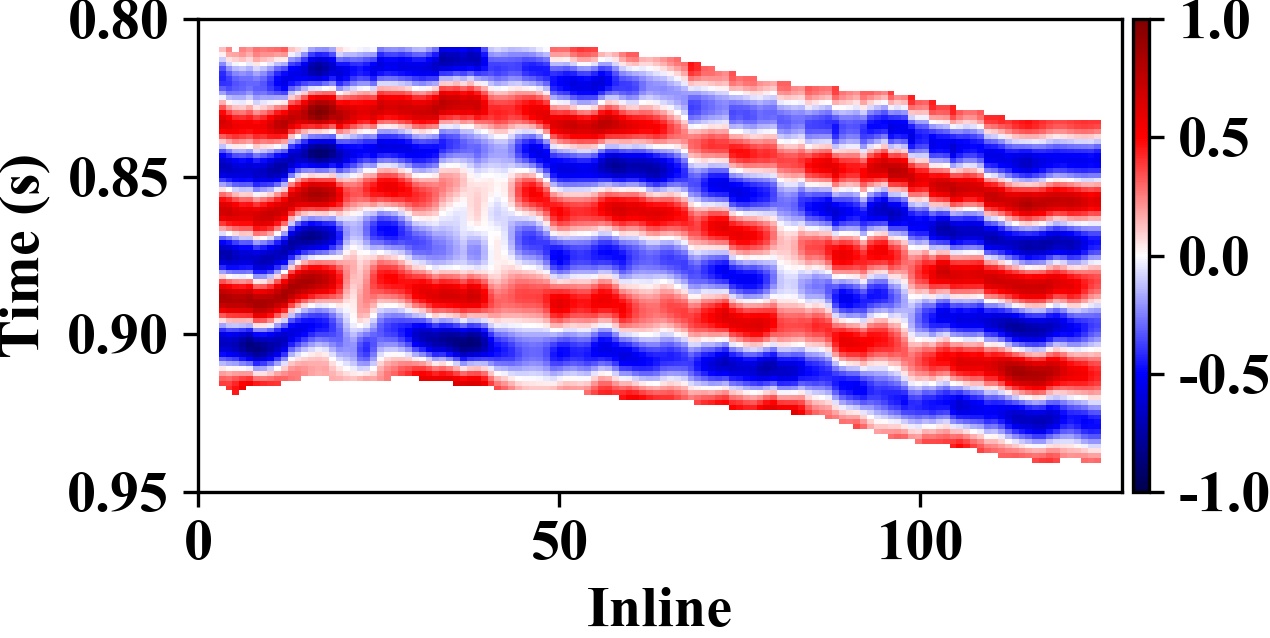}} 
  \caption{Field data reconstructed using inverted acoustic impedance. (a) field Data, (b) SAII-CLDM, (c) SDL, (d)USDL, (e) 2D-TV.}
    \label{fig:field_records_reconstrust}
\end{figure*}

Moreover, to validate the integrity of the inverted impedance within the frequency band of the seismic data, we apply the forward operator defined in Eq.~\eqref{eq:synthetic_record_noise} to reconstruct the seismic data. The observed field seismic data and the seismic data reconstructed from the inversion results in Fig.~\ref{fig:field_imps} are shown in Fig.~\ref{fig:field_records_reconstrust}(a) and Fig.~\ref{fig:field_records_reconstrust}(b)--(e), respectively. Visually, all reconstructed datasets exhibit a strong resemblance to the field data, and the result of SAII-CLDM appears slightly smoother.
This smoothing occurs because the seismic data consistency in the model-driven optimization (Eq.~\eqref{eq:model_driven_optimization}) is not strictly enforced, as the optimization intensity is governed by the learning rate and the number of iterations. A moderate optimization strength enables the model to balance the learned prior and seismic data consistency, rather than strictly fitting the observed data, thereby avoiding potential inaccuracies and ensuring more reliable inversion results.

We also calculate the PCC values between the reconstructed and observed data, which are 0.9363, 0.8812, 0.9665, and 0.9436 for the SAII-CLDM, SDL, USDL, and 2D-TV methods, respectively. These results further confirm that the proposed inversion method effectively preserves the impedance integrity within the frequency band of seismic data. In summary, the field data experiments demonstrate that the proposed SAII-CLDM method exhibits significant potential for high-resolution impedance inversion.

\begin{table}[ht]
    \centering
    \caption{Comparison of Inversion Results Using Different Methods At Well Locations}
    \begin{tabular}{lcccc}
        \toprule 
        \cmidrule(lr){2-5}
       PCC & SAII-CLDM & SDL & USDL & 2D-TV \\
        \midrule
        well1 & 0.7160 & 0.2666 & 0.3392 & 0.4046 \\
        well2 & 0.6948 & 0.5889 & 0.4709 & 0.4575 \\
        \bottomrule
    \end{tabular}
    \label{tab:well_pcc}
\end{table}

\section{Discussion}
\label{sec:discussion}
We propose a seismic acoustic impedance inversion method based on the conditional latent generative diffusion model (SAII-CLDM), which offers four distinct advantages in addressing impedance inversion problems.  

First, SAII-CLDM effectively captures the complex prior conditional distribution within the training dataset through the conditional latent generative diffusion model. This capability enhances the accuracy and resolution of inversion results. As synthetic and field data experiments demonstrate, the proposed method achieves higher accuracy, clearer structures, and greater noise tolerance than the SDL, USDL, and 2D-TV methods. These results confirm the statistical advantages of SAII-CLDM over traditional and alternative deep learning techniques. Secondly, SAII-CLDM operates within a low-dimensional latent space, allowing for an increase in the inversion size. This demonstrates the potential of diffusion models in handling large-scale seismic inversion problems, making them more applicable in real-world scenarios. Third, we incorporate a model-driven strategy in the sampling process to address the distribution shifts between the training and test datasets, further enhancing inversion accuracy and achieving high accuracy in a few timesteps. Fourth, the proposed method establishes a versatile inversion framework that can be extended to a wide range of inverse problems.

Beyond the above advantages, it is also worth noting that other generative paradigms such as GANs and VAEs have also been explored for inverse problems. However, recent studies report that diffusion-based generative priors tend to offer more stable training and higher-quality reconstructions than GAN or VAE-based counterparts, especially when dealing with highly ill-posed imaging tasks~\cite{df_Dhariwal2021guide,multimodal_llerFran2023Science}. For this reason, and to keep the scope focused, we restrict our investigation to diffusion models in this work, while a systematic experimental comparison with alternative generative priors is left for future research. In addition,due to the conditional training mechanism of SAII-CLDM, retraining or fine-tuning may be necessary when the distribution of conditional inputs differs significantly from that in the training data. Fine-tuning techniques such as ControlNet \cite{controlnet_zhang2023addingconditionalcontroltexttoimage} can alleviate this issue by enabling efficient adaptation to new datasets without full retraining.

\section{Conclusion}  
\label{sec:conclusion}
We propose an impedance inversion framework based on a conditional latent generative diffusion model. The framework learns the impedance conditional distribution through the conditional diffusion process in the latent space, enabling larger inversion sizes compared to pixel space. Besides, we design a wavelet-based module to efficiently project the seismic data into the latent space without training overhead. After training, seismic data and low-frequency impedance can be used as conditional input to infer impedance. Additionally, we introduce a model-driven sampling strategy to further enhance inversion accuracy. The synthetic experiments show that the proposed method can invert high-resolution impedance results in a few timesteps and has great robustness to noise and generalization capabilities. The field experiments show that the result of the proposed method is better aligned with impedance logs compared to SDL, USDL, and 2D-TV methods.

%\section{ACKNOWLEDGEMENTS}  
%%The research was supported by the National Natural Science Foundation of China under grant No. 42404122, the China Postdoctoral Science Foundation under grant No. 2024M762570, and the Postdoctoral Fellowship Program of CPSF under grant No. GZB20230580. 
%We appreciate Yang Tao's help in testing the code and applying it to real data.
%

\bibliography{bib/IEEEexample}
%\bibliography{bib/IEEEexample.bib}

\appendices
\section{Conditional Diffusion Model}
\label{appen:condition_ddpm_format}
Referring to \cite{df_Dhariwal2021guide}, we prove that the conditional diffusion process \(\hat{q}\) employed in our method retains the same forward and reverse transition properties as the standard, unconditional diffusion process \(q\) (see Section~\ref{sec:review}). 
 For brevity, all conditional inputs (\(\boldsymbol{l_z}\) and \(\boldsymbol{d_z}\)) are collectively denoted as \(\boldsymbol{y}\). The conditional process \(\hat{q}\) is defined analogously to \(q\), except that it is augmented with \(\boldsymbol{y}\):
\begin{align}
\hat{q}(\boldsymbol{x}_0) &:= q(\boldsymbol{x}_0), \\
\hat{q}(\boldsymbol{y} \mid \boldsymbol{x}_0) &:= \text{Known condition per sample},\\
\hat{q}(\boldsymbol{x}_{t+1} \mid \boldsymbol{x}_t, \boldsymbol{y}) &:= q(\boldsymbol{x}_{t+1} \mid \boldsymbol{x}_t), \\
\hat{q}(\boldsymbol{x}_{1:T} \mid \boldsymbol{x}_0, \boldsymbol{y}) &:= \prod_{t=1}^{T} \hat{q}(\boldsymbol{x}_t \mid \boldsymbol{x}_{t-1}, \boldsymbol{y})
\end{align}
We now demonstrate that when not conditioning on \(\boldsymbol{y}\), \(\hat{q}\) behaves exactly the same as \(q\). For the one-step transition:
\begin{align}
\hat{q}(\boldsymbol{x}_{t + 1} \mid \boldsymbol{x}_t) &= \int_{\boldsymbol{y}} \hat{q}(\boldsymbol{x}_{t + 1}, \boldsymbol{y} \mid \boldsymbol{x}_t) \, d\boldsymbol{y} \\
&= \int_{\boldsymbol{y}} \hat{q}(\boldsymbol{x}_{t + 1} \mid \boldsymbol{x}_t, \boldsymbol{y}) \hat{q}(\boldsymbol{y} \mid \boldsymbol{x}_t) \, d\boldsymbol{y} \\
&= \int_{\boldsymbol{y}} q(\boldsymbol{x}_{t + 1} \mid \boldsymbol{x}_t) \hat{q}(\boldsymbol{y} \mid \boldsymbol{x}_t) \, d\boldsymbol{y} \\
&= q(\boldsymbol{x}_{t + 1} \mid \boldsymbol{x}_t) \int_{\boldsymbol{y}} \hat{q}(\boldsymbol{y} \mid \boldsymbol{x}_t) \, d\boldsymbol{y} \\
&= q(\boldsymbol{x}_{t + 1} \mid \boldsymbol{x}_t).
\end{align}
Similarly, the full trajectory satisfies
\begin{align}
\hat{q}(\boldsymbol{x}_{1:T} \mid \boldsymbol{x}_0) &= \int_{\boldsymbol{y}} \hat{q}(\boldsymbol{x}_{1:T}, \boldsymbol{y} \mid \boldsymbol{x}_0) \, d\boldsymbol{y} \\
&= \int_{\boldsymbol{y}} \hat{q}(\boldsymbol{y} \mid \boldsymbol{x}_0) \hat{q}(\boldsymbol{x}_{1:T} \mid \boldsymbol{x}_0, \boldsymbol{y}) \, d\boldsymbol{y} \\
&= \prod_{t = 1}^T q(\boldsymbol{x}_t \mid \boldsymbol{x}_{t-1}) \\
&= q(\boldsymbol{x}_{1:T} \mid \boldsymbol{x}_0). \label{eq:qq}
\end{align}
From Equation \ref{eq:qq}, we can derive \(\hat{q}(\boldsymbol{x}_t)\):
\begin{align}
\hat{q}(\boldsymbol{x}_t) 
&= \int_{\boldsymbol{x}_{0:t-1}} \hat{q}(\boldsymbol{x}_0) \hat{q}(\boldsymbol{x}_1, \ldots, \boldsymbol{x}_t \mid \boldsymbol{x}_0) \, d \boldsymbol{x}_{0:t-1} \\
&= \int_{\boldsymbol{x}_{0:t-1}} q(\boldsymbol{x}_0) q(\boldsymbol{x}_1, \ldots, \boldsymbol{x}_t \mid \boldsymbol{x}_0) \, d \boldsymbol{x}_{0:t-1} \\
&= \int_{\boldsymbol{x}_{0:t-1}} q(\boldsymbol{x}_0, \ldots, \boldsymbol{x}_t) \, d \boldsymbol{x}_{0:t-1} \\
&= q(\boldsymbol{x}_t). \label{eq:q1}
\end{align}

Using Bayes' rule, we then conclude that the reverse (conditional) transitions satisfy
\[
\hat{q}(\boldsymbol{x}_t \mid \boldsymbol{x}_{t+1}) = q(\boldsymbol{x}_t \mid \boldsymbol{x}_{t+1}).
\]
This demonstrates that incorporating the conditional variable \(\boldsymbol{y}\) does not alter the core transition properties of the diffusion process. Consequently, the loss function for our conditional diffusion model remains consistent with that of the unconditional version. In addition, the auxiliary parameter \(\phi\) (associated with the SHWT module) is considerably smaller than the primary UNet parameter \(\theta\); hence, its effect on the overall loss function is negligible. Collectively, these derivations confirm that the loss function in Eq.~\eqref{eq:cldm_loss} remains valid.

%transition between the high-dimensional impedance domain and the learned latent space 
%with minimal information loss.并且低频背景可以不经过训练共用vq-gan.

\section{Validation of the Learned Latent Space}
\label{sec:latent_validation}

\begin{figure}[htbp]
  \centering
  % Marmousi
  \subfloat[\label{fig:vq_marm_origin}]{
      \includegraphics[width=0.16\textwidth]{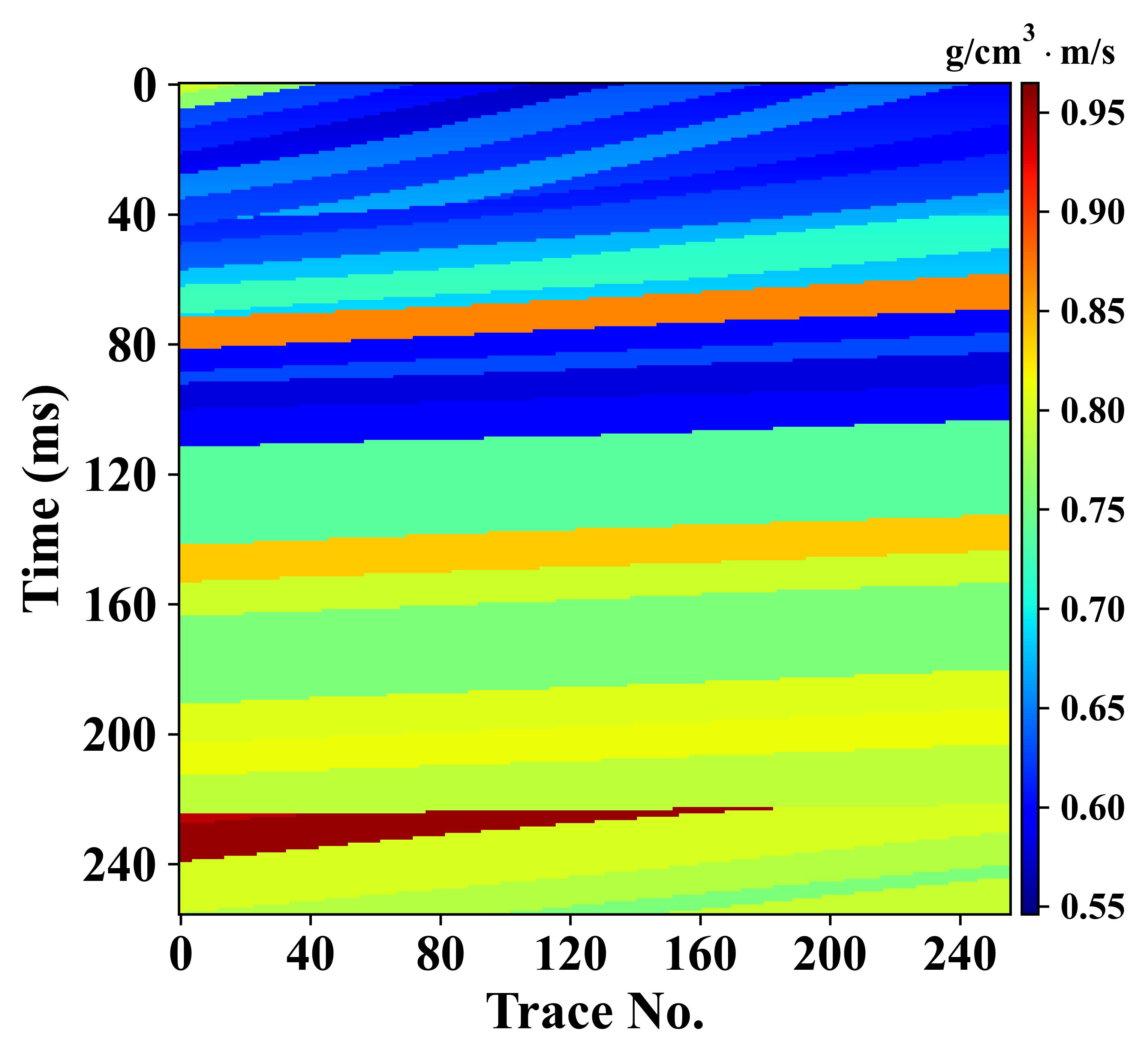}}
  \subfloat[\label{fig:vq_marm_recon}]{
      \includegraphics[width=0.16\textwidth]{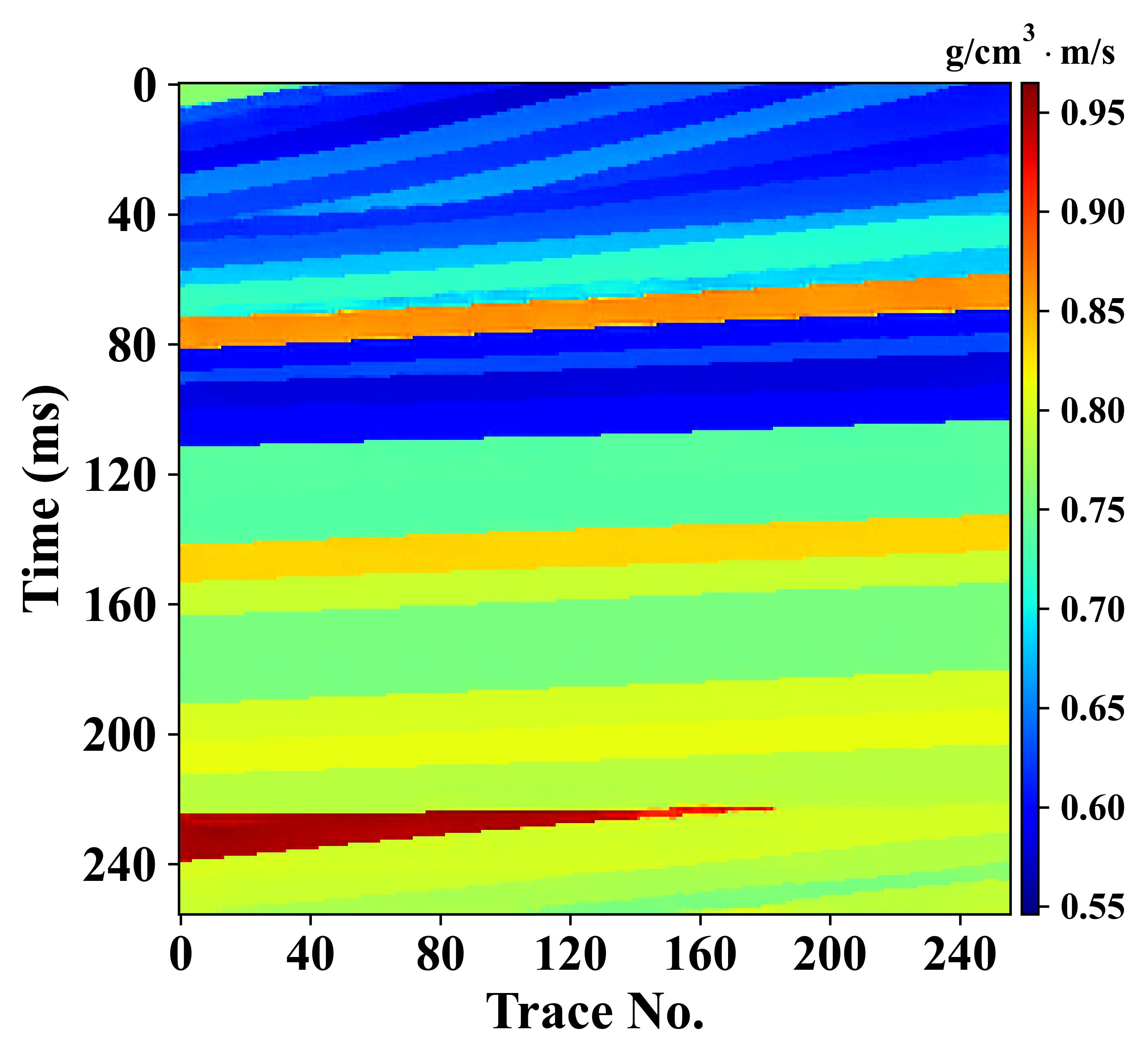}}
  \subfloat[\label{fig:vq_marm_quant}]{
      \includegraphics[width=0.16\textwidth]{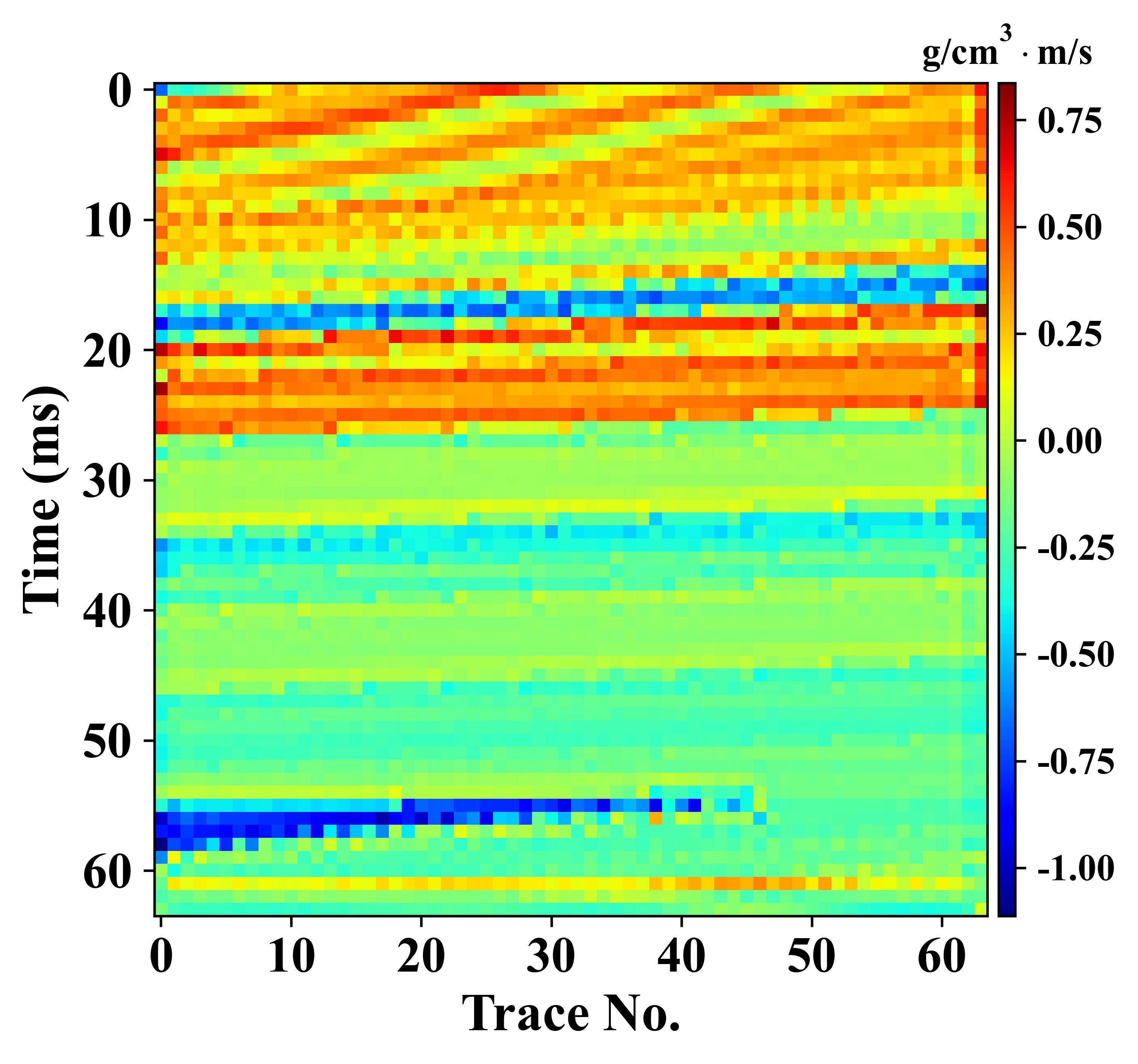}}\\
  % Overthrust
  \subfloat[\label{fig:vq_over_origin}]{
      \includegraphics[width=0.16\textwidth]{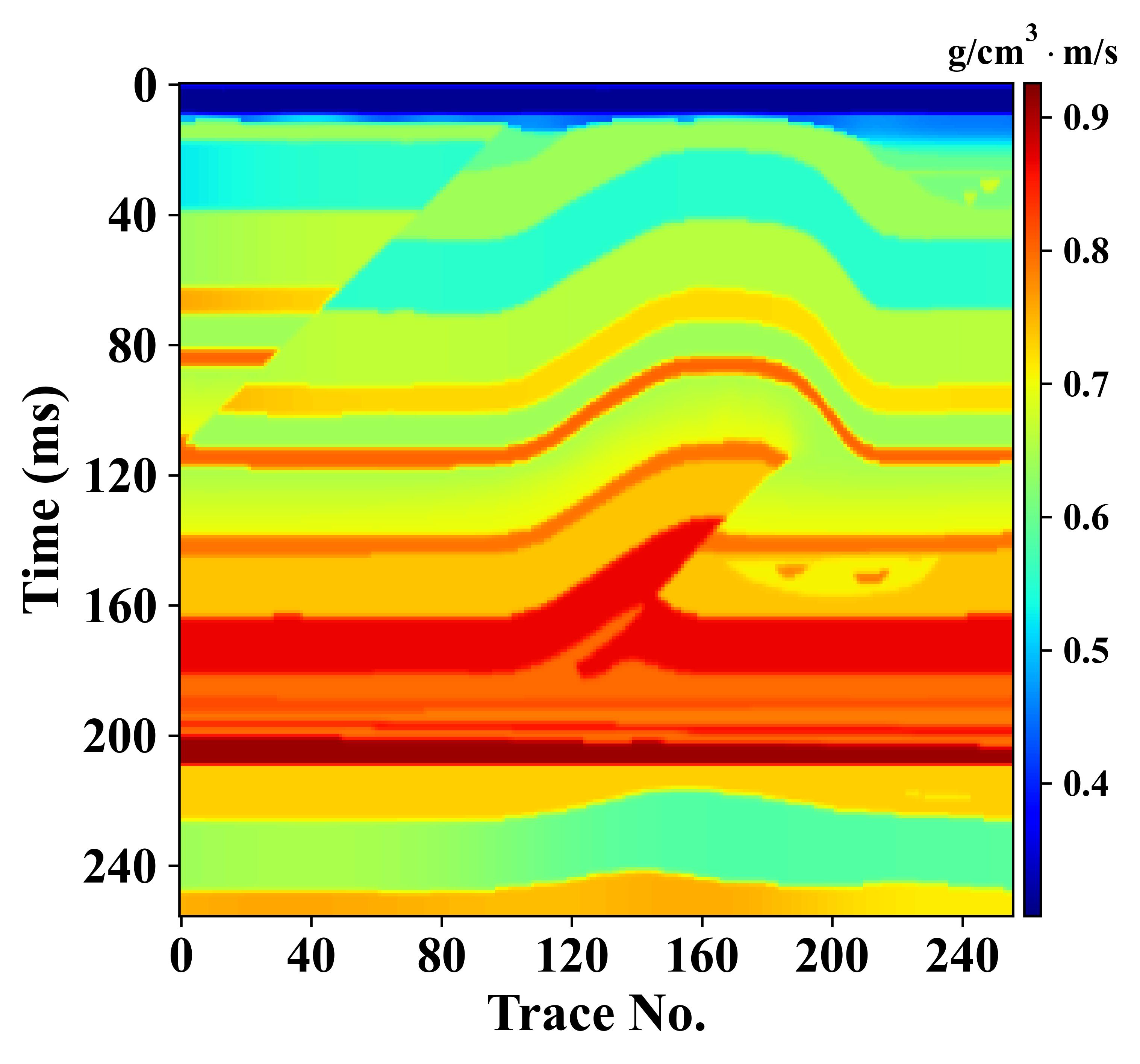}}
  \subfloat[\label{fig:vq_over_recon}]{
      \includegraphics[width=0.16\textwidth]{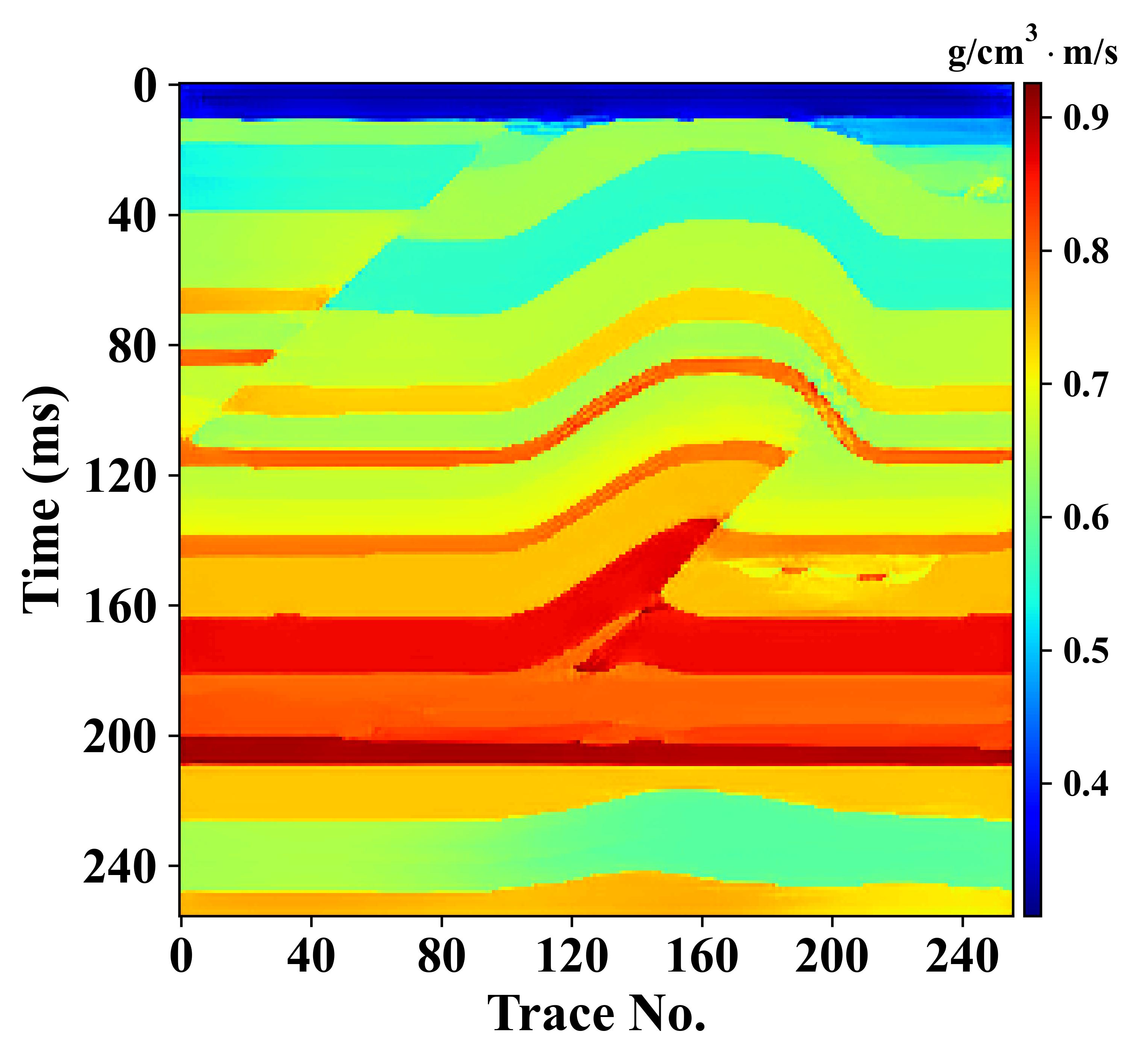}}
  \subfloat[\label{fig:vq_over_quant}]{
      \includegraphics[width=0.16\textwidth]{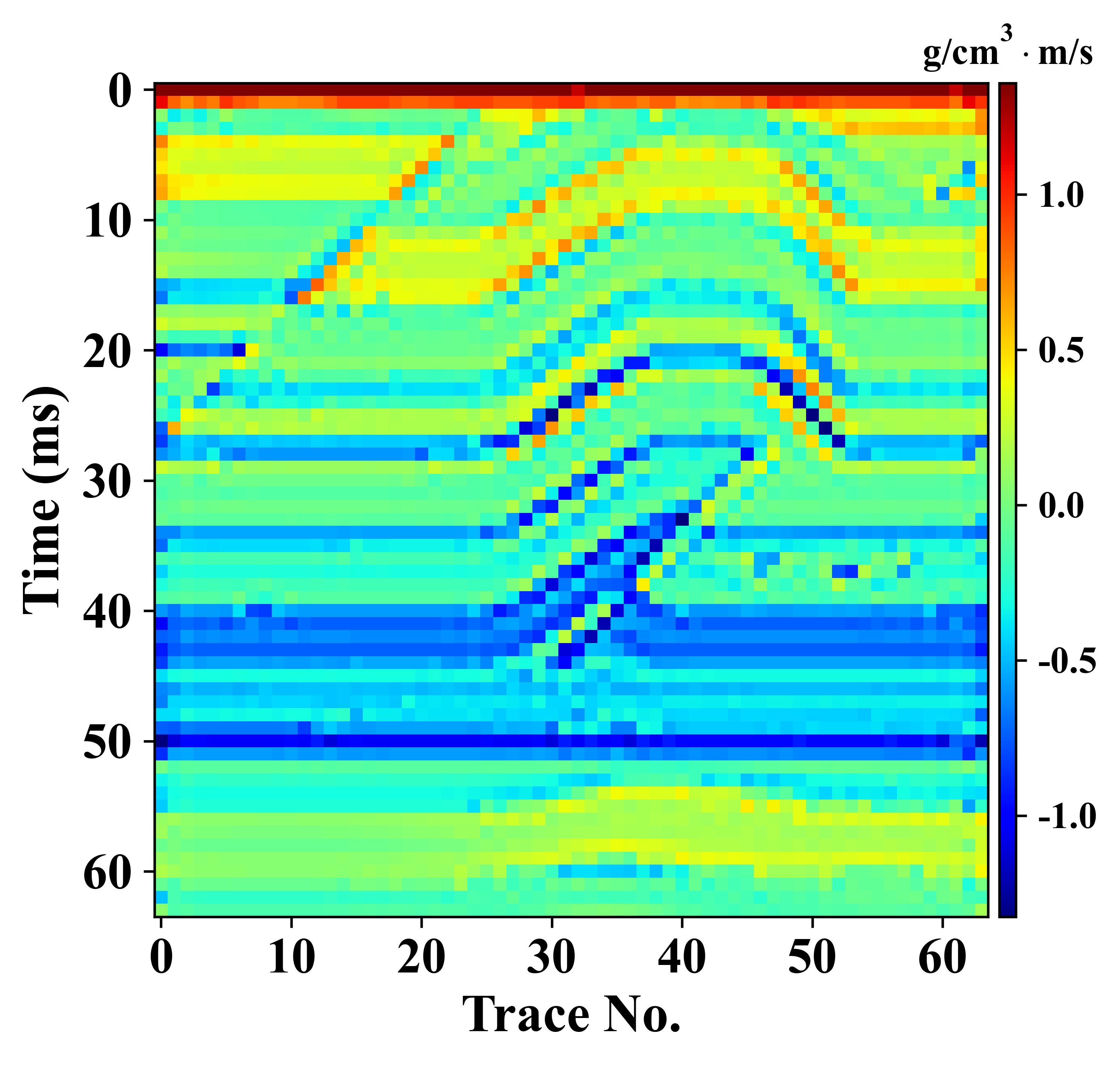}}

  \caption{
    VQ-GAN projection and reconstruction.  
    (a–c) Marmousi dataset: original, reconstruction, and latent representation.  
    (d–f) Overthrust dataset: original, reconstruction, and latent representation.
  }
  \label{fig:vqgan_recon}
\end{figure}

This section validates that the VQ-GAN encoder--decoder provides a compact latent representation that is suitable for subsequent inversion. As shown in Fig.~\ref{fig:vqgan_recon}, the reconstructed Marmousi patch remains highly consistent with the original impedance, with a Pearson correlation coefficient (PCC) of 0.9985. For the Overthrust patch from the testing domain, the reconstruction still achieves a PCC of 0.9922, indicating that the main impedance structures are well preserved, although some fine-scale details are indeed lost.
In addition, the latent representations (Figs.~\ref{fig:vq_marm_quant} and~\ref{fig:vq_over_quant}) remain spatially aligned with the original impedance models, preserving the overall geological structures in a compressed form. This is supported by the convolutional–attention architecture of VQ-GAN, where each latent token corresponds to a fixed receptive-field region enhanced by local self-attention, allowing the latent space to retain the semantic and geophysical structure of the impedance model.

\begin{figure}[htbp]
  \centering
  \subfloat[\label{fig:vq_low_origin}]{
      \includegraphics[width=0.16\textwidth]{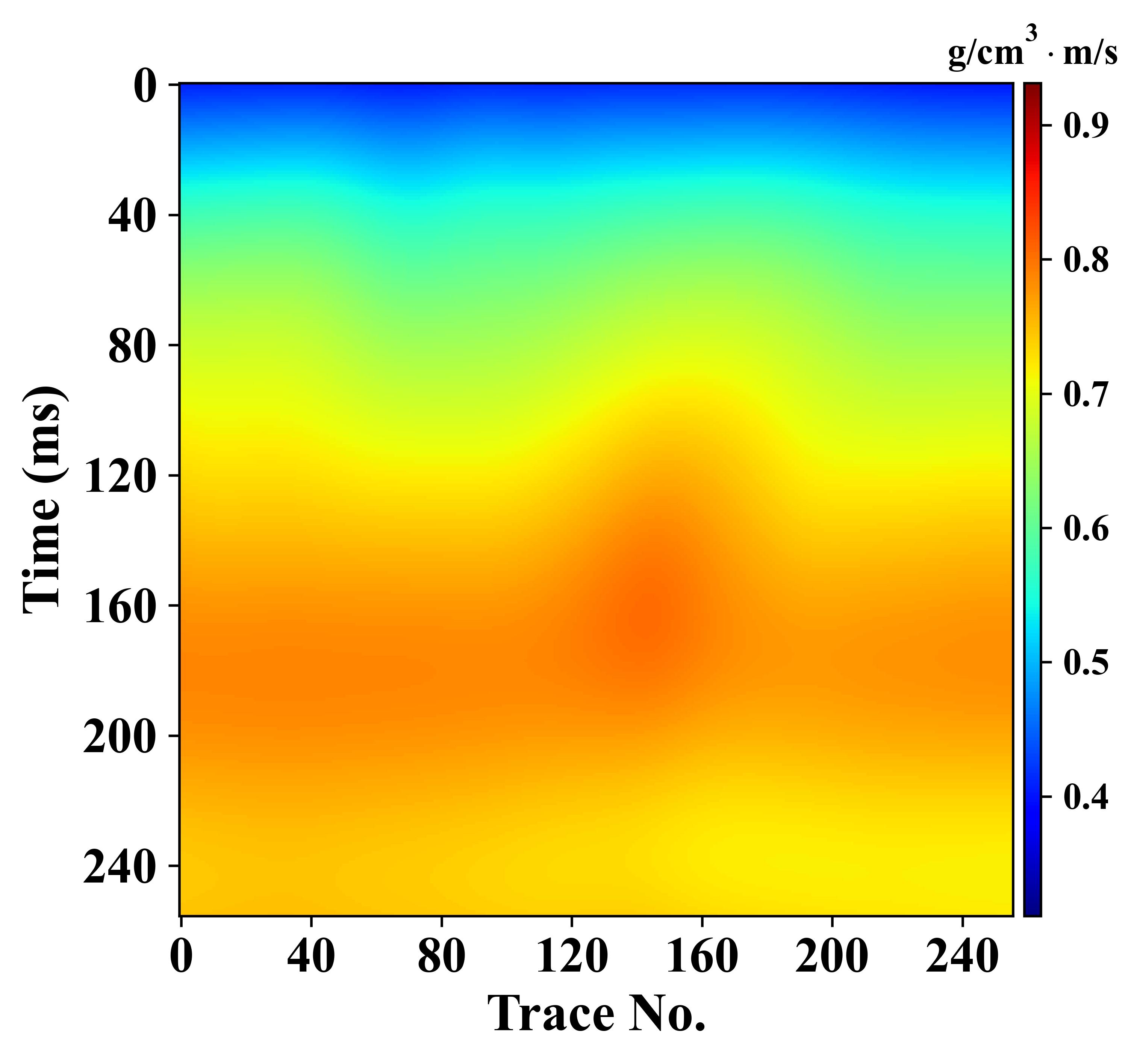}}
  \subfloat[\label{fig:vq_low_recon}]{
      \includegraphics[width=0.16\textwidth]{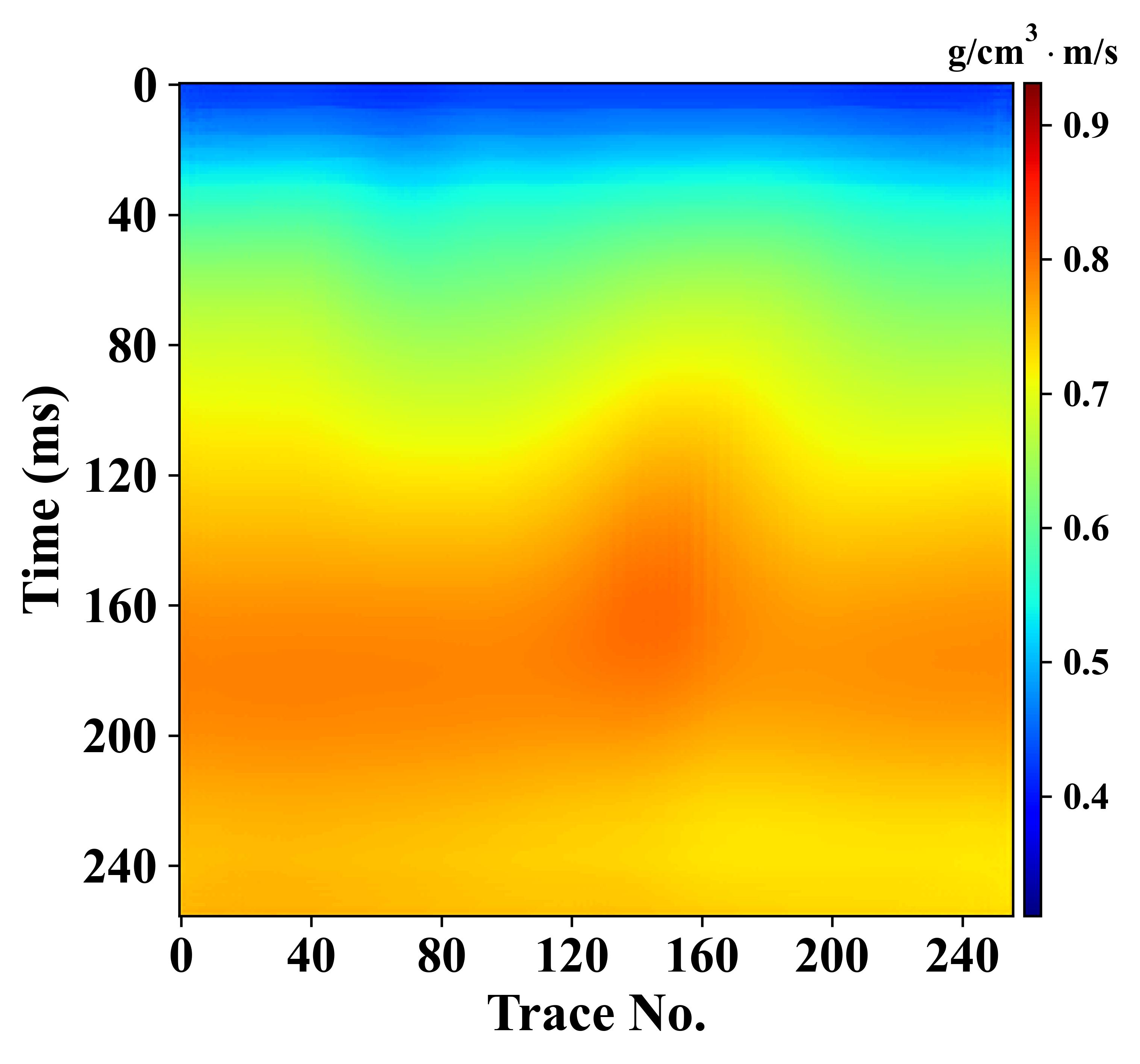}}
  \subfloat[\label{fig:vq_low_quant}]{
      \includegraphics[width=0.16\textwidth]{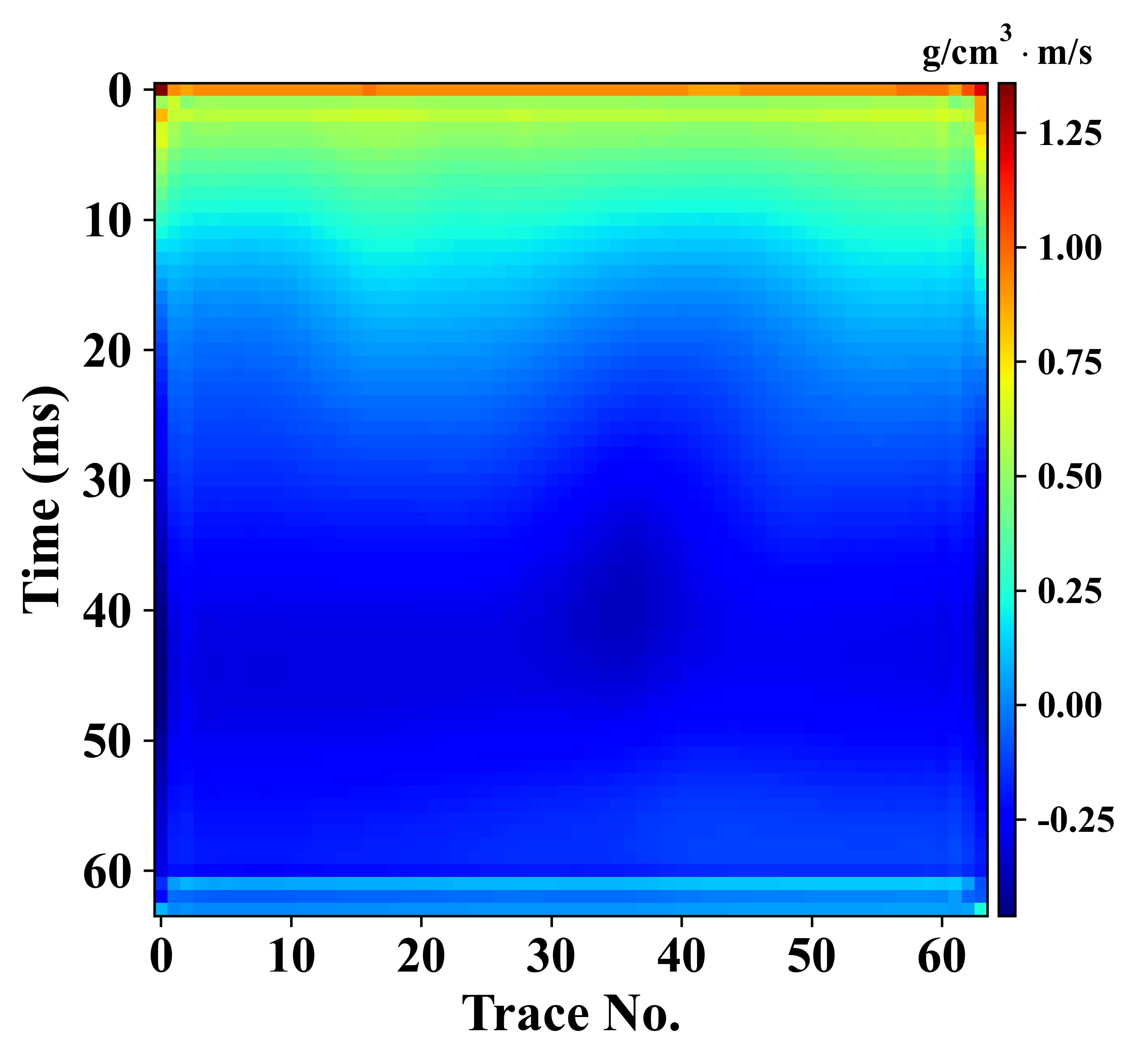}}

  \caption{
    VQ-GAN projection of 6~Hz low-frequency Overthrust impedance.  
    (a) filtered impedance, (b) reconstruction, and (c) latent representation.
  }
  \label{fig:vqgan_lowfreq}
\end{figure}

We further apply a 6~Hz low-pass filter to the Overthrust impedance and reuse the same VQ-GAN, without retraining, to encode and decode the resulting low-frequency background (Fig.~\ref{fig:vqgan_lowfreq}). The reconstruction attains a PCC of 0.9995, demonstrating that low-frequency backgrounds can be robustly projected into the same latent space by a VQ-GAN trained only on impedance model. These results confirm that the learned latent space is expressive enough for conditional diffusion-based inversion, while remaining low-dimensional and computationally efficient.

% Can use something like this to put references on a page
% by themselves when using endfloat and the captionsoff option.
\ifCLASSOPTIONcaptionsoff
  \newpage
\fi

% trigger a \newpage just before the given reference
% number-used to balance the columns on the last page
% adjust value as needed-may need to be readjusted if
% the document is modified later
\IEEEtriggeratref{8}
% The "triggered" command can be changed if desired:
\IEEEtriggercmd{\enlargethispage{-5in}}

% references section

% can use a bibliography generated by BibTeX as a .bbl file
% BibTeX documentation can be easily obtained at:
% http://mirror.ctan.org/biblio/bibtex/contrib/doc/
% The IEEEtran BibTeX style support page is at:
% http://www.michaelshell.org/tex/ieeetran/bibtex/

%\bibliography{bib/IEEEabrv}

% You can push biographies down or up by placing
% a \vfill before or after them. The appropriate
% use of \vfill depends on what kind of text is
% on the last page and whether or not the columns
% are being equalized.

%\vfill

% Can be used to pull up biographies so that the bottom of the last one
% is flush with the other column.
%\enlargethispage{-5in}

\end{CJK}
% that's all folks
\end{document}